\documentclass{article}

\usepackage{arxiv}

\usepackage[utf8]{inputenc} % allow utf-8 input
\usepackage[T1]{fontenc}    % use 8-bit T1 fonts
\usepackage{url}            % simple URL typesetting
\usepackage{booktabs}       % professional-quality tables
\usepackage{amsfonts}       % blackboard math symbols
\usepackage{nicefrac}       % compact symbols for 1/2, etc.
\usepackage{microtype}      % microtypography
\usepackage{lipsum}
\usepackage{amsmath}
\usepackage{amsthm}
\usepackage{amssymb}   
\usepackage{bigints}
\usepackage{bm}
\usepackage{diagbox}
\usepackage{makecell}
\usepackage{multirow}
\usepackage{caption}
\usepackage{subcaption}
\usepackage{graphicx}
\usepackage[ruled,vlined]{algorithm2e}
\include{pythonlisting}
\usepackage{xcolor}
\usepackage{mathtools}
\usepackage{enumitem}

\usepackage{hyperref}
\hypersetup{
    colorlinks=true,
    citecolor=black,
    linkcolor=black,
    filecolor=black,
    urlcolor=black,
}

\newcommand{\R}{\mathbb{R}}

\theoremstyle{definition}

\numberwithin{equation}{section}

\theoremstyle{plain}

\newtheorem{theorem}{Theorem}[section]
\newtheorem{definition}[theorem]{Definition}

\newtheorem{lemma}[theorem]{Lemma}

\title{Improved architectures and training algorithms for deep operator networks}

\author{
  Sifan Wang \\
  Graduate Group in Applied Mathematics \\
  and Computational Science \\
  University of Pennsylvania\\
  Philadelphia, PA 19104 \\
  \texttt{sifanw@sas.upenn.edu} \\
   \And
    Hanwen Wang \\
  Graduate Group in Applied Mathematics \\
  and Computational Science \\
  University of Pennsylvania\\
  Philadelphia, PA 19104 \\
  \texttt{wangh19@sas.upenn.edu} \\
   \And
  Paris Perdikaris \\
  Department of Mechanichal Engineering \\
  and Applied Mechanics\\
  University of Pennsylvania\\
  Philadelphia, PA 19104 \\
  \texttt{pgp@seas.upenn.edu} \\
}

\begin{document}
\maketitle
\begin{abstract}
Operator learning techniques have recently emerged as a powerful tool for learning maps between infinite-dimensional Banach spaces. Trained under appropriate constraints, they can also be effective in learning the solution operator of partial differential equations (PDEs) in an entirely self-supervised manner. In this work we analyze the training dynamics of deep operator networks (DeepONets) through the lens of Neural Tangent Kernel (NTK) theory, and reveal a bias that favors the approximation of functions with larger magnitudes. To correct this bias we propose to adaptively re-weight the importance of each training example, and demonstrate how this procedure can effectively balance the magnitude of back-propagated gradients during training via gradient descent. We also propose a novel network architecture that is more resilient to vanishing gradient pathologies. Taken together, our developments provide new insights into the training of DeepONets and consistently improve their predictive accuracy by a factor of 10-50x, demonstrated  in the challenging setting of learning PDE solution operators in the absence of paired input-output observations. All code and data accompanying this manuscript will be made publicly available at \url{https://github.com/PredictiveIntelligenceLab/ImprovedDeepONets.}
\end{abstract}

% keywords can be removed

\section{Introduction}

% introduction and motivation
\paragraph{Motivation:} 
Operator learning techniques have recently attracted significant attention thanks to their effectiveness and favorable complexity in approximating maps between infinite-dimensional Banach spaces \cite{lanthaler2021error, kovachki2021universal, yu2021arbitrary}. Techniques such as deep operator networks (DeepONets) \cite{lu2021learning}, the family of neural operator methods \cite{kovachki2021neural}, and operator-valued kernel methods \cite{owhadi2020ideas, kadri2016operator} have demonstrated promise in building fast surrogate models for emulating complex physical processes, opening new avenues for sampling, inference and uncertainty quantification in very high-dimensional parameter spaces. More recently, Wang {\it et al.} \cite{wang2021learning, wang2021long} introduced physics-informed DeepONets; a self-supervised framework for learning the solution operator of parametric partial differential equations (PDEs), even in the absence of labelled training data. These new approaches allow deep neural networks to seamlessly generalize across different scenarios (e.g. initial/boundary conditions, random inputs, different geometries, etc.), making it easier to model complex physical and engineering systems, and to do so orders of magnitude faster.

% expand on the limitations and open questions
\paragraph{Open challenges and related work:} 
Network initialization and data normalization are known to be extremely important factors in facilitating the effectiveness of deep neural networks. Effective heuristics such as the Glorot and He initialization schemes \cite{glorot2010understanding, he2016deep}, batch and weight normalization \cite{ioffe2015batch, salimans2016weight}, and data standardization \cite{lecun2012efficient} have been become standard practice in training neural networks for supervised learning tasks in finite-dimensional vector spaces. The main goal of such techniques is to reduce inter-dependencies between examples in the training data-set, ultimately leading to more efficient back-propagation and faster  convergence of the training loss under gradient descent.
However, it is still unclear whether such techniques can be effective or need to be revised in the context of learning operators in infinite-dimensional function spaces. For instance, Di Leoni {\it et. al.} \cite{di2021deeponet} observed that DeepONets are biased towards approximating target functions with larger magnitudes, and proposed to address this by using the $L^2$ norm of each output function to normalize the training data. A similar approach was adopted by Li {\it et. al.} \cite{li2020fourier}, where the relative $L^2$ norm was used as a training objective, instead of the commonly used mean square error. Although the authors do not discuss the motivation behind this choice, here we conjecture that it serves the exact same purpose: mitigate unbalanced gradients during back-propagation that bias the model towards approximating target functions with larger magnitudes. Here we must notice that, while it is evident that such heuristics are crucial, they may not always be applicable and effective in practice. For instance, if only sparse measurements of the target functions are available one cannot accurately estimate their norm and employ the aforementioned normalization. Moreover, in a semi-supervised or self-supervised setting (as is the case for physics-informed DeepONets \cite{wang2021learning, wang2021long}), the training data-set may contain no measurements of the target functions at all. Therefore, there is a pressing need to develop a deeper mathematical understanding of the training dynamics of deep operator networks, and use it to guide the development of new effective and robust architectures and training algorithms.

% contributions of this work
\paragraph{Contributions of this work:} 
Motivated by prior work on physics-informed neural networks \cite{wang2021understanding, wang2020and, mcclenny2020self, wang2021deep}, in this work we employ Neural Tangent Kernel (NTK) theory \cite{jacot2018neural,du2019gradient,allen2019convergence} to investigate the training dynamics of DeepONets and physics-informed DeepONets. Our analysis provides theoretical insight on the magnitude bias of DeepONets observed by Di Leoni {\it et. al.} \cite{di2021deeponet}, and provides a principled framework for re-weighting the importance of each training example, even in the absence of any measurements for the target output functions. This results into an adaptive learning rate annealing scheme that can continuously interpolate between balancing the magnitude of back-propagated gradients during training and balancing the convergence rate of the training loss associated with each individual example. Motivated by these findings, we also put forth a novel DeepONet architecture that is more resilient to vanishing gradient pathologies and is demonstrated to outperform conventional DeepONet for all benchmarks considered in this work. Taken together, the proposed developments introduce a 10-50x improvement in the predictive accuracy of DeepONets, which is demonstrated in the challenging setting of learning PDE solution operators in an entirely self-supervised manner (i.e. in the absence of any paired input-output observations). 
The computational infrastructure developed in this work can have a broad technical impact in enhancing the performance of DeepONets  and physics-informed DeepONets in emulating PDE systems, paving a new way to accelerating scientific modeling of complex non-linear, non-equilibrium processes in science and engineering.

\paragraph{Structure of this paper:} 
In section \ref{sec: Preliminaries}, we provide an overview of the Neural Tangent Kernel (NTK) theory following the original formulation of Jacot {\em et. al.} \cite{jacot2018neural}, the DeepONet architecture \cite{lu2021learning}, and the physics-informed DeepONet framework  \cite{wang2021learning}. Section \ref{sec: grad_pathologies} introduces a simple benchmark problem that will guide our analysis and highlight the bias of DeepONets towards learning output functions with a larger magnitude. To address this issue,  in section \ref{sec: NTK_PI_deeponet} we analyze the training dynamics of physics-informed DeepONets through the lens of NTK theory. Based on it, we propose an NTK-guided weighting scheme along with a novel DeepONet architecture in sections \ref{sec: NTK-guided weights}-\ref{sec: modified_deeponet}. To validate the effectiveness the proposed techniques, in section \ref{sec: results} we present a series of comprehensive numerical studies across a range of representative benchmarks for which conventional physics-informed DeepONets struggle.  Finally, section \ref{sec: discussion} concludes with a discussion of our main findings, potential pitfalls, and shortcomings, as well as future research directions emanating from this study.

\section{Preliminaries}
\label{sec: Preliminaries}

% move the intro of NTK to appendix?

\subsection{The Neural Tangent Kernel (NTK) of a fully-connected neural network}
\label{sec: NTK}

The Neural Tangent Kernel (NTK) is a recently proposed theoretical framework for establishing provable convergence and generalization guarantees for wide (over-parametrized) neural networks \cite{jacot2018neural,du2019gradient,allen2019convergence}. In this section, we will present a brief introduction to NTK theory and its connection to spectral bias \cite{cao2019towards, xu2019frequency, rahaman2019spectral} in the context of training multi-layer perceptron networks (MLPs). We begin by considering a scalar-valued MLP network $f(\bm{x}, \bm{\theta})$  with inputs $\bm{x} \in \R^{d_0}$  recursively defined by, 
\begin{align}
    & f(\bm{x}, \bm{\theta}) = \bm{W}^{(L+1)} \bm{f}^{(L)}(\bm{x}) + \bm{b}^{(L+1)}, \\
    &\bm{f}^{(l)}(\bm{x}) =  \frac{1}{\sqrt{d_{l-1}}}\bm{W}^{(l)} \sigma(\bm{f}^{(l-1)}(\bm{x})) + \bm{b}^{(l)} \in \R^{d_{l}}, \text{ for } k = 2, \dots, L, \\
    & \bm{f}^{(1)}(\bm{x}) = \frac{1}{\sqrt{d_{0}}}\bm{W}^{(1)}\bm{x} + \bm{b}^{(1)},
\end{align}
where $\bm{\theta} = \{\bm{W}^{(k)}, \bm{b}^{(k)}\}_{k=1}^{L+1}$ denotes all trainable parameters of the network, with all weights and biases initialized by sampling a Gaussian distribution $\mathcal{N}(0, 1)$. Now we can define the NTK as
\begin{align}
    k_{\bm{\theta}}(\bm{x}, \bm{x}') = \left\langle \frac{d f(\bm{x}, \bm{\theta}) }{d \bm{\theta}} , \frac{d f(\bm{x}', \bm{\theta}) }{d \bm{\theta}}  \right\rangle.
\end{align}
The seminal work of Jacot {\em et. al.} \cite{jacot2018neural} shows that the NTK converges to a deterministic kernel under the infinite width limit and training via gradient descent with an infinitesimally small learning rate. As a consequence the training dynamics of a sufficiently wide MLP network behave like a linearized model governed by the NTK.

Now consider a set of training data $\left\{\bm{X}_{\text {train}}, \bm{Y}_{\text {train}}\right\}$ with $\bm{X}_{\text {train}} = (\bm{x}_i)_{i=1}^N$, $\bm{Y}_{\text {train}} = (y_i)_{i=1}^N$, and the $\ell^2$ regression loss $\mathcal{L}(\bm{\theta})=\frac{1}{N} \sum_{i=1}^{N}\left|f\left(\bm{x}_{i}, \bm{\theta}\right)-y_{i}\right|^{2}$. Under the asymptotic conditions stated in Lee {\em et. al.} \cite{lee2019wide}, one may derive
\begin{align}
    \label{eq: training_dynamics}
    \frac{d f\left(\bm{X}_{\text {train }}, \bm{\theta}(t)\right)}{d t} \approx-\bm{K} \cdot\left(f\left(\bm{X}_{\text {train }}, \bm{\theta}(t)\right)-\bm{Y}_{\text {train }}\right),
\end{align}
where $\bm{\theta}$ denotes the parameters of the network at gradient descent iteration $t$ and $f\left(\bm{X}_{\text {train }}, \bm{\theta}(t)\right)=\left(f\left(\bm{x}_{i}, \bm{\theta}(t)\right)_{i=1}^{N}\right)$ is a vector in $\R^N$. Besides, $\bm{K} \in \R^{N \times N}$ is the NTK matrix whose entries are defined by
\begin{align}
    \bm{K}_{ij} = k_{\bm{\theta}}(\bm{x}_i, \bm{x}_j) = \left\langle \frac{d f(\bm{x}_i, \bm{\theta}) }{d \bm{\theta}} , \frac{d f(\bm{x}_j, \bm{\theta}) }{d \bm{\theta}}  \right\rangle, \quad \text{for } i,j =1,2, \dots, N.
\end{align}
One can then obtain that
\begin{align}
    f\left(\bm{X}_{\text {train}}, \bm{\theta}(t)\right) \approx    e^{-\bm{K} t}  f\left(\bm{X}_{\text {train}}, \bm{\theta}(0)\right) + \left(I-e^{-\bm{K} t}\right) \cdot \bm{Y}_{\text {train}}.
\end{align}
From the NTK definition, it is easy to see that $\bm{K}$ is positive semi-definite. Therefore, there exists an orthogonal matrix $\bm{Q} = [\bm{q}_1, \bm{q}_2, \dots, \bm{q}_N]$ such that $\bm{K} = \bm{Q \Lambda Q}^T $, where $\bm{\Lambda}$ is a diagonal matrix with diagonal entries $\lambda_i$ being the eigenvalue corresponding to the eigenvector $\bm{q}_i$. Note that $e^{-\bm{K} t} = \bm{Q} e^{-\bm{\Lambda} t} \bm{Q}^T$, yielding
\begin{align}
    \bm{Q}^{T}\left(f\left(\bm{X}_{\text{train}}, \bm{\theta}(t)\right) -     \bm{Y}_{\text{train}}\right) \approx e^{-\bm{\Lambda} t} \bm{Q}^{T} \left(f\left(\bm{X}_{\text {train}}, \bm{\theta}(0)\right) -  \bm{Y}_{\text {train}} \right).
\end{align}
Therefore, 
\begin{align}
    \left[\begin{array}{c}
\bm{q}_{1}^{T} \\
\bm{q}_{2}^{T} \\
\vdots \\
\bm{q}_{N}^{T}
\end{array}\right]\left(f\left(\bm{X}_{\text{train}}, \bm{\theta}(t)\right) - \bm{Y}_{\text {train}}\right)
\approx \left[\begin{array}{llll}
e^{-\lambda_{1} t} & & & \\
& e^{-\lambda_{2} t} & & \\
& & \ddots & \\
& & & e^{-\lambda_{N} t}
\end{array}\right]\left[\begin{array}{c}
\bm{q}_{1}^{T} \\
\bm{q}_{2}^{T} \\
\vdots \\
\bm{q}_{N}^{T}
\end{array}\right]  \left(f\left(\bm{X}_{\text {train}}, \bm{\theta}(0)\right) - \bm{Y}_{\text {train}}\right).
\end{align}
The above equation reveals that the $i$-th term of the left-hand side  will converge to $0$ at the speed of $e^{-\lambda_i t}$ if $\lambda_i > 0$. In other words, the eigenvalues of the NTK characterize the convergence rate of the total training error. We remark that this conclusion holds for any network architecture in the NTK regime \cite{wang2020and, wang2020eigenvector}.

\subsection{DeepONets and Physics-informed DeepONets}
\label{sec: deeponets}

\begin{figure}
    \centering
    \includegraphics[width=0.8\textwidth]{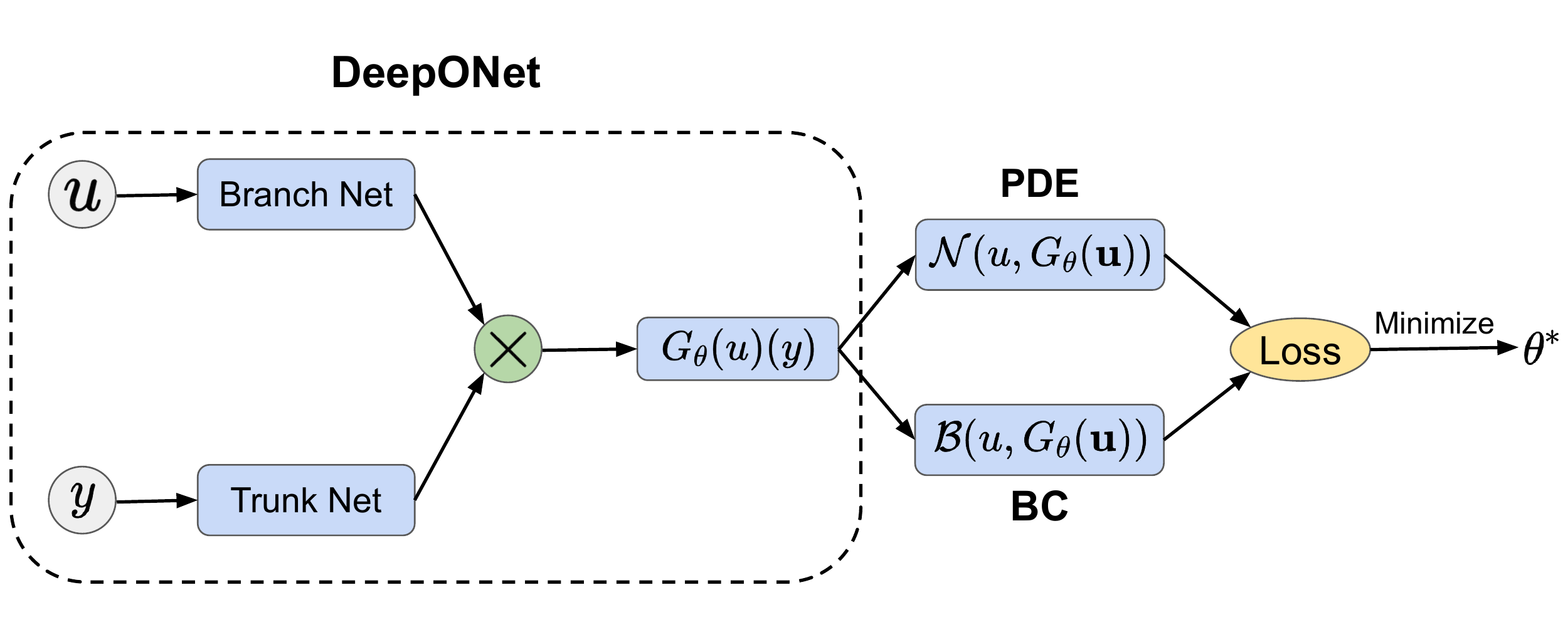}
    \caption{{\em  Physics-informed DeepONets:}  The DeepONet architecture \cite{lu2021learning} consists of two sub-networks referred as the branch network and the trunk network, which  extract  latent  representations of input functions $\bm{u}$ and input coordinates $\bm{y}$ at which the output functions are evaluated, respectively. A continuously  differentiable representation of the output functions is then obtained by merging the outputs of each sub-network via a dot product. Automatic differentiation can then be employed to formulate appropriate regularization mechanisms for biasing the DeepONet outputs to satisfy a given system of PDEs.}
    \label{fig: arch}
\end{figure}

Deep operator networks (DeepONets) \cite{lu2021learning} are a specialized deep learning architecture that aims to learn abstract nonlinear operators  between infinite-dimensional function spaces. This network architecture is inspired and validated by the rigorous universal approximation theorem for operators \cite{chen1995universal, lu2021learning}. Now we present a brief overview of DeepONets in the setting of learning the solution operator of parametric partial differential equations (PDEs) \cite{wang2021learning}. Let $(\mathcal{U}, \mathcal{V}, \mathcal{S})$ be a triplet of abstract function spaces and $\mathcal{N}: \mathcal{U} \times \mathcal{S} \rightarrow \mathcal{V}$ be a linear or nonlinear differential operator. We consider general parametric PDEs of the form
\begin{align}
    \label{eq: parametric_PDE}
    \mathcal{N}(\bm{u} , \bm{s})= 0,
\end{align}
subject to boundary conditions
\begin{align}
    \label{eq: BC}
    \mathcal{B}(\bm{u}, \bm{s}) = 0,
\end{align}
where $\bm{u} \in \mathcal{U}$ denotes the parameters (i.e. input functions), and $\bm{s} \in \mathcal{S}$ denotes an unknown latent function that is governed by the PDE system in equation \ref{eq: parametric_PDE}. Moreover,  $\mathcal{B}[\cdot] $ denotes a boundary conditions operator that enforces any Dirichlet, Neumann, Robin, or periodic boundary conditions. For time-dependent problems, we consider time $t$ as an extra component of $\bm{x}$, and $\Omega$ is extended to contain the temporal domain. In that case, initial conditions can be simply treated as a special type of boundary condition defined in the joint spatio-temporal domain.  Assume that, for any $\bm{u} \in \mathcal{U}$, there exists an unique solution $\bm{s} = \bm{s}(\bm{u}) \in \mathcal{U}$ to equation (\ref{eq: parametric_PDE}), subject to appropriate initial and boundary conditions. Then, one can define the PDE solution operator $G: \mathcal{U} \rightarrow \mathcal{S}$ as
\begin{align}
    G(\bm{u}) = \bm{s}(\bm{u}).
\end{align}
Now we can proceed by representing the solution map with a DeepONet $G_{\bm{\theta}}$, where $\bm{\theta}$ denotes all trainable parameters of the DeepOnet network. As illustrated in Fig \ref{fig: arch},  the  DeepONet is composed of two separate neural networks referred  as the “branch” and “trunk” networks, respectively. The branch network takes a function $\bm{u}$ as input and returns a features embedding $[b_1, b_2,\dots, b_q]^T \in \mathbb{R}^q$ as output, where $\bm{u} = [\bm{u}(\bm{x}_1), \bm{u}(\bm{x}_2), \dots, \bm{u}(\bm{x}_m) ]$ represents a function $\bm{u} \in \mathcal{U}$ evaluated at a collection of fixed locations $\{\bm{x}_i\}_{i=1}^m$.  The trunk network takes the continuous coordinates $\bm{y}$ as inputs, and outputs a features embedding $[t_1, t_2,\dots, t_q]^T \in \mathbb{R}^q$. The final DeepONet output is obtained by merging the outputs of branch and trunk networks together via an inner product. Mathematically, one can express the DeepONet $G_{\bm \theta}$ representation of a function $\bm{u}$ evaluated at $\bm{y}$ as
\begin{align}
    \label{eq: deeponet_output}
    G_{\bm{\theta}}(\bm{u})(\bm{y}) = \sum_{k=1}^{q} \underbrace{b_{k}\left(\bm{u}\left(\bm{x}_{1}\right), \bm{u}\left(\bm{x}_{2}\right), \ldots, \bm{u}\left(\bm{x}_{m}\right)\right)}_{\text {branch }} \underbrace{t_{k}(\bm{y})}_{\text {trunk }},
\end{align}
From this  definition,  one may note that the output of a DeepONet is a scalar function. However, sometimes we may require a DeepONet to return a vector-valued function.  To resolve this issue,  Wang {\em et. al}  \cite{wang2021long} modified the original forward pass as follows
\begin{align}
\label{eq: deeponet_multiple_outputs}
G_{\bm{\theta}}^{(i)}(\bm{u})(\bm{y}) = \sum_{k=q_{i-1}+1}^{q_{i}} \underbrace{b_{k}\left(\bm{u}\left(\bm{x}_{1}\right), \bm{u}\left(\bm{x}_{2}\right), \ldots, \bm{u}\left(\bm{x}_{m}\right)\right)}_{\text {branch }} \underbrace{t_{k}(\bm{y})}_{\text {trunk }}, \ \ i = 1, \dots, n,
\end{align}
where $0 = q_0 < q_1 < \cdots < q_{n} = q$.  This simple modification enables a DeepONet to output an n-dimensional vector.

Now let us define
\begin{align}
    \mathcal{L}(\bm{u}, \bm{\theta}) =  \frac{1}{P} \sum_{j=1}^P   \left|G_{\bm{\theta}} (\bm{u})(\bm{y}_j)- s(\bm{y}_j)  \right|^2,
\end{align}
where $\{s(y_j)\}_{j=1}^P$ denotes the associated PDE solution of equation (\ref{eq: parametric_PDE}) evaluated at $P$ locations $\{\bm{y}_j\}_{j=1}^P$ in the domain of $G(u)$. Then a DeepONet model can be trained by minimizing the following loss 
\begin{align}
    \label{eq: deeponet_loss}
       \mathcal{L}(\bm{\theta})  =
       \frac{1}{N} \sum_{i=1}^N \mathcal{L}(\bm{u}^{(i)}, \bm{\theta}),
\end{align}
where 
\begin{align}
    \mathcal{L}(\bm{u}^{(i)}, \bm{\theta}) &=  \frac{1}{P} \sum_{j=1}^P   \left|G_{\bm{\theta}} (\bm{u}^{(i)})(\bm{y}^{(i)}_j)- s^{(i)}(\bm{y}^{(i)}_j)  \right|^2 \\
    &= \frac{1}{NP} \sum_{i=1}^N \sum_{j=1}^P \left|G_{\bm{\theta}} (\bm{u}^{(i)})(\bm{y}^{(i)}_j)- s^{(i)}(\bm{y}^{(i)}_j)  \right|^2.
\end{align}
We remark that that the locations $\bm{y}$ may vary for different input samples $\bm{u}$.

In follow up work, Wang {\em et. al.} \cite{wang2021learning} developed physics-informed DeepONets, which leverages automatic differentiation \cite{baydin2018automatic} to constrain the outputs of a DeepONet model to satisfy a given governing PDE. Specifically, a physics-informed DeepONet can be trained by minimizing
\begin{align}
    \label{eq: physics_informed_deeponet_loss}
    \mathcal{L}(\bm{\theta}) = \mathcal{L}_{\text{operator}}(\bm{\theta}) +  \mathcal{L}_{\text{physics}}(\bm{\theta}),
\end{align}
where $ \mathcal{L}_{\text{operator}}(\bm{\theta})$ can be defined exactly the same as in equation (\ref{eq: deeponet_loss}), which aims to fit available experimental data and numerical estimations. Notice that this enables one to train DeepONet models even if no paired input-output observations are available, except for assuming knowledge of the initial and boundary conditions \ref{eq: BC}. Then one can define 
\begin{align}
     \sum_{i=1}^N \mathcal{L}(\bm{u}, \bm{\theta}) = \frac{1}{P}  \sum_{j=1}^P   \left|\mathcal{B}(u(\bm{x}_j), G_{\bm{\theta}}(\bm{u})(\bm{y}_j) ) \right|^2.
\end{align}
For each input sample $\bm{u}$,  $\{\bm{x}_j\}_{j=1}^P$ and  $\{\bm{y}_j\}_{j=1}^P$ denote two sets of points that are randomly sampled from the domain of $\bm{u}$ and the boundary of $ G(\bm{u})$, respectively, for imposing boundary conditions.  Consequently we can define
\begin{align}
    \mathcal{L}_{\text{operator}}(\bm{\theta}) =  \frac{1}{N} \sum_{i=1}^N \mathcal{L}(\bm{u}^{(i)}, \bm{\theta}) = \frac{1}{NP} \sum_{i=1}^N \sum_{j=1}^P   \left|\mathcal{B}(u^{(i)}(\bm{x}_j^{(i)}), G_{\bm{\theta}}(\bm{u}^{(i)})(\bm{y}_j^{(i)}) ) \right|^2,
\end{align}
and
\begin{align}
    \mathcal{L}_{\text{physics}}(\bm{u}, \bm{\theta}) = \frac{1}{Q} \sum_{j=1}^Q  \left|\mathcal{N}(u(\bm{x}_{r,j}), G_{\bm{\theta}}(\bm{u})(\bm{y}_{r,j}) ) \right|^2,
\end{align}
where $\{\bm{x}_{r,j}\}_{j=1}^Q$ and  $\{\bm{y}_{r,j}\}_{j=1}^Q$ denote two sets of collocation points that are randomly sampled from the domain of $u$ and $G(\bm{u})$, respectively, for enforcing the set of parametric PDE constraints described equation (\ref{eq: parametric_PDE}). It also follows that
\begin{align}
    \mathcal{L}_{\text{physics}}(\bm{\theta}) 
    & = \frac{1}{N}  \sum_{i=1}^N \mathcal{L}_{\text{physics}}(\bm{u}^{(i)}, \bm{\theta}) \\
    &=  \frac{1}{NQ}  \sum_{i=1}^N \sum_{j=1}^Q  \left|\mathcal{N}(u^{(i)}(\bm{x}_{r,j}^{(i)}), G_{\bm{\theta}}(\bm{u}^{(i)})(\bm{y}_{r,j}^{(i)}) ) \right|^2
\end{align}
As shown in Wang {\em et. al.} \cite{wang2021learning, wang2021long}, it is worth  pointing out that physics-informed DeepONets are capable of learning the solution operator of parametric PDEs in an entirely self-supervised manner, i.e. without any paired input-output observations.

\section{Methods}

%1. NTK with anti derivative example
%2. New archtiecture perhaps with some analysis? 

\subsection{Gradients pathologies in DeepONets}
\label{sec: grad_pathologies}

% use anti derivative example
Although DeepONets and physics-informed DeepONets  have demonstrated a series of promising results in learning nonlinear operators and solving multi-physics problems \cite{lu2021learning,cai2020deepm, di2021deeponet, wang2021learning, wang2021long}, it is worth noting that the original formulation may have difficulties in tackling cases  with multi-scale input parameters. To this end, poor approximations  can arise even in the simplest possible setting. As an pedagogical example, let us consider the following one-dimensional initial value problem
\begin{align}
    \label{eq: ODE}
    &\frac{d s(x)}{d x}=u(x), \quad x \in [0, 1], \\
    &s(0) = 0.
\end{align}
Here, our goal is to learn the solution operator $G$ mapping the  forcing term $u(x)$ to the corresponding solution $s(x)$ in $[0,1]$. Indeed, $G$ is the so-called anti-derivative operator with an explicit expression of
\begin{align}
    G: u(x) \longrightarrow s(x)=s(0)+\int_{0}^{x} u(t) d t,  \quad  x \in[0,1].
\end{align}
To assess the performance of DeepONets in learning the anti-derivative operator, we generate a training data-set by sampling $N = 10^4$ forcing terms $u$ from a Gaussian random field (GRF) with a fixed length scale $l=0.2$, and a random output scale $k \in [0.01, 100]$. 
To balance the training data-set, we set $k = 10^{\alpha}$ where $\alpha$ is sampled from a uniform distribution $\mathcal{U}(-2, 2)$. The  associated ODE solutions $s$ are obtained by integrating the ODE \ref{eq: ODE} using an explicit Runge-Kutta method (RK45) \cite{iserles2009first}. Input functions $u$ are represented via point-wise evaluations at a set of $m=100$ fixed sensors $\{x_i\}_{i=1}^m$ evenly spaced in $[0, 1]$. For each observed pair $(u, s)$, we randomly select $P=1$ observation of $s(\cdot)$ in $[0,1]$.

\begin{figure}
     \centering
     \begin{subfigure}[b]{0.45\textwidth}
         \centering
         \includegraphics[width=\textwidth]{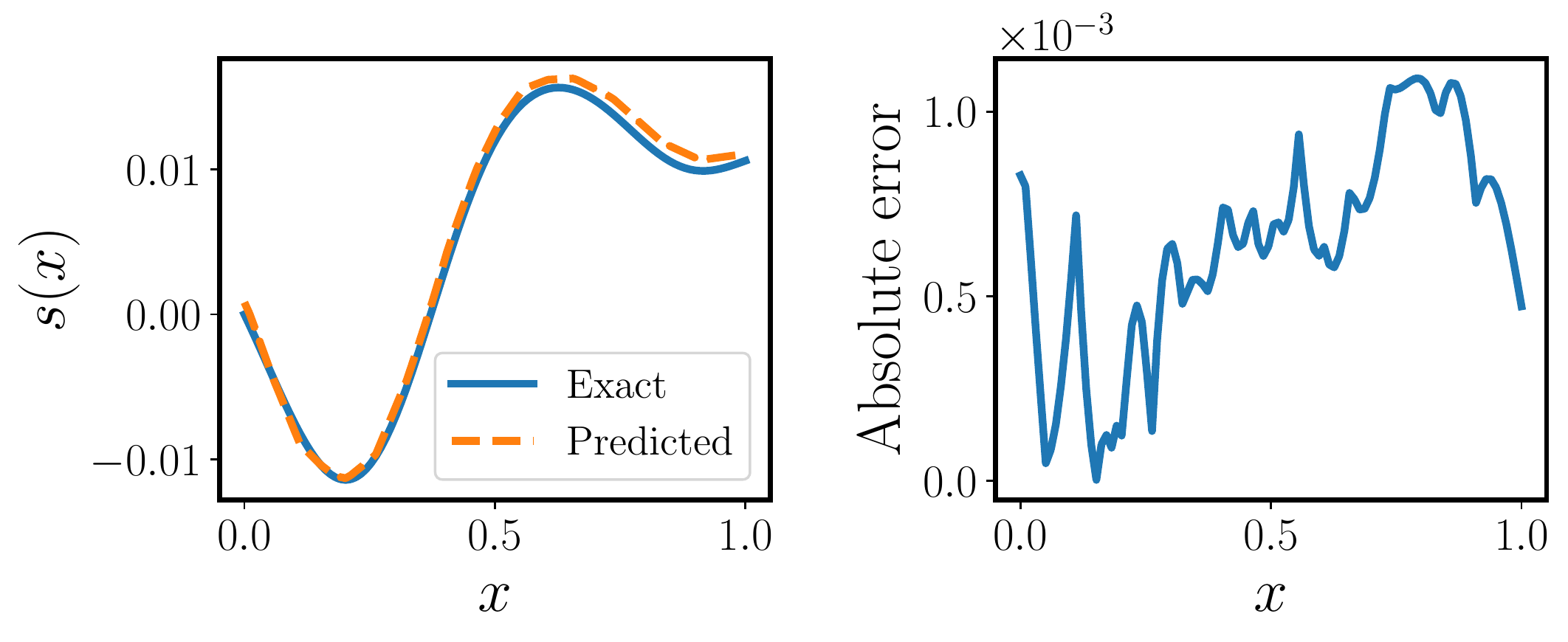}
         \label{fig: ODE_deeponet_pred_001}
     \end{subfigure}
     \begin{subfigure}[b]{0.45\textwidth}
         \centering
         \includegraphics[width=\textwidth]{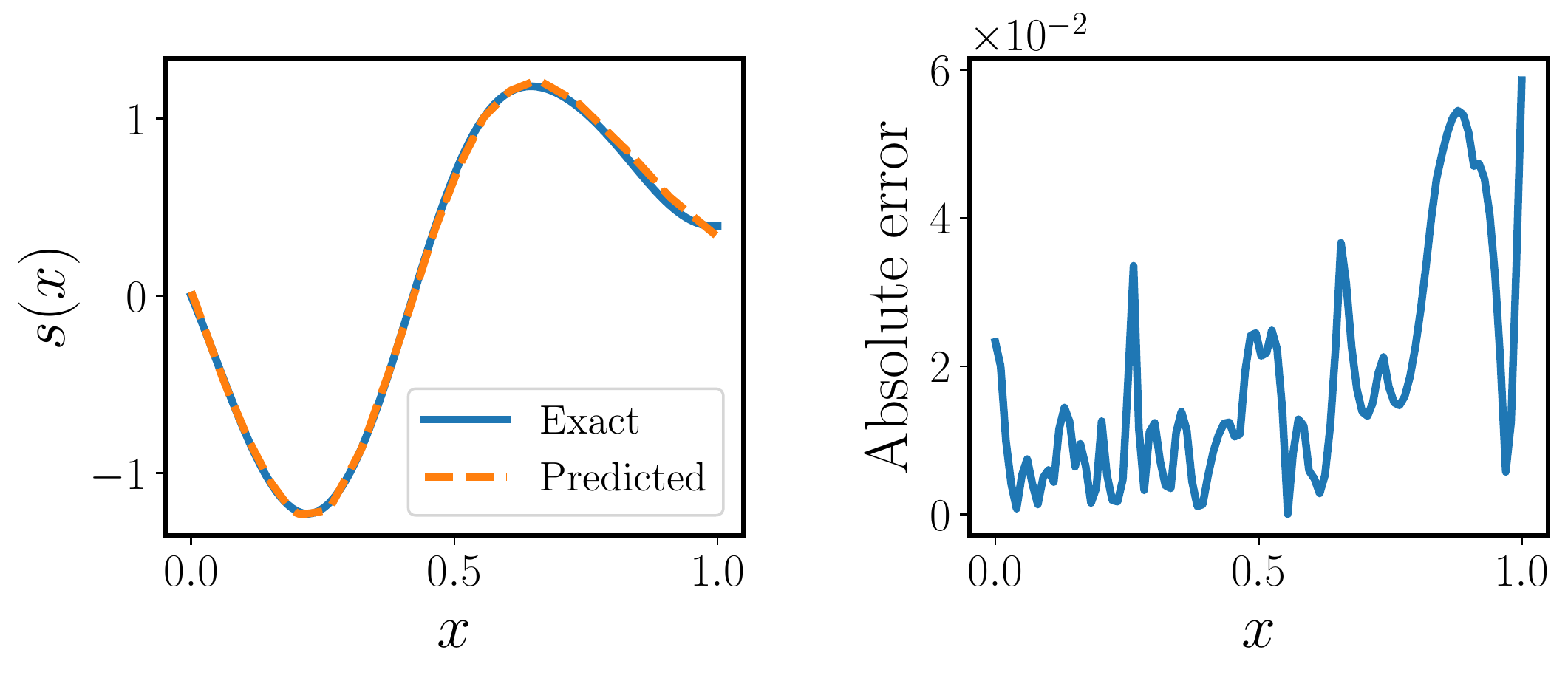}
         \label{fig: ODE_deeponet_pred_100}
     \end{subfigure}
        \caption{{\em Anti-derivative operator:} Exact solution versus the predicted solution of
       a trained physics-informed DeepONet for the same input sample with different output scale $k=0.01$ (left) and $k= 100$ (right). }
        \label{fig: ODE_deeponet_pred}
\end{figure}

We now proceed by approximating the anti-derivative operator with a DeepONet $G_{\bm{\theta}}$, where the branch and trunk networks are two 3-layer MLP networks with ReLU activation functions and $100$ neurons per hidden layer. 
We train this model by minimizing the loss function of equation (\ref{eq: deeponet_loss}) for $4 \times 10^4$ iterations of gradient descent using the Adam optimizer \cite{kingma2014adam}.  Figure \ref{fig: ODE_deeponet_pred} shows the predicted solutions for representative input samples with different output scales. One may observe that the model prediction corresponding to the large output scale is more accurate than the one corresponding to the small output scale.  To further validate this observation, we compute the relative $L^2$ errors over $N=10^3$ random input functions sampled from a GRF with a different fixed output scale $k$ and length scale $l=0.2$. The results are summarized in Table \ref{tab: ODE_deeponet_error}.
Intuitively, since the DeepONet is trained on a balanced training data-set, one would expect that the trained model is capable of achieving  approximately the same predictive accuracy across all input samples. However, this is contradicted with the observed positive correlation between the test error and the output scale $k$ reported in  Table \ref{tab: ODE_deeponet_error}.

\begin{figure}
    \centering
    \includegraphics[width=0.3\textwidth]{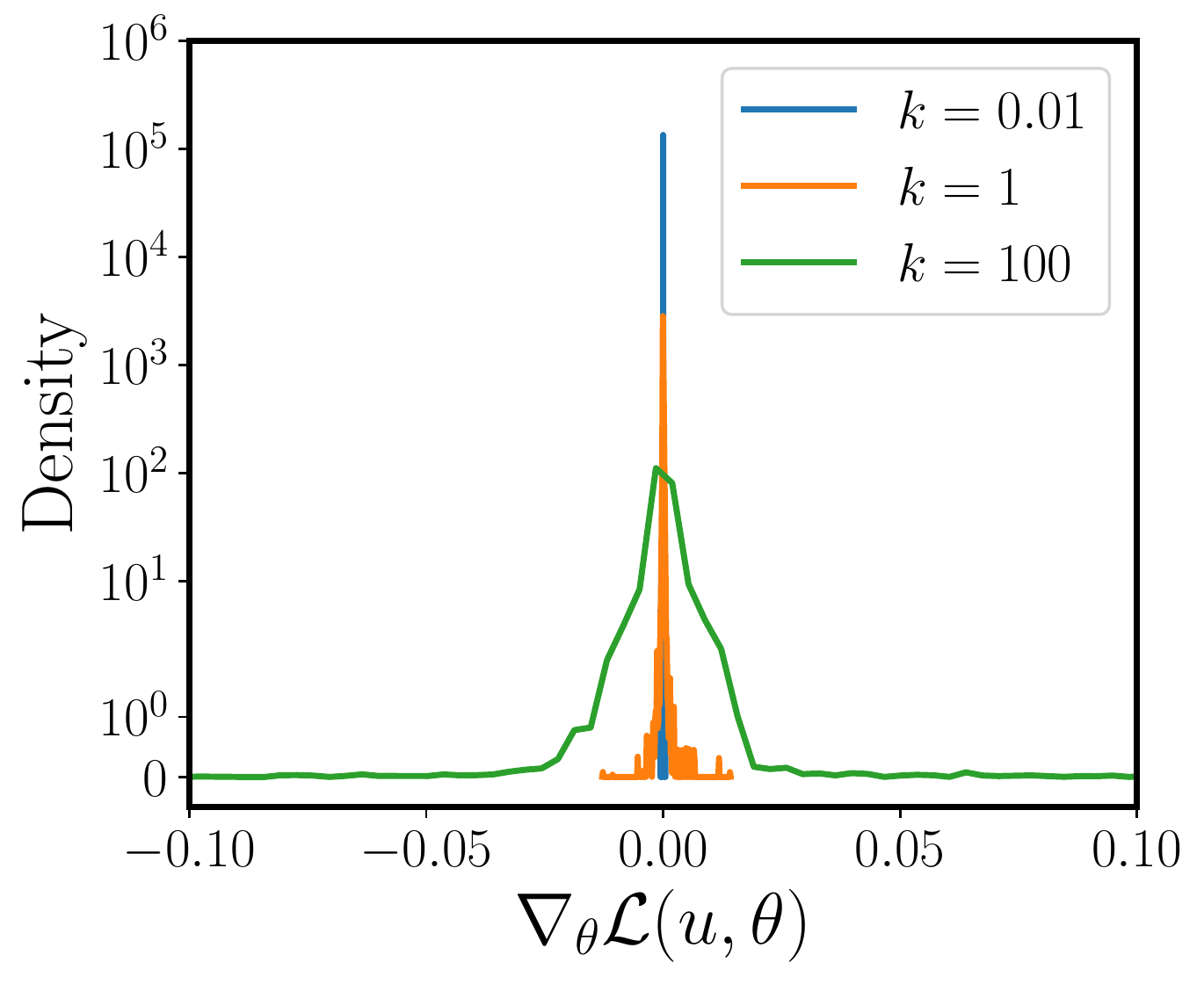}
    \caption{{\em Anti-derivative operator:} Histogram of back-propagated gradients corresponding to the input samples with different output scale $k \in [10^{-2}, 10^2]$.    }
    \label{fig: ODE_deeponet_grad}
\end{figure}

To investigate the inner mechanism that triggers this interesting behavior, we draw motivation from Wang {\em et. al.} \cite{wang2021understanding} and monitor the back-propagated gradients with respect to the network parameters during training. Rather than tracking the gradients of the aggregate loss, we monitor the gradients of each individual term in $\mathcal{L}(\bm{u}, \bm{\theta})$. In Figure \ref{fig: ODE_deeponet_grad}, we present a histogram of the gradients $\nabla_{\bm{\theta}} \mathcal{L}(\bm{u}, \bm{\theta})$ for three representative input functions $\bm{u}$ with different output scales, at the end of training. One key observation is that the gradients corresponding to small output scales are sharply concentrated at the origin and overall attain significantly smaller values than the gradients corresponding to large output scales.  In fact, if  we take a closer look at the definition of the loss function \ref{eq: deeponet_loss}, then (for $P=1$) the corresponding update rule of gradient descent is given by 
\begin{align}
    \frac{d \mathcal{L}}{d \bm{\theta}} = \frac{1}{N} \sum_{i=1}^N \frac{d \mathcal{L}(\bm{u}^{(i)}, \bm{\theta})}{d \bm{\theta}}  = \frac{2}{N} \sum_{i=1}^N (G_{\bm{\theta}}(\bm{u}^{(i)})(y_j^{(i)}) - s^{(i)}(y_j^{(i)})) \frac{d G_{\bm{\theta}}(\bm{u}^{(i)})(y^{(i)})}{d \bm{\theta}}.
\end{align}
From this we can deduce that for any input sample $\bm{u}$
\begin{align*}
    \|G_{\bm{\theta}}(\bm{u})- s\| \propto \|s\|,
    \propto \|u\|
\end{align*}
where $s(y) = s(0) + \int_0^y u(x) dx$ denotes the associated ODE solution. As a result, the solutions corresponding to large-value input functions yield large changes to the loss function and thus large magnitudes in the corresponding back-propagated gradients during training. For this specific example, it is no surprise that  minimizing the mean square error corresponding to large-value input samples dominates the total training process, and, consequently, our trained model is  biased towards yielding accurate solutions corresponding to input samples with large output scales. 
Therefore, it is natural to consider assigning some weight to each term of the loss function to balance the back-propagated gradients, i.e
\begin{align}
     \mathcal{L}(\bm{\theta}) = \frac{1}{N} \sum_{i=1}^N  \lambda_i \mathcal{L}(\bm{u}^{(i)}, \bm{\theta})
\end{align}
Here we must point out that this issue has been also noticed by Di Leoni {\em et. al.} \cite{di2021deeponet}, and the authors used $\lambda_i = \frac{1}{\|s^{(i))}\|_\infty}$ as weights to mitigate it. However, this solution is inapplicable to physics-informed DeepONets \cite{wang2021learning}, not only due to the absence of paired input-output solution measurements, but also because the interplay between the operator loss $\mathcal{L}_{\text{operator}}$ and the physics loss $\mathcal{L}_{\text{physics}}$ is not taken into consideration. As illustrated  in \cite{wang2021learning, wang2021long}, the latter generally plays an important role in the loss convergence and the predictive accuracy of a DeepONet model. In the following section, we will propose a novel algorithm that assigns adaptive weights to the loss function via the lens of NTK theory, which is capable of tackling all the observed issues in a unified manner.

\begin{table}[]
\renewcommand{\arraystretch}{1.4}
    \centering
    \begin{tabular}{|c|c|c|c|c|c|}
    \hline
     Output scale   &  $k = 0.01$ & $k = 0.1$ & $k = 1$ & $k=10$ & $k=100$ \\
      \hline
      Rel. $L^2$ error   &  $3.29\% \pm 3.01 \%$   & $1.73\% \pm 1.40 \%$  & $1.36\% \pm 1.09 \%$ & $1.27\% \pm 1.01 \%$  &  $1.17\% \pm 0.86\%$  \\
         \hline
    \end{tabular}
    \caption{{\em Anti-derivative operator:} Mean and standard deviation of the relative $L^2$ errors over the test data sampled from a GRF with different output scale $k \in [10^{-2}, 10^2]$.    }
    \label{tab: ODE_deeponet_error}
\end{table}

\subsection{NTK analysis of physics-informed DeepONets}
\label{sec: NTK_PI_deeponet}

In this section, we employ the NTK framework to analyze and understand the training dynamics of physics-informed DeepONets, in the specific context of learning the solution operator for a given parametric PDE system (see equations (\ref{eq: parametric_PDE}) - (\ref{eq: BC}) in section \ref{sec: deeponets}). Without loss of generality, here we will assume  $P= Q = R$ and rewrite the loss function of equation (\ref{eq: physics_informed_deeponet_loss}) by combining all the individual terms in one summation by rearranging notation and indices as
\begin{align}
    \label{eq: rearranged_loss}
    \mathcal{L}(\bm{\theta}) = \frac{2}{N^*} \sum_{k=1}^{N^*} \left| T^{(k)}(u^{(k)}(\bm{x}_k), G_{\bm{\theta}}(\bm{u}^{(k)})(\bm{y}_k))\right|^2
\end{align}
where $N^* = 2 NR$, and for every $k$, $T^{(k)}(\cdot, \cdot)$ can be the identity operator, differential operator $\mathcal{N}(\cdot, \cdot)$ or the boundary condition $\mathcal{B}(\cdot, \cdot)$. We will refer to a loss function written in the above equation  as a "fully-decoupled" loss. Similarly, we will refer to a loss function  written in the form of equation (\ref{eq: physics_informed_deeponet_loss}) as a "partially-decoupled" loss.

Now we generalize the original definition of the NTK to physics-informed DeepONets.
\begin{definition}
\label{def: deeponet_ntk}
Given the loss function \ref{eq: rearranged_loss},  the Neural Tangent Kernel of physics-informed DeepONets (NTK of PI-DeepONets) is  a matrix $\bm{H}(\bm{\theta}) \in \R^{N^* \times N^*}$ whose entries are defined by
\begin{align}
    \label{eq: def_deeponet_ntk}
    \bm{H}_{ij}(\bm{\theta}) =  \left\langle \frac{d T^{(i)} \left(u^{(i)} (\bm{x}_i), G_{\bm{\theta}} (\bm{u}^{(i)})(\bm{y}_i)  \right)  }{d \bm{\theta}}  , \frac{d T^{(j)} \left(u^{(j)} (\bm{x}_j), G_{\bm{\theta}} (\bm{u}^{(j)})(\bm{y}_j) \right)  }{d \bm{\theta}}                  \right\rangle, 
\end{align}
for $i,j = 1, 2, \dots, N^*$.
\end{definition}

\begin{lemma} The NTK of PI-DeepONets $\bm{H}(\bm{\theta})$ has the following properties
\label{prop: deeponet_NTK_property}
\begin{enumerate}[label=(\alph*)]
    \item $\bm{H}(\bm{\theta})$ is a positive semi-definite matrix.
    \item $\|\bm{H}(\bm{\theta})\|_\infty$ is achieved by some diagonal entry of $\bm{H}(\bm{\theta})$, i.e
    \begin{align}
        \|\bm{H}(\bm{\theta})\|_\infty = \max_{1\leq k \leq N^*} \bm{H}_{kk}(\bm{\theta}).
    \end{align}
\end{enumerate}
\end{lemma}
\begin{proof}
The proof is provided in Appendix \ref{proof: deeponet_NTK_property}.
\end{proof}

For simplicity, we also introduce the following notation.
\begin{definition} 
\label{def: notation}
For the loss function \ref{eq: rearranged_loss}, we define
\begin{enumerate}[label=(\alph*)]
    \item $\bm{U} = \left(\bm{u}^{(k)}\right)_{k=1}^{N^*}$  $\bm{X} = \left(\bm{x}_k \right)_{k=1}^{N^*}$,  $\bm{Y} = \left(\bm{y}_k \right)_{k=1}^{N^*}$.
    \item  $\bm{U}(\bm{X}) = \left( u^{(k)} (\bm{x}_k) \right)_{k=1}^{N^*}  $, $\bm{G}_{\bm{\theta}}(\bm{U})(\bm{Y}) = \left( G_{\bm{\theta}}  (\bm{u}^{(k)} ) (\bm{y}_k)\right)_{k=1}^{N^*}$.
    \item $ \bm{T}(\bm{U}(\bm{X}), G_{\bm{\theta}} (\bm{U}) (\bm{Y}) )) = \left(T^{(k)}(u^{(k)}(\bm{x}_k), G_{\bm{\theta}}(\bm{u}^{(k)})(\bm{y}_k))   \right)_{k=1}^{N^*}$. 
    \item $\mathcal{L}^{(k)}(\bm{\theta}) =   \left| T^{(k)}(u^{(k)}(\bm{x}_k), G_{\bm{\theta}}(\bm{u}^{(k)})(\bm{y}_k))\right|^2 $, then $\mathcal{L}(\bm{\theta}) = \frac{2}{N^*} \sum_{k=1}^{N^*}\mathcal{L}^{(k)}(\bm{\theta}) $.
\end{enumerate}
\end{definition}
With these definitions, we can prove the following lemma, which characterizes the training dynamics of physics-informed DeepONets.
\begin{lemma}
\label{lemma: ODE}
Given the loss function \ref{eq: rearranged_loss}, the physics-informed DeepONet outputs is governed by the following ODE: 
\begin{align}
    \label{eq: deeponet_training_dynamics}
    \frac{d \bm{T}\left(\bm{U}(\bm{X}), G_{\bm{\theta}(t)} (\bm{U}) (\bm{Y}) )\right)   }{ d t} = - \frac{4}{ N^*} \bm{H}(\bm{\theta}) \cdot \bm{T}(\bm{U}(\bm{X}), G_{\bm{\theta}(t)} (\bm{U}) (\bm{Y}) ))
\end{align}
where $\bm{H} \in \R^{N^* \times N^*}$ is the NTK of PI-DeepONets defined in equation \ref{eq: def_deeponet_ntk} 
\end{lemma}
\begin{proof}
The proof is provided in Appendix \ref{proof: ODE}.
\end{proof}

Although the recent work of \cite{fort2020deep,leclerc2020two} has empirically observed that the NTK of finite-width MLP networks evolves at a constant slow velocity after a rapid change at the early training stages, the behavior of the NTK of physics-informed DeepONets is unclear during training, and worth of further investigation in the future. Nevertheless, performing a similar analysis as in Section \ref{sec: NTK} we can show that 
\begin{align}
    \bm{T}\left(\bm{U}(\bm{X}), G_{\bm{\theta}(t)} (\bm{U}) (\bm{Y}) )\right) \approx e^{-\bm{H} t} \bm{T}\left(\bm{U}(\bm{X}), G_{\bm{\theta}(0)} (\bm{U}) (\bm{Y}) )\right).
\end{align}
Letting $\bm{H} = \bm{Q \Lambda} Q^T$ be an orthogonal decomposition of $\bm{H}$, where $\bm{\Lambda}$ is a diagonal matrix  whose entries are the eigenvalues of $\bm{H}$, we obtain
\begin{align}
    \bm{Q}^T \bm{T}\left(\bm{U}(\bm{X}), G_{\bm{\theta}(t)} (\bm{U}) (\bm{Y}) )\right) \approx e^{-\bm{\Lambda t}} \bm{Q}^T \bm{T}\left(\bm{U}(\bm{X}), G_{\bm{\theta}(0)} (\bm{U}) (\bm{Y}) )\right). 
\end{align}
Consequently,  the eigenvalues of $\bm{H}(\bm{\theta})$ not only determine the stiffness of the ODE system \ref{eq: deeponet_training_dynamics}, but also characterizes the convergence rate of total training error. 

\subsection{NTK-guided weights for physics-informed DeepONets}

\label{sec: NTK-guided weights}
Recall that our goal is to assign appropriate weights to each individual term in the loss function of physics-informed DeepONets to balance the back-propagated gradients, as well as calibrate the convergence rate of each term. To this end, we consider a weighted loss
\begin{align}
    \label{eq: weighted_loss}
      \mathcal{L}(\bm{\theta}) = \frac{2}{N^*} \sum_{k=1}^{N^*} \lambda_k \mathcal{L}^{(k)}(\bm{\theta}) =  \frac{2}{N^*} \sum_{k=1}^{N^*} \lambda_k \left| T^{(k)}(u^{(k)}(\bm{x}_k), G_{\bm{\theta}}(\bm{u}^{(k)})(\bm{y}_k))\right|^2.
\end{align}

\begin{algorithm}
\SetAlgoLined
Consider a physics-informed DeepONet $G_{\bm{\theta}}$ with parameters $\bm{\theta}$
and the corresponding weighted loss
\begin{align}
      \mathcal{L}(\bm{\theta}) = \frac{2}{N^*} \sum_{k=1}^{N^*} \lambda_k \left| T^{(k)}(u^{(k)}(\bm{x}_k), G_{\bm{\theta}}(\bm{u}^{(k)})(\bm{y}_k))\right|^2,
\end{align}
where all weights $\{\lambda_k\}_{k=1}^{N^*}$ are initialized to $1$. Then, use $S$ steps of a gradient descent algorithm to update the parameters $\bm{\theta}$ as:
 \For{$n = 1, \dots, S$}{
  (a) Compute and update $\lambda_k$ by
  \begin{align}
        \label{eq: lambda_update}
    \lambda_k =  \left( \frac{ \|\bm{H}(\bm{\theta})\|_\infty }{ \bm{H}_{kk}(\bm{\theta}_n)} \right)^\alpha = \left( \frac{\max_{1 \leq k \leq N^*} \bm{H}_{kk} (\bm{\theta}_n) }{ \bm{H}_{kk}(\bm{\theta}_n)} \right)^\alpha
  \end{align}
  where $\bm{\theta}_n$ denotes the DeepONet parameters at $n$-th step, and $\bm{H}_{kk}$ is the $k$-th diagonal entry of the NTK  matrix defined in Definition \ref{def: deeponet_ntk}. Here $\alpha\in[0,1]$ is a user-defined hyper-parameter that determines the magnitude of each weight.
  
  (b) Update the parameters $\bm{\theta}$ via gradient descent
  \begin{align}
      \bm{\theta}_{n+1} = \bm{\theta}_{n} - \eta \nabla_{\bm{\theta}}\mathcal{L}(\bm{\theta}_n)
  \end{align}
  The recommended hyper-parameter values are: $\alpha = 1 $ or $\alpha = \frac{1}{2}$.
  % \alpha = [0, 1]
}
\caption{NTK-guided weighting scheme for training deep operator networks}
\label{alg: NTK_weights}
\end{algorithm}

Motivated by Wang {\em et. al.} \cite{wang2020and}, we propose Algorithm \ref{alg: NTK_weights} to calibrate the interplay between each individual term of the loss function. 
One may note that Algorithm \ref{alg: NTK_weights} reduces to training a conventional physics-informed DeepONet  when $\alpha=0$. However, in this work we will mainly consider the case of $\alpha=\frac{1}{2}$, and $\alpha=1$.   In the rest of this work, we may refer to the NTK-guided weights with $\alpha=1$ as NTK weights while referring the NTK-guided weights with $\alpha=\frac{1}{2}$ as moderate NTK weights.

To get a deeper understanding of this algorithm and the role of the hyper-parameter $\alpha$, let us boldly assume that each individual term of the loss function \ref{eq: rearranged_loss} does not affect one another, and is minimized independently via gradient descent. This simplifying assumption enables us to decouple the total loss into individual terms, thus enabling us to analyze their training dynamics separately.  For every $1 \leq k \leq N^*$, we then obtain
\begin{align}
    \frac{d T^{(k)}(u^{(k)}(\bm{x}_k), G_{\bm{\theta}}(\bm{u}^{(k)})(\bm{y}_k)) }{d t} &= - \frac{4 \lambda_k}{N^*}  \left\|  \frac{d T^{(k)}(u^{(k)}(\bm{x}_k), G_{\bm{\theta}}(\bm{u}^{(k)})(\bm{y}_k)) }{d \bm{\theta}}  \right\|_2^2  \cdot \left(T^{(k)}(u^{(k)}(\bm{x}_k), G_{\bm{\theta}}(\bm{u}^{(k)})(\bm{y}_k))   \right) \\
    & =  - \frac{4}{N^*}  \lambda_k \bm{H}_{kk}(\bm{\theta})  \cdot \left(T^{(k)}(u^{(k)}(\bm{x}_k), G_{\bm{\theta}}(\bm{u}^{(k)})(\bm{y}_k))   \right)
\end{align}
This implies that (under our independence assumption) $ \lambda_k \bm{H}_{kk}(\bm{\theta})$ essentially quantifies the convergence rate of every weighted individual loss. Then, taking $\alpha =1$ in equation \ref{eq: lambda_update} yields
\begin{align}
    \frac{d T^{(k)}(u^{(k)}(\bm{x}_k), G_{\bm{\theta}}(\bm{u}^{(k)})(\bm{y}_k)) }{d t} = - \frac{4}{N^*}  \|\bm{H}(\bm{\theta})\|_\infty  \cdot \left(T^{(k)}(u^{(k)}(\bm{x}_k), G_{\bm{\theta}}(\bm{u}^{(k)})(\bm{y}_k))   \right), \ \ 1 \leq k \leq N^*.
\end{align}
Consequently, computing the NTK weights by setting $\alpha=1$ enables each individual loss to achieve the same maximum convergence rate  $\| \bm{H}(\bm{\theta})\|_\infty$.

Next, let us examine the back-propagated gradients of each individual weighted term when applying gradient descent updates. For $k=1,2, \dots, N^*$,  we have
\begin{align}
     \left\| \nabla_{\bm{\theta} } \lambda_k \mathcal{L}^{(k)} (\bm{\theta}) \right\|_2 &= 2 \frac{ \|\bm{H}(\bm{\theta})\|_\infty }{ \bm{H}_{kk}(\bm{\theta}_n)} \left\|  \left( T^{(k)}(u^{(k)}(\bm{x}_k), G_{\bm{\theta}}(\bm{u}^{(k)})(\bm{y}_k)) \right) \nabla_{\bm{\theta}} T^{(k)}(u^{(k)}(\bm{x}_k), G_{\bm{\theta}}(\bm{u}^{(k)})(\bm{y}_k))\right\|_2  \\
    &= 2\frac{\| \bm{H}(\bm{\theta})\|_\infty }{ \left\|  \nabla_{\bm{\theta}} T^{(k)}(u^{(k)}(\bm{x}_k), G_{\bm{\theta}}(\bm{u}^{(k)})(\bm{y}_k))\right\|_2 } \left| T^{(k)}(u^{(k)}(\bm{x}_k), G_{\bm{\theta}}(\bm{u}^{(k)})(\bm{y}_k)) \right|.
\end{align}
As a result, the NTK weights obtained by setting $\alpha=1$ tend to penalize the gradients with large magnitude and to amplify the gradients with small magnitude. However, in practice we observe that this update rule sometimes yields extremely large weights , potentially due to the presence of vanishing gradients, which can lead to an unstable training process.  In contrast, updating the NTK weights by setting $\alpha=\frac{1}{2}$ gives
\begin{align}
    \left\| \nabla_{\bm{\theta} } \lambda_k \mathcal{L}^{(k)} (\bm{\theta}) \right\|_2 &= 2 \sqrt{\frac{ \|\bm{H}(\bm{\theta})\|_\infty }{ \bm{H}_{kk}(\bm{\theta}_n)}}  \left\|  \left( T^{(k)}(u^{(k)}(\bm{x}_k), G_{\bm{\theta}}(\bm{u}^{(k)})(\bm{y}_k)) \right) \nabla_{\bm{\theta}} T^{(k)}(u^{(k)}(\bm{x}_k), G_{\bm{\theta}}(\bm{u}^{(k)})(\bm{y}_k))\right\|_2  \\
    &= 2\sqrt{\| \bm{H}(\bm{\theta})\|_\infty } \left| T^{(k)}(u^{(k)}(\bm{x}_k), G_{\bm{\theta}}(\bm{u}^{(k)})(\bm{y}_k)) \right|.
\end{align}
This implies that NTK weights computed with $\alpha=\frac{1}{2}$  normalize the loss function such that the gradients of each weighted individual loss are of the same magnitude $\sqrt{\| \bm{H}(\bm{\theta})\|_\infty }$.

In summary, Algorithm \ref{alg: NTK_weights} with $\alpha=1$  aims to balance the convergence rate derived from the associated  gradient flow, while Algorithm \ref{alg: NTK_weights}  with $\alpha=\frac{1}{2}$  aims to balance the back-propagated gradients.
Rather than constructing the whole NTK matrix, it is worth emphasizing that the Algorithm \ref{alg: NTK_weights} only requires one to compute the NTK diagonal entries, which will greatly save computational costs and reduce the total training time in practice. 

\begin{figure}
    \centering
    \includegraphics[width=0.7\textwidth]{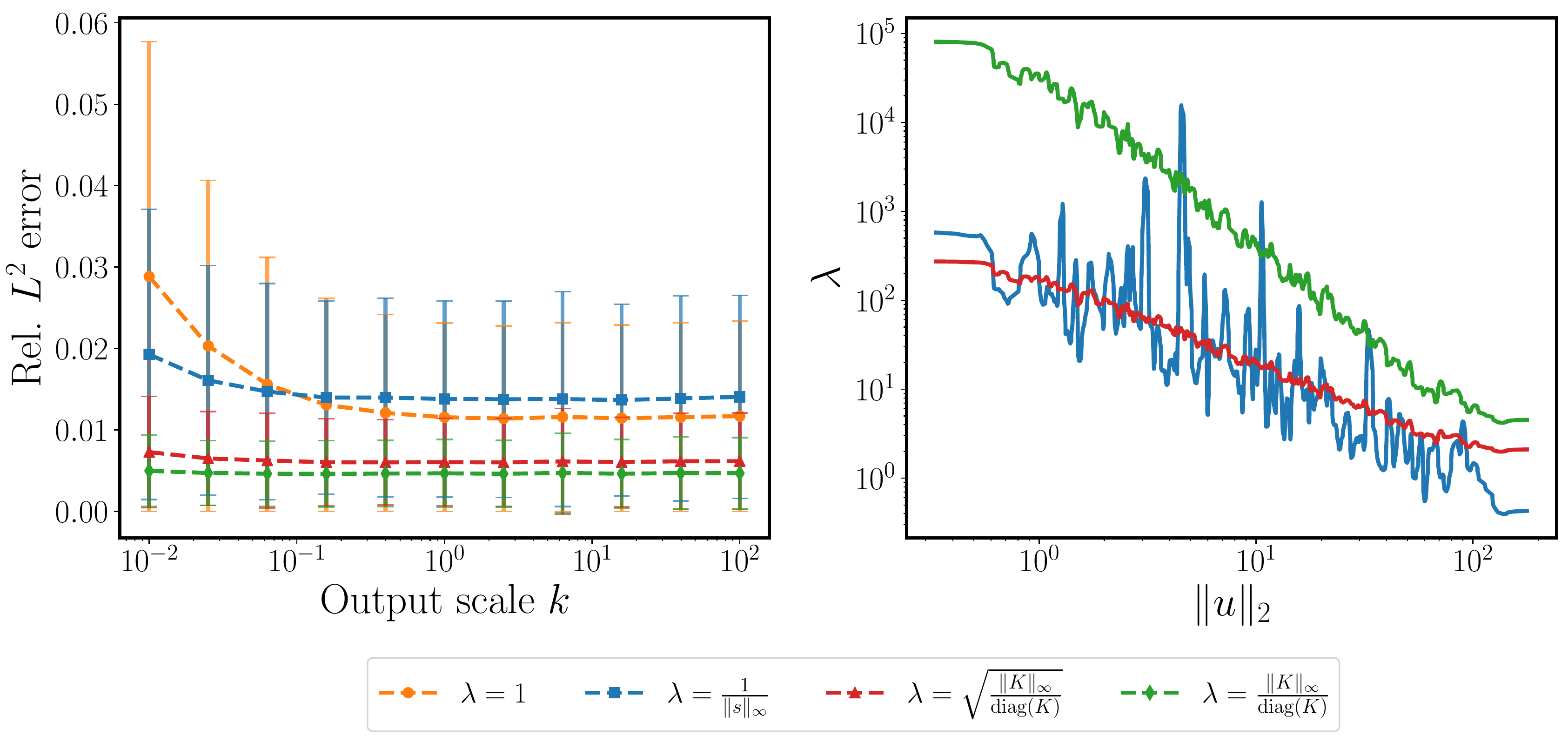}
    \caption{{\em Anti-derivative operator:} {\em Left:} Mean and standard deviation of the relative $L^2$ prediction errors of the trained DeepONets using different weighting schemes, averaged over the test data sampled from a GRF with different output scale $k \in [10^{-2}, 10^2]$.  A detailed description of the legend is summarized in Table \ref{tab: weighting_schemes}. {\em Right:} Weight distributions of trained models with different weighting schemes  in increasing order of $\|u\|_2$. }
    \label{fig: ODE_deeponet_error}
\end{figure}

To compare the performance of different weight schemes mentioned in Section \ref{sec: grad_pathologies} - \ref{sec: NTK_PI_deeponet}, let us first revisit the example of learning anti-derivative operator shown in Section \ref{sec: grad_pathologies}. In all cases we will employ the same DeepONet architecture to represent the solution operator $G$ (see Appendix \ref{app: hp_settings} for additional details). The DeepONet model can be trained by minimizing the weighted loss function with different weighting schemes under exactly the same hyper-parameter settings (i.e. \# of iterations, learning rate, batch size, etc.). The left panel of Figure \ref{fig: ODE_deeponet_error} summarizes the averaged resulting relative $L^2$ prediction errors along with their standard deviation corresponding to input functions with different output scales in the test data-set. Among all weighting schemes, NTK weights with $\alpha=1$ yields the best predictive accuracy as well as the smallest standard deviation, which indicates that the trained DeepONet is unbiased towards different output-scale input samples. 
Moreover, using $\lambda= \frac{1}{\|s\|_\infty}$ as weights slightly mitigates the gradient pathology issue and achieves a similar performance to the original un-weighted loss. To explore the underlying reasons, we investigate the weight distribution of different update rules after training. Particularly, we compute the weights for each input sample $\bm{u}$ in a random mini-batch and visualize them in a descending order of $\|\bm{u}\|_2$. As shown in the right panel of Figure \ref{fig: ODE_deeponet_error}, all weight distributions are inversely proportional to the magnitude of input samples, which potentially balance the back-propagated gradients of each individual term in the loss function. However, one may note that the weights for $\lambda = \frac{1}{\|s\|_\infty}$  exhibit high variance. This is because of the inaccurate estimate of $\|s\|_\infty$ since only one corresponding observation is provided for each input sample. We believe that such inaccurate estimation of $\|s\|_\infty$ leads to a poor model performance.

\subsection{An improved DeepONet architecture}

\label{sec: modified_deeponet}
Besides effective training algorithms, the network architecture itself also plays a significant role in deep learning. In this section, we propose a novel DeepONet architecture, which is empirically proved to be uniformly better than the conventional DeepONet architecture of Lu {\it et al.} \cite{lu2021learning}. To demonstrate our motivation, we start with a perspective of deep information propagation \cite{schoenholz2016deep}, which hypothesizes that a necessary condition for a random neural network to be trainable is that information should be able to pass through it in a stable manner. One may notice that the conventional DeepONet architecture merges the latent representations of DeepONet inputs $\bm{u}$ and $\bm{y}$ only in the last layer via an inner product. Consequently, the final information fusion may be inefficient if the DeepONet input signals fail to propagate through a deep branch network or trunk network at initialization, leading to an ineffective training process and poor model performance. To design a network architecture that is more resilient to vanishing signals,  we draw motivation from Wang {\em et. al.} \cite{wang2021understanding} and modify the forward pass of an L-layer  DeepONet as follows
\begin{align}
     & \bm{U} = \phi( \bm{W}_u \bm{u} + \bm{b}_u), \ \  \bm{V} = \phi( \bm{W}_y \bm{y} + \bm{b}_y) \\
     & \bm{H}_u^{(1)} = \phi(\bm{W}^{(1)}_u \bm{u}  + \bm{b}^{(1)}_u), \ \ \bm{H}_y^{(1)} = \phi( \bm{W}^{(1)}_y \bm{y} + \bm{b}^{(1)}_y) \\
    & \bm{Z}_u^{(l)} = \phi(\bm{W}^{(l)}_u \bm{H}^{(l)}_u  + \bm{b}^{(l)}_u), \ \ \bm{Z}_y^{(l)} = \phi( \bm{W}^{(l)}_y \bm{H}^{(l)}_y + \bm{b}^{(l)}_y), \quad l = 1, 2, \dots, L-1 \\
    \label{eq: arch_embed_1}
    & \bm{H}^{(l+1)}_u = (1 - \bm{Z}^{(l)}_u) \odot \bm{U}  +  \bm{Z}^{(l)}_u  \odot \bm{V}, \quad l = 1, \dots, L-1 \\
     \label{eq: arch_embed_2}
    & \bm{H}^{(l+1)}_y = (1 - \bm{Z}^{(l)}_y) \odot \bm{U}  +  \bm{Z}^{(l)}_y  \odot \bm{V}, \quad l = 1, \dots, L-1 \\
    & \bm{H}_u^{(L)} = \phi(\bm{W}^{(L)}_u \bm{H}^{(L-1)}_u  + \bm{b}^{(L)}_u), \ \ \bm{H}_y^{(L)} = \phi( \bm{W}^{(L)}_y \bm{H}^{(L-1)}_y   + \bm{b}^{(L)}_y) \\
    &G_{\bm{\theta}}(\bm{u})(\bm{y}) = \left\langle \bm{H}_u^{(L)}, \bm{H}_y^{(L)}        \right\rangle,
\end{align}
where $\odot$ denotes point-wise multiplication, $\phi$ denotes a activation function, and  $\bm{\theta}$ represents all  trainable parameters of the DeepONet model. In particular, $\{\bm{W}_u^{(l)},  \bm{b}_u^{(l+1)} \}_{l=1}^{L+1}$ and $ \{\bm{W}_y^{(l)},  \bm{b}_y^{(l+1)} \}_{l=1}^{L+1} $ are the weights and biases of the branch and trunk networks, respectively. 

\begin{figure}
    \centering
    \includegraphics[width=0.8\textwidth]{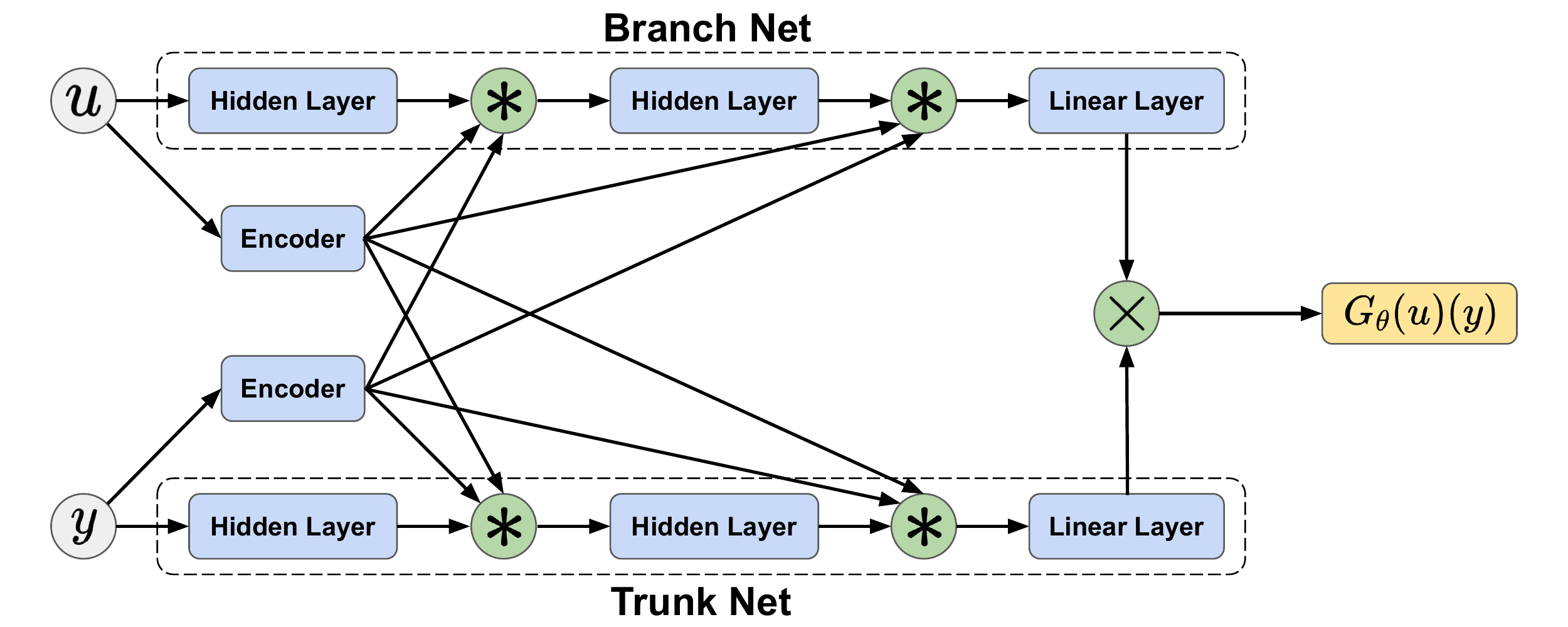}
    \caption{{\em Modified DeepONet architecture:} Two encoders are applied to the input samples and the input coordinates, respectively. The embedded features are then inserted into each hidden layer of the branch network and the trunk network using a $\mathbf{*}$ operation, which is defined in  equation (\ref{eq: arch_embed_1}) - (\ref{eq: arch_embed_2}). The final outputs is obtained by the same way as conventional DeepONets, i.e,  merging the outputs of the branch and trunk networks via a dot production.  }
    \label{fig: modified_deeponet}
\end{figure}

At first glance, this novel architecture seems to appear a bit complicated. However, notice that the proposed architecture is almost the same as a standard DeepONet model using MLPs as backbone, with the addition of two encoders and a minor modification in the forward pass. As illustrated in Figure \ref{fig: modified_deeponet}, we embed the DeepONet inputs $\bm{u}$ and $\bm{y}$ into a high-dimensional feature space via two encoders $\bm{U}, \bm{V}$, respectively. Instead of just merging the propagated information in the output layer of the branch and trunk networks, we merge the embeddings $\bm{U}, \bm{V}$ in each hidden layer of these two sub-networks using  a point-wise multiplication (equation (\ref{eq: arch_embed_1}) - (\ref{eq: arch_embed_2})). Heuristically, this design may not only help input signals propagate through the DeepONet, but also enhance its capability of representing non-linearity due to the extensive use of point-wise multiplications. In section \ref{sec: results}, we will demonstrate that this simple modification uniformly leads to more accurate predictions than the original DeepONet architecture across all benchmarks considered in this work.

\section{Results}
\label{sec: results}

To validate the proposed training algorithms and network architectures, we present a series of comprehensive numerical studies across a range of representative benchmarks for which the conventional physics-informed DeepONets struggle.
More specifically, we quantify the predictive accuracy of trained physics-informed DeepONets using the different weighting schemes listed in Table \ref{tab: weighting_schemes}, as well as the different DeepONet architectures listed in Table \ref{tab: archs}. 

The error metric employed throughout all numerical experiments to assess model performance is the relative $L^2$ norm. Specifically, the reported test errors correspond to the mean of the relative $L^2$ error of a trained model over all examples in the test data-set, i.e
\begin{align}\label{eq: test_error}
    \text{Test error} := \frac{1}{N} \sum_{i=1}^N \frac{\|G_{\theta}(\bm{u}^{(i)})(y) - G(\bm{u}^{(i)})(y)\|_2 }{\|G(\bm{u}^{(i)})(y)\|_2},
\end{align}
where $N$ denotes the number of examples in the test data-set and $y$ is typically a set of equi-spaced points in the domain of $G(u)$. Here $G_{\theta}(\bm{u}^{(i)})(\cdot)$ denotes a predicted DeepONet output function, while $G(\bm{u}^{(i)})(\cdot)$ corresponds to the ground truth target functions.

Throughout all benchmarks we  employ hyperbolic tangent activation functions (Tanh) and initialize the DeepOnet networks using the Glorot normal scheme \cite{glorot2010understanding}, unless otherwise stated. All networks are trained via  mini-batch stochastic gradient descent using the Adam optimizer \cite{kingma2014adam} with default settings.  Particularly, we set the batch size to be 10,000 and use exponential learning rate decay with a decay-rate of 0.9 every 2, 000 training iterations. 
The detailed hyper-parameters and computational cost for all examples are listed in Appendix \ref{app: hp_settings} and \ref{sec: computational_cost}. We can see that the proposed methods generally incur a higher computational costs than the original DeepONet formulation. For example, training physics-informed DeepONets with NTK weights is $\sim$3-5x slower than without any weighting schemes, while employing the modified DeepONet architecture takes about twice the training time as the conventional DeepONet architecture. The code and data accompanying this manuscript will be made publicly available at \url{https://github.com/PredictiveIntelligenceLab/ImprovedDeepONets}.

% \subsection{Diffusion reaction}

\begin{table}[]
\renewcommand{\arraystretch}{2.2}
    \centering
    \begin{tabular}{|c|c|c|c|} 
    \hline
        Weighting scheme & Notation &  Loss form  & Update rule \\
        \hline
        No weights &  $\lambda = 1$ &  Fully-decoupled   &  $\lambda_k = 1$\\
         \hline
     Fixed weights &  $\lambda_{\text{bc}} = \lambda$ &  Partially-decoupled   & $\lambda_{\text{bc}} = \lambda, \lambda_{\text{r}} = 1$   \\
         \hline
    Data-guided weights &  $\lambda = \frac{1}{\|s\|_\infty}$ &  Fully-decoupled   &  $\lambda_k = \frac{1}{\|s^{(k)}\|_\infty}$ \\
         \hline 
    NTK-guided weights &  $\lambda =\left( \frac{\|\bm{H}\|_\infty}{\text{diag}(\bm{H})} \right)^\alpha$ &  Fully-decoupled   &  $\lambda_k = \left( \frac{\|\bm{H}\|_\infty}{\text{diag}(\bm{H}_{kk})} \right)^\alpha$ \\
         \hline 
    \end{tabular}
    \caption{Weighting schemes and their associated notation, as considered in the numerical studies presented in this work.}
    \label{tab: weighting_schemes}
\end{table}

\begin{table}[]
\renewcommand{\arraystretch}{1.8}
    \centering
    \begin{tabular}{|c|l|}
    \hline
       DeepONet architecture &  Description  \\
        \hline
      DeepONet with MLP &  The branch and trunk networks are MLPs\\
        \hline
     DeepONet with  modified MLP & The branch and trunk networks are modified MLPs, see \cite{wang2021understanding}. \\
              \hline
     Modified DeepONet & The proposed novel DeepONet architecture in section \ref{sec: modified_deeponet}.\\
        \hline
    \end{tabular}
    \caption{DeepONet architectures considered in the numerical studies presented in this work. We remark that DeepONet with MLP refers to a  conventional DeepONet model,  while DeepONet with modified MLP consists of two independent sub-networks employing modified MLP networks as backbone that does not share any parameters. One main difference between the modified DeepONet and conventional DeepONet architectures is that two encoders are employed to exchange  information between the branch and the trunk networks before merging them together. To simplify the notations, we may omit “DeepONet” in the rest of this work.}
    \label{tab: archs}
\end{table}

\subsection{Advection equation}

% systematical study for fixed weights using different architectures

We start with a one-dimensional linear advection equation with variable coefficients of the form   
\begin{align}
     \frac{\partial s}{\partial t} + u(x) \frac{\partial s}{\partial x} = 0, \quad (x, t) \in (0,1) \times (0,1),
\end{align}
subject to the initial and  boundary  condition
\begin{align}
    &s(x, 0) = f(x), \quad x \in [0, 1],\\
    &s(0, t) = g(t), \quad t \in [0, 1]
\end{align}
where $f(x) = \sin(\pi x)$ and $g(t) = \sin(\frac{\pi}{2} t)$. To ensure numerical stability, we make the input function $u(x)$  strictly positive by letting $u(x) = v(x) - \min_{x}v(x) + 1$  where $v(x)$ is sampled from a GRF with a length scale $l=0.2$.  Our goal is to learn the solution operator $G$ mapping the variable coefficient $u(x)$ to the associated spatio-temporal PDE solution $s(x, t)$.

We represent the solution map by a DeepONet $G_{\bm{\theta}}$ and define PDE residual as 
\begin{align}
    \mathcal{R}_{\bm{\theta}}(\bm{u})(x,t) = \frac{\partial G_{\bm{\theta}}(\bm{u})(x, t)}{\partial t} - u(x) \frac{\partial G_{\bm{\theta}}(\bm{u})(x, t)}{\partial x}. 
\end{align}
Then, we formulate a weighted physics-informed as follows
\begin{align}
    \mathcal{L}(\bm{\theta}) &= \lambda_{\text{bc}} \mathcal{L}_{\text{BC}}(\bm{\theta}) + \lambda_{\text{ic}} \mathcal{L}_{\text{IC}}(\bm{\theta}) + \lambda_{\text{r}} \mathcal{L}_{\text{PDE}}(\bm{\theta}), 
\end{align}
where
\begin{align}
  & \mathcal{L}_{\mathrm{IC}}(\bm{\theta})=\frac{1}{N P} \sum_{i=1}^{N} \sum_{j=1}^{P}\left|G_{\bm{\theta}}(\bm{u}^{(i)})(x_{i c, j}^{(i)}, 0) - f(x_{i c, j}^{(i)}) \right|^{2}, \\
   &\mathcal{L}_{\mathrm{BC}}(\bm{\theta})=\frac{1}{N P} \sum_{i=1}^{N} \sum_{j=1}^{P}\left|G_{\bm{\theta}}(\bm{u}^{(i)})(0, t_{b c, j}^{(i)})- g( t_{b c, j}^{(i)})\right|^{2}, \\
   &\mathcal{L}_{\text{PDE}}(\bm{\theta}) = \frac{1}{N Q} \sum_{i=1}^N \sum_{j=1}^Q \left|  \mathcal{R}_{\bm{\theta}}(\bm{u}^{(i)})(x_{r,j}^{(i)}, t_{r,j}^{(i)})  \right|^2,
\end{align}
and $\lambda_{\text{bc}}, \lambda_{\text{ic}}, \lambda_{\text{r}}$ can be set to $1$ or some other user-specified value, and remain fixed during training. For using different weighting schemes, we reformulate the loss function into a fully-decoupled form
\begin{align}
        \mathcal{L}(\bm{\theta}) = \frac{1}{N^*} \sum_{k=1}^{N^*} \lambda_k \left| T^{(k)}(u^{(k)}(x_k), G_{\bm{\theta}}(\bm{u}^{(k)})(x_k, t_k))\right|^2,
\end{align}
where $T^{(k)}$ is a operator corresponding to either boundary, initial conditions or differential operators (e.g $\mathcal{R}_{\bm{\theta}}$), and $\lambda_k$ are weights that will be automatically updated by Algorithm \ref{alg: NTK_weights} during training.  

Here, we take $N=2,000$, $P=200$ and $Q=2,500$. The input samples $\bm{u}^{(i)}$ are evaluated at equi-spaced points $\{x_i\}_{i=1}^m$ in $[0,1]$. To generate a set of test data, we sample $N=100$ input functions from the same GRF and solve the advection equation using the Lax–Wendroff scheme \cite{iserles2009first} on a $100 \times 100$ uniform grid. 

\begin{figure}
     \centering
     \begin{subfigure}[b]{0.8\textwidth}
         \centering
         \includegraphics[width=\textwidth]{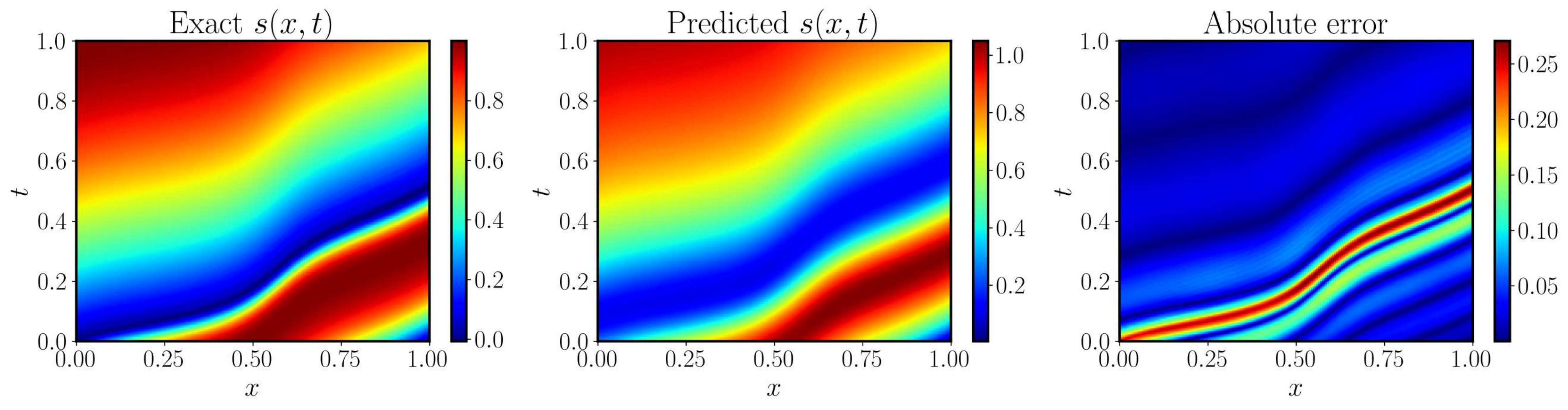}
     \end{subfigure}
     \begin{subfigure}[b]{0.8\textwidth}
         \centering
         \includegraphics[width=\textwidth]{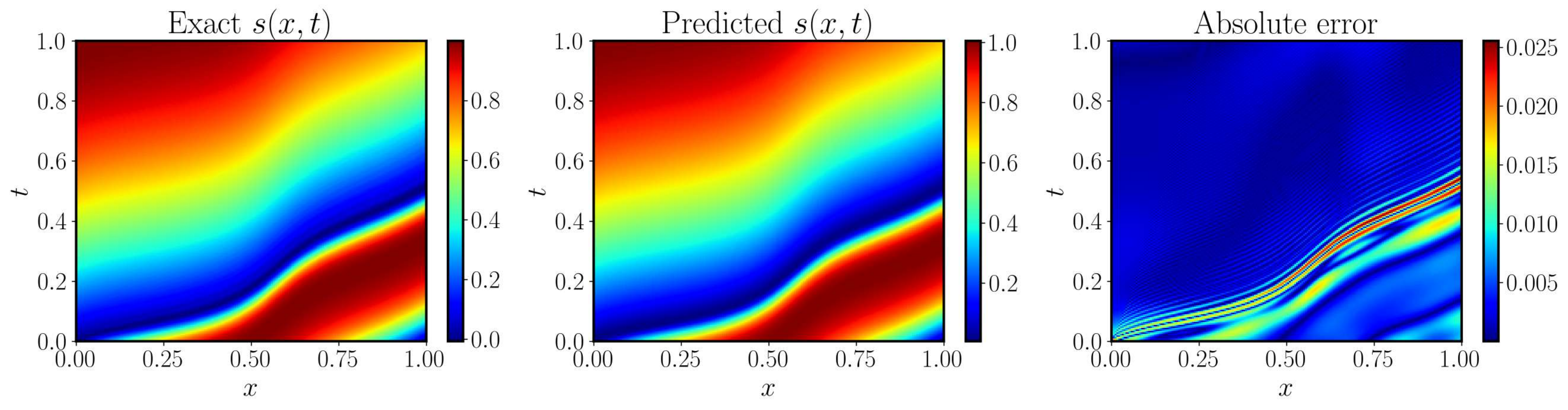}
     \end{subfigure}
        \caption{{\em Advection equation:} {\em Top:}  Exact solution versus the prediction of a trained conventional physics-informed DeepONet for a representative example in the test data-set. The resulting relative $L^2$ error is $9.32\%$.  {\em Bottom:}  Exact solution versus the prediction of a   physics-informed DeepONet represented by modified DeepONet architecture and trained  using Algorithm \ref{alg: NTK_weights} with $\alpha = \frac{1}{2}$ for the same example. The resulting relative $L^2$ error is $0.73\%$, which is 10x more accurate than the original formulation. }
        \label{fig: ADV_PI_deeponet_pred}
\end{figure}

We train the physics-informed DeepONet using different weighting schemes and DeepONet architectures  for $3 \times 10^5$ iterations,  and report the resulting test error in Table \ref{tab: adv_l2_error}.  
Despite some improvements brought by all weighting schemes, the results obtained with both the MLP and modified MLP architectures are unsatisfactory. In contrast, the modified DeepONet architecture with moderate NTK weights ($\alpha = 1/2$) achieves the best accuracy, which has been improved by a factor of x6 compared to the baseline  physics-informed DeepONets \cite{wang2021learning}.
A more detailed visual assessment of this comparison is shown in Figure \ref{fig: ADV_PI_deeponet_pred}. In contrast to the inaccurate approximation of sharp gradients  by  the conventional physics-informed DeepONet, the result of the improved one shows excellent agreement between the predictions and the numerical estimations with a relative $L^2$ error of $0.73\%$. 

\begin{figure}
    \centering
    \includegraphics[width=0.3\textwidth]{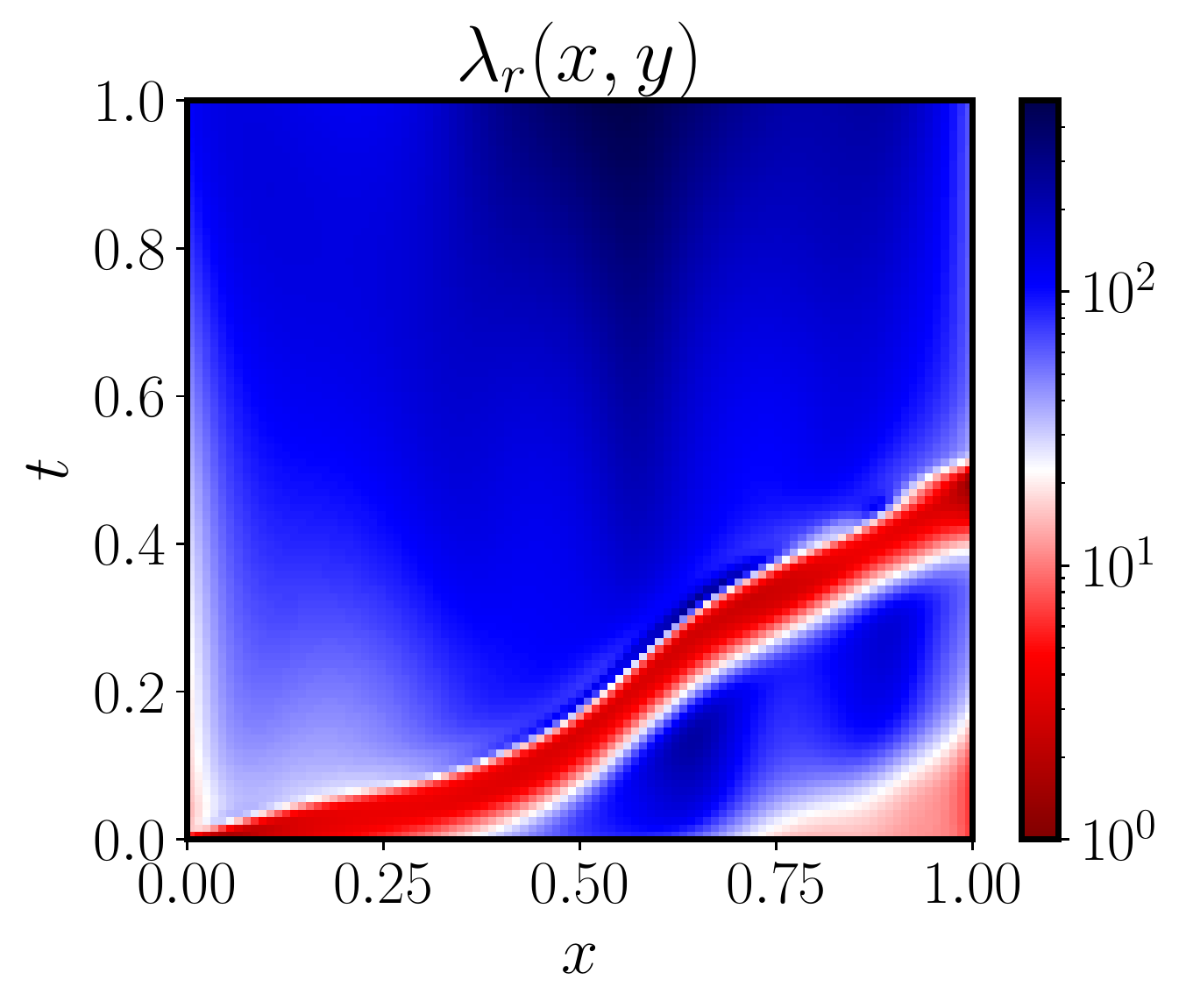}
    \caption{{\em Advection equation:} Weight distribution defined in equation (\ref{eq: adv_lam_r}) for the same input sample as in Figure \ref{fig: ADV_PI_deeponet_pred}.}
    \label{fig: ADV_lam_r}
\end{figure}

To further elucidate the important role played by the adaptive per-example weights $\lambda_k$, we can the weight distribution over the computational domain during training. Specifically, for some input sample $\bm{u}$, we can define the weight distribution associated with the PDE residual loss as
\begin{align}
    \label{eq: adv_lam_r}
    \lambda_r(\bm{u})(x, y) =  \sqrt{\frac{\|\bm{H}\|_\infty}{\|\nabla_{\bm{\theta}} \mathcal{R}_{\bm{\theta}}(\bm{u})(x,y) \|_2^2} }.
\end{align}
Figure \ref{fig: ADV_lam_r} shows the distribution of the residual weights corresponding to the same input sample as in Figure \ref{fig: ADV_PI_deeponet_pred}, at the end of training. Interestingly, the computed weights resemble an "attention map" that identifies regions of sharp gradients, in which small weights are assigned. This implies that Algorithm \ref{alg: NTK_weights} tends to relax the PDE constraint in such regions, hence making the residual loss easier to minimize. More visualizations of the model predictions and the associated weight distribution maps can be found in Appendix Figure \ref{fig: ADV_examples}. We can observe that this phenomenon is not accidental and strikingly manifests itself across all input samples.

% Because of the large improvement brought by moderate NTK weights, we are interested in 
% the weight distribution over the domain and its effect during training. 
% According to the update rule \ref{eq: lambda_update_moderate} , we have
% \begin{align}
%     \lambda_{\text{r},ij} \propto \frac{1}{\left\| \nabla_{\bm{\theta}} \mathcal{R}_{\bm{\theta}}(\bm{u}^{(i)})(x^{(i)}_{\text{r},j}, t^{(i)}_{\text{r},j}) \right\|_2}
% \end{align}
% where $\mathcal{R}_{\bm{\theta}}(\bm{u})(x,y)$ is defined in equation \ref{eq: Burger_residual}. Therefore, it suffices to fix some input sample $\bm{u}$ and visualize $1 / \|\nabla_{\bm{\theta}} \mathcal{R}(\bm{u})(x, t) \|_2$ over the domain. As shown in Figure 

\begin{table}[]
\renewcommand{\arraystretch}{1.8}
    \centering
    \begin{tabular}{|c|c|c|c|}
    \hline
        \diagbox{Method}{Architecture}   & MLP & Modified MLP & Modified DeepONet\\
        \hline
         $\lambda = 1$ & $6.39\% \pm 1.76\%$ & $6.83\% \pm 1.82\%$   &  $1.78\% \pm 0.42\%$ \\  
        \hline
        $\lambda_{\text{bc}} = \lambda_{\text{ic}} = \lambda$ & $4.73\% \pm 1.43\%$  & $3.74\% \pm 1.17\%$ & $1.03\% \pm 0.25\%$  \\
          \hline
        $\lambda = \frac{\|\bm{H}\|_\infty}{\text{diag}(\bm{H}) }$ &  $3.88\% \pm 1.42\%$ &   $2.74\% \pm 1.04\%$  & $1.19\% \pm 0.35 \%$\\
        \hline
        $\lambda = \sqrt{  \frac{\|\bm{H}\|_\infty}{\text{diag}(\bm{H}) }}$ & $4.22 \%  \pm 1.28 \% $ &   $3.37\% \pm 1.12\%$    &   $\mathbf{0.95\% \pm 0.23\%}$   \\
        \hline
    \end{tabular}
    \caption{{\em Advection equation:} Test errors of trained physics-informed DeepONets using different weighting schemes and network architectures. In particular, the results of $\lambda_{\text{ic}}  = \lambda_{\text{bc}}  = \lambda$  shows the best accuracy obtained by  training the physics-informed DeepONets for different $\lambda \in [10^{-2}, 10^{2}]$. The resulting test errors and their standard deviations are summarized in Figure \ref{fig: ADV_fixed_weights_error}. }
    \label{tab: adv_l2_error}
\end{table}

\subsection{Burgers' equation}
To demonstrate the effectiveness of the proposed approaches in tackling  nonlinear PDEs, let us consider the one-dimensional Burgers' equation
\begin{align}
    \label{eq: Burger_eq}
    \frac{ds}{dt} + s \frac{ds}{dx} - \nu \frac{d^2 s}{dx^2} = 0, \quad (x, t) \in (0,1) \times (0,1], 
\end{align}
subject to initial and boundary conditions
\begin{align}
        s(x,0) &= u(x), \quad x \in (0,1), \\
        s(0, t) &= s(1,t), \quad t \in (0,1),\\
     \frac{ds}{dx}(0,t) &= \frac{ds}{dx}(1,t), \quad t \in (0,1),
\end{align}
where the viscosity is set to $\nu = 10^{-3}$, and the initial condition $u(x)$ is generated from a GRF \cite{yang2018physics}.

Our objective here is to learn the solution operator mapping initial conditions $u(x)$ to the associated full spatio-temporal solution $s(x,t)$. We proceed by representing  the solution operator by a DeepONet $G_{\bm{\theta}}$. For any input sample $\bm{u}$, the PDE residual is then defined by
\begin{align}
    \label{eq: Burger_residual}
    R_{\bm{\theta}}(\bm{u})(x, t) = \frac{\partial G_{\bm{\theta}}(\bm{u})(x,t)}{\partial t} + G_{\bm{\theta}}(\bm{u})(x,t) \frac{\partial G_{\bm{\theta}}(\bm{u})(x,t)}{\partial x} 
    - \nu \frac{\partial^2 G_{\bm{\theta}}(\bm{u})(x,t)}{\partial x^2},
\end{align}
Consequently, a physics-informed DeepONet can be trained by minimizing the following weighted loss function with two equivalent forms
\begin{align}
    \label{eq: Burger_loss}
    \mathcal{L}(\bm{\theta}) &= \lambda_{\text{bc}} \mathcal{L}_{\text{BC}}(\bm{\theta}) + \lambda_{\text{ic}} \mathcal{L}_{\text{IC}}(\bm{\theta}) + \lambda_{\text{r}} \mathcal{L}_{\text{PDE}}(\bm{\theta}), \\
     \mathcal{L}(\bm{\theta}) &= \frac{1}{N^*} \sum_{k=1}^{N^*} \lambda_k \left| T^{(k)}(u^{(k)}(x_k), G_{\bm{\theta}}(\bm{u}^{(k)})(x_k, t_k))\right|^2,
\end{align}
where 
\begin{align}
     \mathcal{L}_{\text{IC}}(\bm{\theta}) &= \frac{1}{NP}\sum_{i=1}^N \sum_{j=1}^{P}  \left|G_{\bm{\theta}}(\bm{u}^{(i)})(x^{(i)}_{\text{ic}, j}, 0 ) - u^{(i)}(x^{(i)}_{\text{ic}, j})  \right|^2 \\
      \mathcal{L}_{\text{BC}}(\bm{\theta}) &= \frac{1}{NP}\sum_{i=1}^N \sum_{j=1}^{P} 
     \left|G_{\bm{\theta}}(\bm{u}^{(i)})(0, t^{(i)}_{\text{bc},j} ) - G_{\bm{\theta}}(\bm{u}^{(i)})(1, t^{(i)}_{\text{bc},j} )\right|^2 
    \\
    &+   \frac{1}{NP}\sum_{i=1}^N \sum_{j=1}^{P}  \left|\frac{\partial G_{\bm{\theta}}(\bm{u}^{(i)})(0, t^{(i)}_{\text{bc},j}) }{\partial x} - \frac{\partial G_{\bm{\theta}}(\bm{u}^{(i)})(1, t^{(i)}_{\text{bc},j}) }{\partial x} \right|^2 \\
    \mathcal{L}_{\text{PDE}}(\bm{\theta}) &= \frac{1}{NQ}\sum_{i=1}^N \sum_{j=1}^{Q} 
    \left|    R_{\bm{\theta}}(\bm{u}^{(i)})(x^{(i)}_{\text{r},j}, t^{(i)}_{\text{r},j})   \right|^2.
\end{align}

To obtain a set of training data, we randomly sample $N = 1,000$ input functions from a GRF $\sim \mathcal{N}\left(0,  25^2(-\Delta+5^2 I)^{-4}\right)$ and for every input sample $\bm{u}^{(i)}$,  we uniformly sample $P=100$ locations $x_{ic,j}^{(i)} = x_j$ and  $ \{(0, t_{ic,j}^{(i)})\}_{j=1}^P, \{(1, t_{ic,j}^{(i)})\}_{j=1}^P$ from each boundary, and $Q = 2500$ collocation points  $\{( x_{r,j}^{(i)}, t_{r,j}^{(i)})\}_{j=1}^Q$ from the domain interior. We generate the test data-set by sampling another $500$ input functions from the same GRF and solve the Burgers' equation (\ref{eq: Burger_eq}) using 
the Chebfun package \cite{driscoll2014chebfun} with a spectral Fourier discretization and a fourth-order stiff time-stepping scheme (ETDRK4) \cite{cox2002exponential} with a time-step size of $10^{-4}$.  Temporal snapshots of the solution are saved every $\Delta t = 0.01$ to give us 101 snapshots in total. Consequently, the test data-set contains $500$ realizations evaluated at a $100 \times 101$ spatio-temporal grid.

\begin{figure}
     \centering
     \begin{subfigure}[b]{0.8\textwidth}
         \centering
         \includegraphics[width=\textwidth]{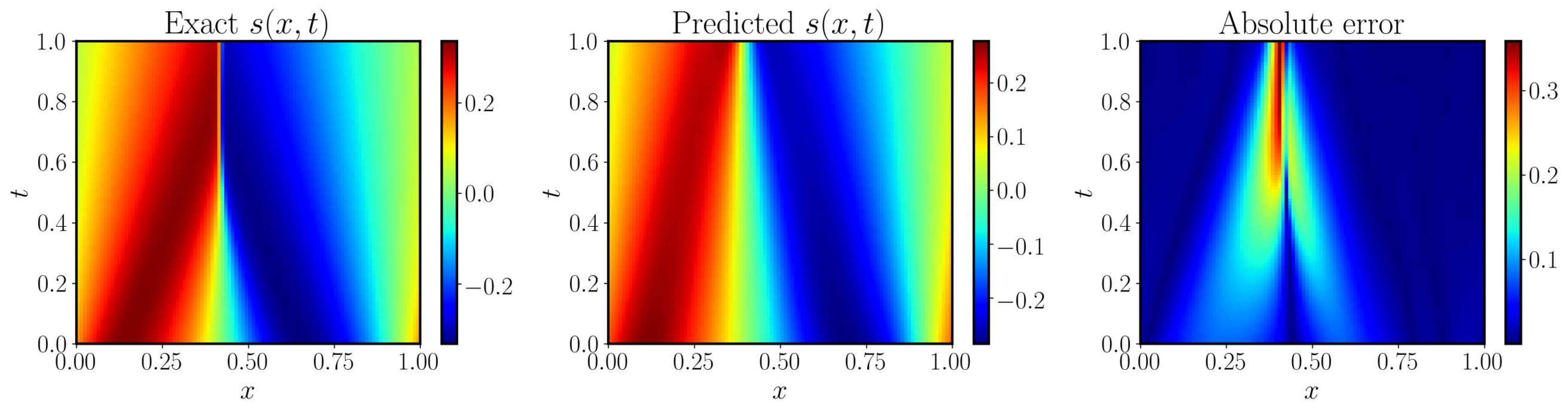}
        %  \caption{30.72\%}
     \end{subfigure}
     \begin{subfigure}[b]{0.6\textwidth}
         \centering
         \includegraphics[width=\textwidth]{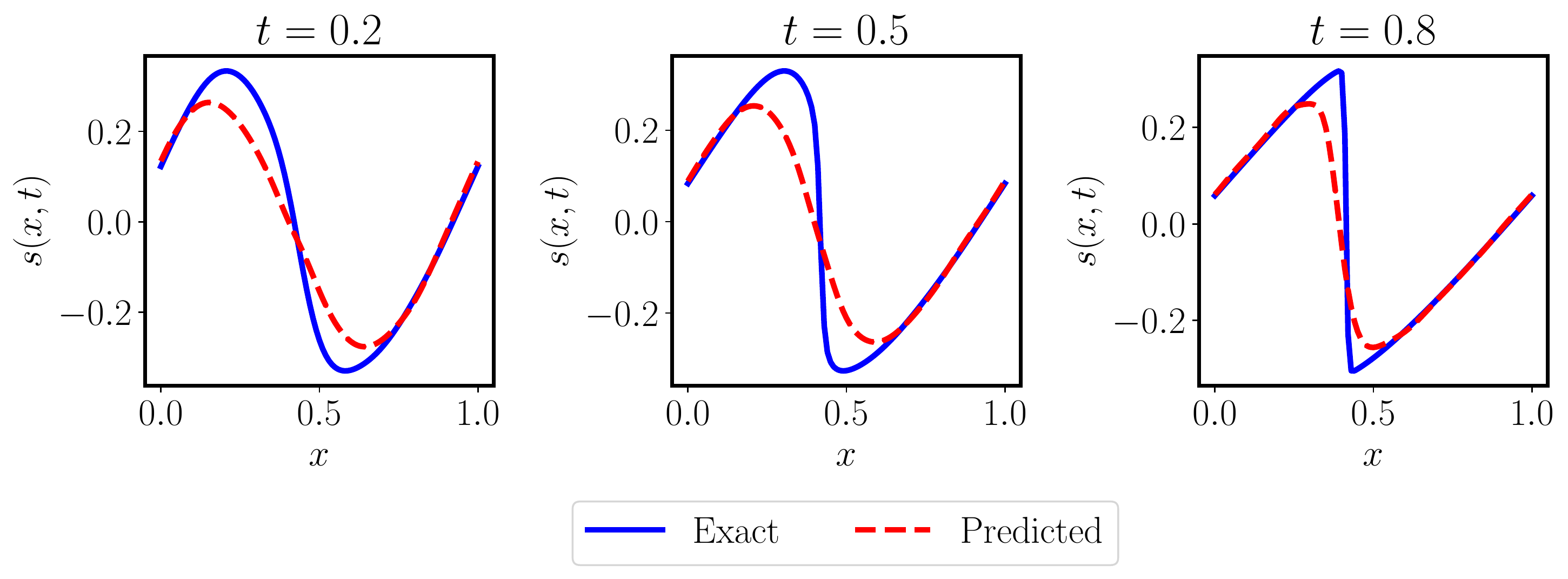}
        %  \caption{1.19\%}
     \end{subfigure}
        \caption{{\em Burger's equation:} {\em Top:}  Exact solution versus the prediction of a trained conventional physics-informed DeepONet for a representative example in the test data-set.  {\em Bottom:}  Comparison of the predicted and exact solutions corresponding to the three temporal snapshots at $t=0.2, 0.5, 0.8$. The resulting relative $L^2$ error is $30.72\%$}
        \label{fig: Burger_PI_deeponet_MLP_no_weights_pred}
\end{figure}

\begin{figure}
     \centering
     \begin{subfigure}[b]{0.8\textwidth}
         \centering
         \includegraphics[width=\textwidth]{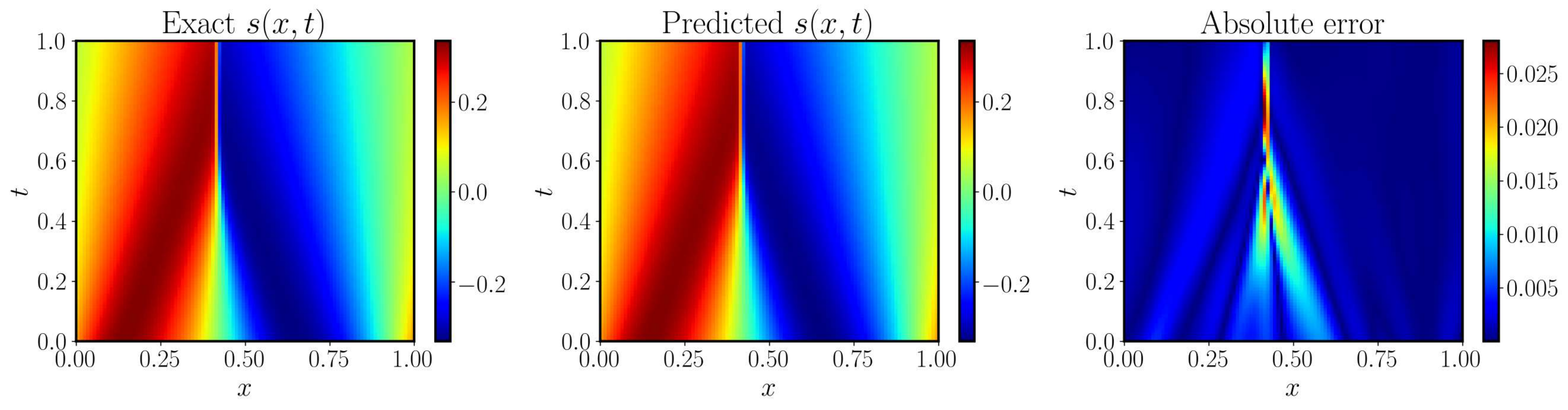}
        %  \caption{30.72\%}
     \end{subfigure}
     \begin{subfigure}[b]{0.6\textwidth}
         \centering
         \includegraphics[width=\textwidth]{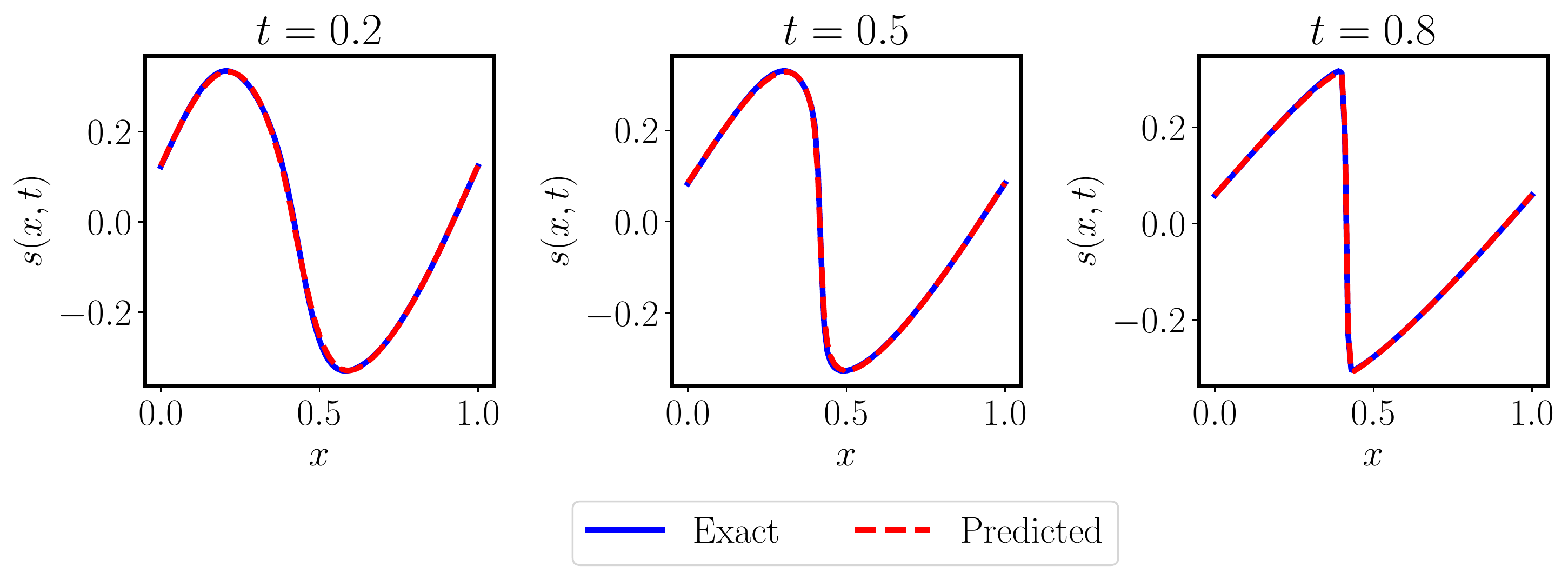}
        %  \caption{1.19\%}
     \end{subfigure}
        \caption{{\em Burger's  equation:} {\em Top:}  Exact solution versus the prediction of a  physics-informed DeepONet represented by modified DeepONet architecture and trained  using Algorithm \ref{alg: NTK_weights} with $\alpha = \frac{1}{2}$ for the same example as in Figure \ref{fig: Burger_PI_deeponet_MLP_no_weights_pred}..   {\em Bottom:}  Comparison of the predicted and exact solutions corresponding to the three temporal snapshots at $t=0.2, 0.5, 0.8$. The resulting relative $L^2$ error is $1.19\%$, which is 30x more accurate than the original formulation. }
        \label{fig: Burger_PI_deeponet_modified_deeponet_moderate_local_NTK_weights_pred}
\end{figure}

\begin{figure}
    \centering
    \includegraphics[width=0.3\textwidth]{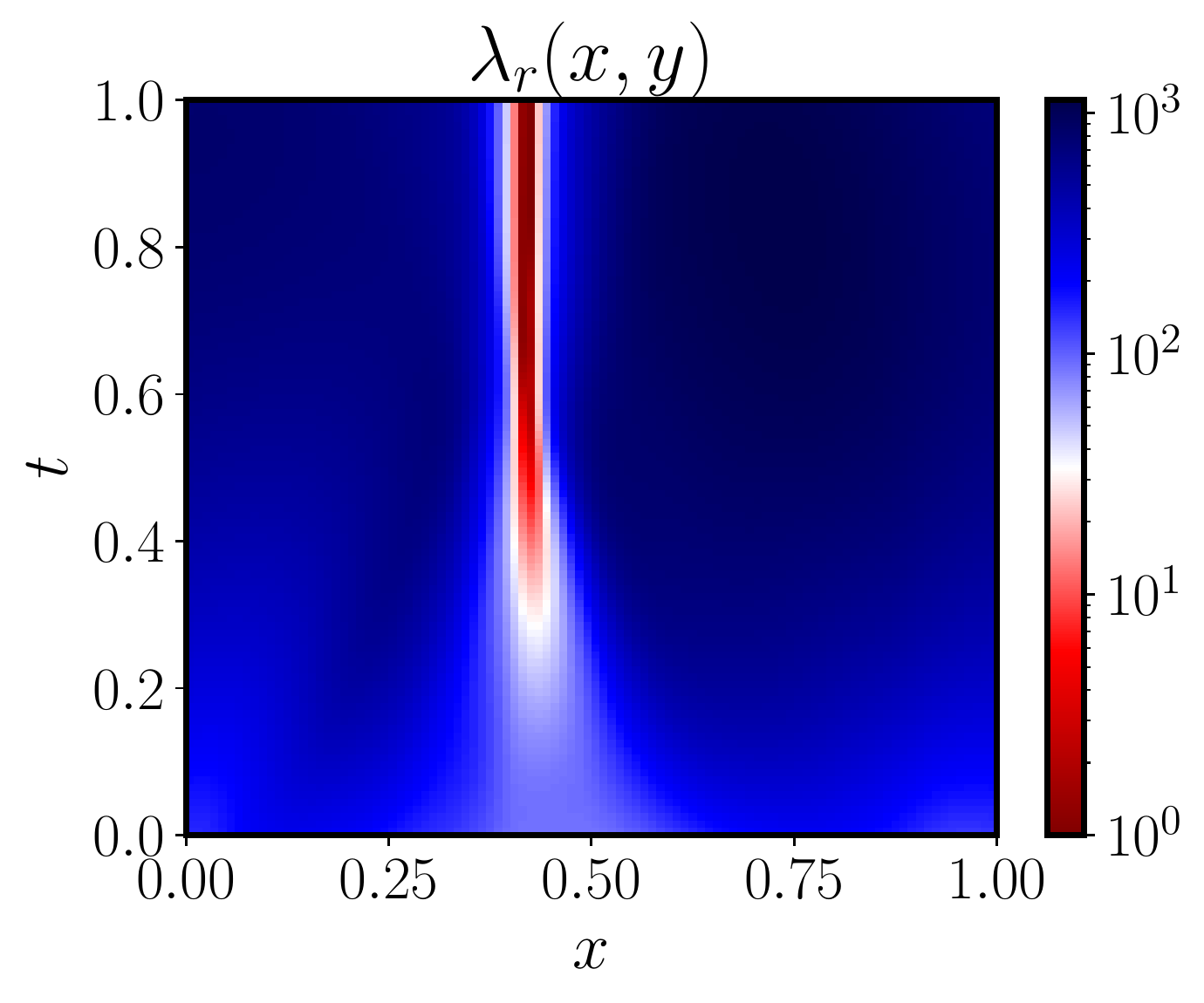}
    \caption{{\em Burger's equation:} Weight distribution defined in equation (\ref{eq: adv_lam_r}) for the same input sample as in Figure \ref{fig: Burger_PI_deeponet_MLP_no_weights_pred}.}
    \label{fig: Burger_lam_r}
\end{figure}

Table \ref{tab: Burger_l2_error} summarizes the test error of the trained physics-informed DeepONets with different weighting schemes and architectures after $2 \times 10^5$ iterations of gradient descent.  Compared to the poor performance of the MLP and modified MLP architectures, the proposed modified DeepONet achieves a test error of $8.10 \%$  even without any weights. 
The error can be further reduced to $3.69 \%$ by using  NTK weights with $\alpha=\frac{1}{2}$, yielding a result that is at least 5x more accurate than the baseline physics-informed DeepONets \cite{wang2021learning}. To further compare the performance of the different trained models, we present a representative predicted solution obtained from the worst and the best trained models in Figure \ref{fig: Burger_PI_deeponet_MLP_no_weights_pred} and Figure \ref{fig: Burger_PI_deeponet_modified_deeponet_moderate_local_NTK_weights_pred}, respectively. We can observe that the latter can accurately capture the viscous shock wave and is in excellent agreement with the ground truth. The resulting relative $L^2$ error is $1.19 \%$, which is about 30x more accurate than the former one ($30.72\%$). In Figure \ref{fig: Burger_lam_r},  we also visualize the weights distribution of the PDE residual corresponding to the same input sample as in Figure \ref{fig: Burger_PI_deeponet_modified_deeponet_moderate_local_NTK_weights_pred}.  
We can observe a similar behavior as in the previous benchmark, i.e., the region of viscous shocks are identified and assigned small weights. Additional representative predictions with their associated residual weight distribution maps corresponding to different input functions are shown in Figure \ref{fig: Burger_001_examples}.

Furthermore, we also investigate the effect of the viscosity parameter in the Burgers' equation on the performance of physics-informed DeepONets. In particular, we vary the viscosity $\nu$ from $10^{-2}$ to $10^{-4}$ and train physics-informed DeepONets  with different architectures and weighting schemes, under exactly the same hyper-parameter settings (i.e. learning rate, batch size, optimizer, etc).  Table \ref{tab: Burger_nu_l2_error} summarizes the test error of the best and the worst trained models. These results further validate the remarkable improvements of the proposed architecture and adaptive training algorithms. It is no surprise that the accuracy decreases as we reduce the viscosity since smaller viscosity typically results in sharper gradients and stiffer dynamics, leading to a harder optimization problem. 
% Nevertheless, our approaches appears promising and we believe that these issues will be eventually resolved in the future by designing  more appropriate architectures, as well as more effective optimization algorithms for training constrained neural network models, such as PINNs and  physics-informed DeepONets. 

\begin{table}[]
\renewcommand{\arraystretch}{1.8}
    \centering
    \begin{tabular}{|c|c|c|c|}
    \hline
        \diagbox{Method}{Architecture}   & MLP & Modified MLP & Modified DeepONet\\
        \hline
        $\lambda=1$ & $26.91\% \pm 12.26\% $   & $27.46\% \pm 16.54\% $    &  $8.10\% \pm 6.44\% $  \\  
        \hline
         $\lambda_{\text{ic}} = \lambda_{\text{bc}} = \lambda$ & $22.20\% \pm 9.50\% $  & $22.87\% \pm 9.22\% $ & $8.10\% \pm 6.44\% $  \\
          \hline
        $\lambda = \frac{\|\bm{H}\|_\infty}{\text{diag}(\bm{H}) }$  &   $ 23.85\% \pm 8.08\% $   &  $15.76\% \pm 12.27\% $   & $ 5.53\% \pm 5.95\% $ \\
        \hline
        $\lambda = \sqrt{  \frac{\|\bm{H}\|_\infty}{\text{diag}(\bm{H}) }}$  & $19.87\% \pm 11.06\% $   &  $20.21\% \pm 14.86\% $ &   $ \mathbf{3.69\% \pm 3.63\%} $   \\
        \hline
    \end{tabular}
    \caption{{\em Burger's equation:} Test errors of trained physics-informed DeepONets using different weighting schemes and network architectures. In particular, the results of $\lambda_{\text{ic}}  = \lambda_{\text{bc}}  = \lambda$  shows the best accuracy obtained by  training the physics-informed DeepONets for different $\lambda \in [10^{-2}, 10^{2}]$. The resulting test errors and their standard deviations are visualized in Figure \ref{fig: Burger_fixed_weights_error}. }
    \label{tab: Burger_l2_error}
\end{table}

\begin{table}[]
\renewcommand{\arraystretch}{1.4}
    \centering
    \begin{tabular}{|c|c|c|c|}
    \hline
     \diagbox{Model}{Viscosity}     & $\nu=10^{-2}$ & $\nu=10^{-3}$ & $\nu=10^{-4}$  \\
     \hline
      Plain PI-DeepONet   & $3.48\% \pm 3.79\%$     &  $26.91\% \pm 12.26\%$    &  $29.21\% \pm 12.38\%$    \\
      \hline
      Improved PI-DeepONet    & $1.03\% \pm 1.35\%$    &$3.69\% \pm 3.63\%$    &  $7.95\% \pm 5.29\%$  \\
      \hline
    \end{tabular}
    \caption{{\color{black}{\em Solving  a parametric Burger's equation:}  Relative $L^2$ prediction error of a trained physics-informed DeepONet averaged over all examples in the test data-set, for different viscosity values.}}
    \label{tab: Burger_nu_l2_error}
\end{table}

\subsection{Stokes flow}

\begin{figure}
    \centering
    \includegraphics[width=0.5\textwidth]{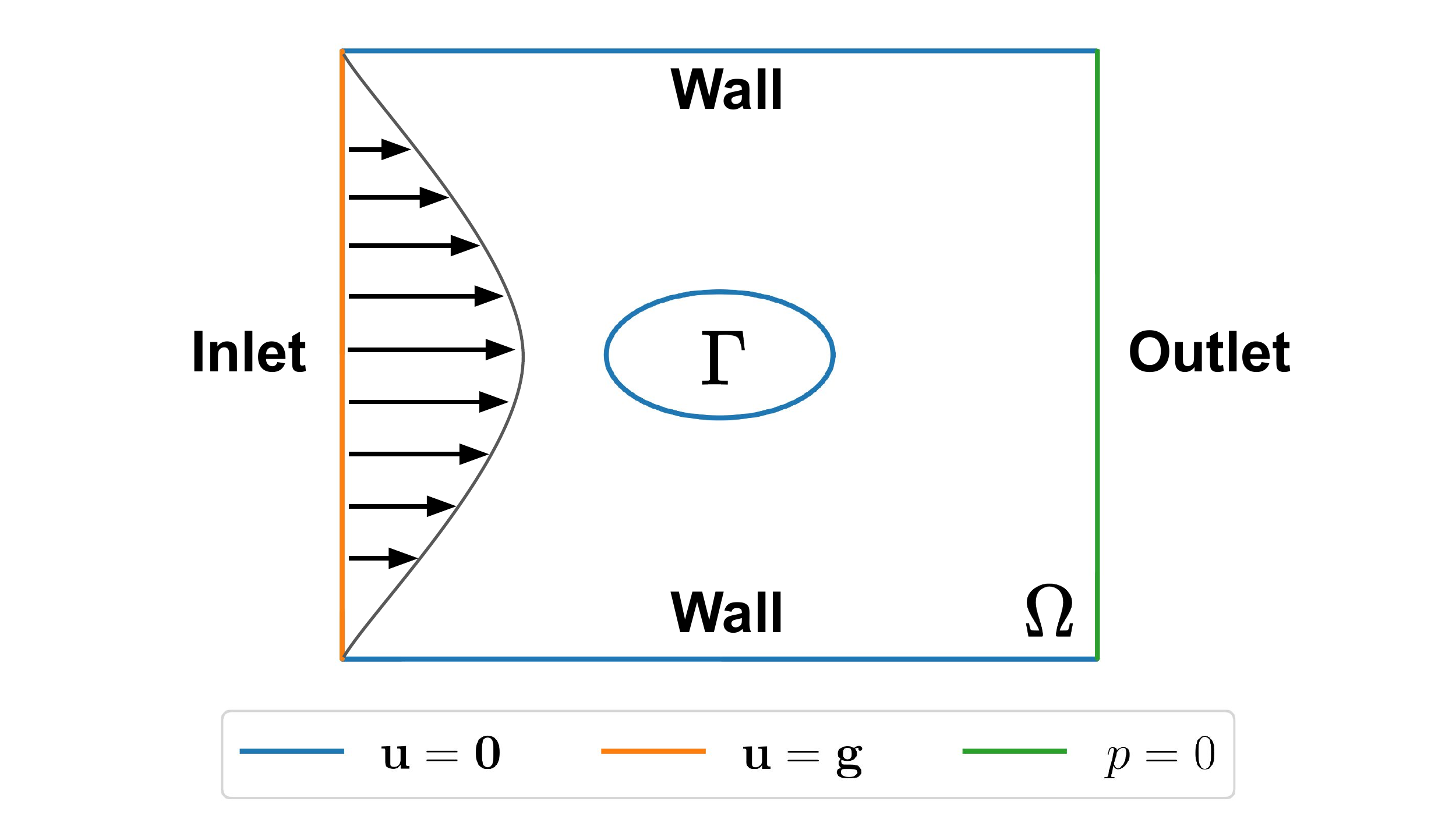}
    \caption{{\em Stokes equation:} Illustration of the computational domain and boundary conditions.}
    \label{fig: Stokes_domain}
\end{figure}

Our last example aims to highlight the capability of the proposed methods on solving PDEs in domains of varying shapes. To this end, we consider an example of a two-dimensional Stokes flow over an obstacle. The associated PDE system takes the form
% \begin{align}
%      - (\frac{\partial^2 u}{\partial x^2} + \frac{\partial^2 u}{\partial y^2}) + \frac{\partial p}{\partial x} &= 0 \\
%      - (\frac{\partial^2 v}{\partial x^2} + \frac{\partial^2 v}{\partial y^2}) + \frac{\partial p}{\partial y} &= 0 \\
%      \frac{\partial u}{\partial x} + \frac{\partial v}{\partial y} &= 0 
% \end{align}
\begin{align}
    \label{eq: stokes_1}
    -\Delta \bm{u} + \nabla p &= 0, \quad (x,y) \in \Omega / \Gamma,    \\
    \label{eq: stokes_2}
    \nabla \cdot \bm{u}    &= 0, \quad (x,y) \in \Omega / \Gamma,
\end{align}
subject to boundary conditions
\begin{align}  
    \label{eq: stokes_3}
    \bm{u} &= \bm{0}, \quad (x,y) \in \Lambda_1 \cup \partial \Gamma, \\
        \label{eq: stokes_4}
    \bm{u} &= \bm{g}, \quad (x,y) \in \Lambda_2 \\
    \label{eq: stokes_5}
        p  & = 0,     \quad (x,y) \in \Lambda_3,
\end{align}
where $\bm{u} = (u, v)$ denotes the velocity field, and $\nabla p$ is the gradient of the pressure. In addition, $\Omega = [0, 1] \times [0, 1]$ denotes the computational domain, $\Gamma$ represents an obstacle, and $\Lambda_1$ denotes wall boundaries,  while $ \Lambda_2$ and $\Lambda_3$ correspond to the inlet and the outlet of the channel, respectively.
As illustrated in Figure \ref{fig: Stokes_domain},  we  assume a no-slip boundary condition at the walls and the obstacle surface, impose an initial velocity as $\bm{g} = (\sin(\pi y), 0 )$ at the inlet, and a zero pressure condition at the outlet.

Our goal is to learn the solution operator $G$ that maps different obstacle shapes $\partial \Gamma$ to the associated PDE solution triplet $(u, v, p)$. In this example, we parameterize $\partial \Gamma$  by a family of ellipses centered at $(\frac{1}{2}, \frac{1}{2})$, i.e
\begin{align}
    \label{eq: curve}
    \partial \Gamma  =  \partial \Gamma (\phi) = (a \cos(\phi) + \frac{1}{2}, b \sin(\phi) + \frac{1}{2}), \quad \phi \in [0, 2\pi),
\end{align}
where $a, b > 0$ denote the length of the long or the short axis.

We approximate the solution operator by a DeepONet $G_{\bm{\theta}}$ with 3-dimensional vector-valued outputs for representing $u, v, p$, respectively (see equation (\ref{eq: deeponet_multiple_outputs})), 
\begin{align}
    \partial \Gamma  \overset{G_{\bm{\theta}}} {\longrightarrow} [G^{(u)}_{\bm{\theta}}, G^{(v)}_{\bm{\theta}}, G^{(p)}_{\bm{\theta}}].
\end{align}
Now we can define the following PDE residuals
\begin{align}
     \mathcal{R}_{\bm{\theta}}^{(1)}(\partial  \bm{\Gamma})(x, y) &= - \frac{\partial^2 G_{\bm{\theta}}^{(u)}( \partial \bm{\Gamma})(x, y) }{\partial x^2} - \frac{\partial^2 G_{\bm{\theta}}^{(u)}(\partial \bm{\Gamma})(x, y) }{\partial y^2} + \frac{\partial G_{\bm{\theta}}^{(p)}(\partial \bm{\Gamma})(x, y) }{\partial x}, \\
          \mathcal{R}_{\bm{\theta}}^{(2)}(\partial  \bm{\Gamma})(x, y) &= - \frac{\partial^2 G_{\bm{\theta}}^{(v)}(\partial \bm{\Gamma})(x, y) }{\partial x^2} - \frac{\partial^2 G_{\bm{\theta}}^{(v)}(\partial \bm{\Gamma})(x, y) }{\partial y^2} + \frac{\partial G_{\bm{\theta}}^{(p)}(\partial \bm{\Gamma})(x, y) }{\partial y}, \\
          \mathcal{R}_{\bm{\theta}}^{(3)}( \partial  \bm{\Gamma})(x, y) &=  \frac{\partial G_{\bm{\theta}}^{(u)}(\partial \bm{\Gamma})(x, y) }{\partial x} + \frac{\partial G_{\bm{\theta}}^{(v)}( \partial \bm{\Gamma})(x, y) }{\partial y},
\end{align}
where $\partial \bm{\Gamma}  = [\partial \Gamma(\phi_1), \partial \Gamma(\phi_2), \dots, \partial \Gamma(\phi_m)]$ denotes an input closed curve evaluated at evenly-spaced grid points $\{\phi\}_{i=1}^m$ in $[0, 2\pi]$.

According to equation (\ref{eq: stokes_1}) - (\ref{eq: stokes_5}), a  physics-informed loss can be formulated as follows
\begin{align}
    \label{eq: Stokes_loss}
    \mathcal{L}(\bm{\theta}) = \sum_{k=1}^3 \mathcal{L}^{(k)}_{\text{BC}}(\bm{\theta}) + \sum_{k=1}^3 \mathcal{L}^{(k)}_{\text{PDE}}(\bm{\theta}),
\end{align}
where
\begin{align}
    \mathcal{L}^{(1)}_{\text{BC}}(\bm{\theta}) &= \frac{1}{NP} \sum_{i=1}^N \sum_{j=1}^{P}  \left[   \left| G^{(u)}_{\bm{\theta}}(\partial \bm{\Gamma}^{(i)}) (x_{\text{bc1}, j}^{(i)}, y_{\text{bc1},j}^{(i)}) \right|^2 +   \left| G^{(v)}_{\bm{\theta}}(\partial \bm{\Gamma}^{(i)}) (x_{\text{bc1}, j}^{(i)}, y_{\text{bc1},j}^{(i)}) \right|^2 \right], \\
    \mathcal{L}^{(2)}_{\text{BC}}(\bm{\theta}) &= \frac{1}{NP} \sum_{i=1}^N \sum_{j=1}^{P}  \left[ \left| G^{(u)}_{\bm{\theta}}(\partial \bm{\Gamma}^{(i)}) (x_{\text{bc2}, j}^{(i)}, y_{\text{bc2},j}^{(i)}) - \sin(\pi y_{\text{bc2},j}^{(i)}) \right|^2 +  \left| G^{(v)}_{\bm{\theta}}(\partial \bm{\Gamma}^{(i)}) (x_{\text{bc2}, j}^{(i)}, y_{\text{bc2},j}^{(i)}) \right|^2 \right], \\
    \mathcal{L}^{(3)}_{\text{BC}}(\bm{\theta}) &= \frac{1}{NP} \sum_{i=1}^N \sum_{j=1}^{P}\left| G^{(p)}_{\bm{\theta}}(\partial \bm{\Gamma}^{(i)}) (x_{\text{bc3}, j}^{(i)}, y_{\text{bc3},j}^{(i)}) \right|^2,
\end{align}
and 
\begin{align}
   \mathcal{L}^{(k)}_{\text{PDE}}(\bm{\theta}) = \frac{1}{NQ} \sum_{i=1}^N \sum_{j=1}^{Q}  \left|  \mathcal{R}_{\bm{\theta}}^{(k)}(\partial \bm{\Gamma}^{(i)})(x^{(i)}_{\text{r}, j}, y^{(i)}_{\text{r}, j})   \right|^2, \quad k = 1,2,3.
\end{align}
Here, for each input sample $\partial \bm{\Gamma}^{(i)}$, $\{x_{\text{bc1}, j}^{(i)}, y_{\text{bc1},j}^{(i)}  \}_{j=1}^P$, $\{x_{\text{bc2}, j}^{(i)}, y_{\text{bc2},j}^{(i)}  \}_{j=1}^P$, $\{x_{\text{bc3}, j}^{(i)}, y_{\text{bc3},j}^{(i)}  \}_{j=1}^P$ and $\{x^{(i)}_{\text{r}, j}, y^{(i)}_{\text{r}, j} \}_{j=1}^Q$ are uniformly sampled at the boundaries $\Lambda_1 \cup \partial \Gamma$, $\Lambda_2$, $\Lambda_3$  and inside the domain $\Omega / \Gamma$, respectively.  In this example, we set $N = 1000, m=P=100, Q=2500$. To prepare the training data-set, we randomly sample  $a, b$ from uniform distribution $\mathcal{U}(0.05, 0.3)$ and obtain a set of input curves $\{\partial \bm{\Gamma}^{(i)} \}_{i=1}^N$ using equation (\ref{eq: curve}). To generate a set of test data, we repeat the same procedure to obtain 500 new curves and obtain the corresponding PDE solutions using the FEniCS solver \cite{alnaes2015fenics}.

One can see that there are several terms in the loss function that it is almost impossible to assign weights manually. Therefore, here we will attempt any hyper-parameter tuning and only test the performance of Algorithm \ref{alg: NTK_weights} for different choices of underlying network architectures. To this end, we reformulate the above loss function into a weighted fully-decoupled form
\begin{align}
     \mathcal{L}(\bm{\theta}) = \frac{1}{N^*} \sum_{k=1}^{N^*} \lambda_k \left| T^{(k)}(\partial \Gamma^{(k)}, G_{\bm{\theta}}(\partial \bm{\Gamma}^{(k)})(x_k, y_k))\right|^2,
\end{align}
where  all weights $\{\lambda_k\}_{k=1}^{N^*}$ can be either set to $1$, or be updated using Algorithm \ref{alg: NTK_weights}  at each training iteration.

\begin{figure}
    \centering
    \includegraphics[width=0.8\textwidth]{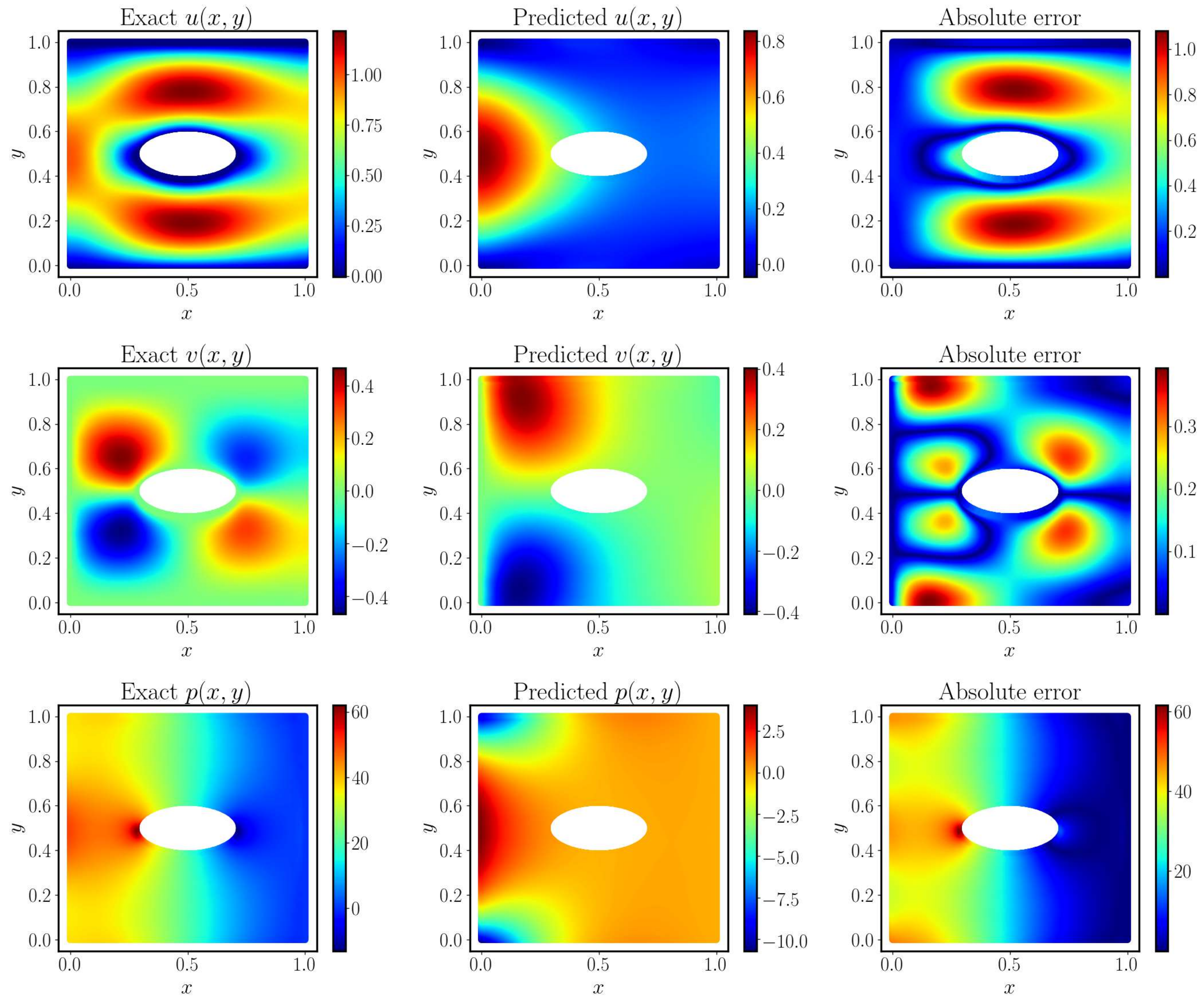}
    \caption{{\em Stokes equation:}  Exact solution versus the prediction of a trained conventional physics-informed DeepONet for a representative example in the test data-set. The resulting relative $L^2$ errors are $74.32\%, 82.64\%, 99.63\%$ for $u,v,p$, respectively. The trained model completely fails to predict the correct solution.}
    \label{fig: Stokes_PI_deeponet_MLP_no_weights_pred}
\end{figure}

\begin{figure}
    \centering
    \includegraphics[width=0.8\textwidth]{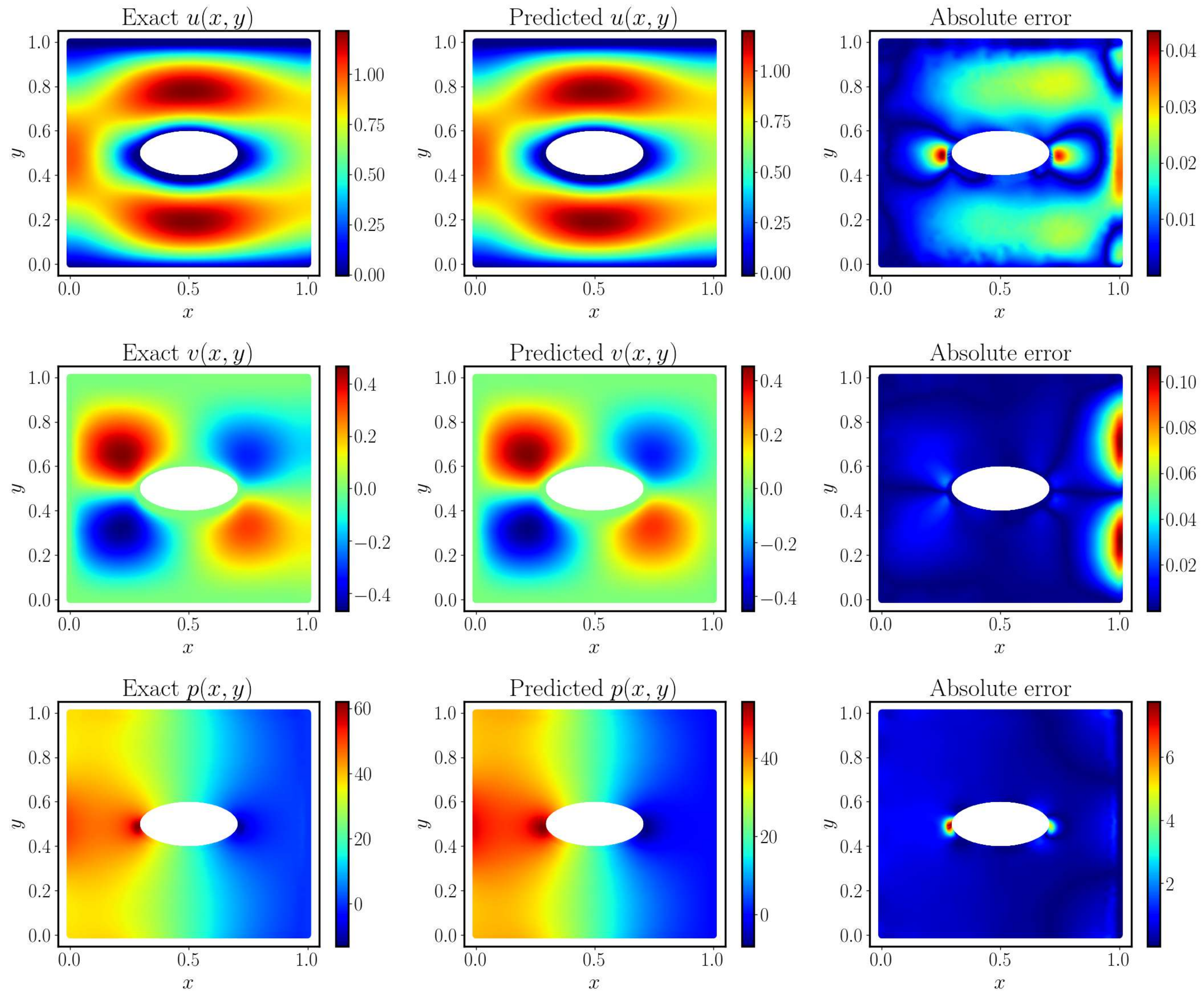}
    \caption{{\em Stokes equation:}  Exact solution versus the prediction of a  physics-informed DeepONet represented by modified DeepONet architecture and trained using Algorithm \ref{alg: NTK_weights} for the same example as in Figure \ref{fig: Stokes_PI_deeponet_MLP_no_weights_pred}. The resulting relative $L^2$ errors are $1.88\%, 5.74\%, 2.41\%$ for $u,v,p$, respectively.}
    \label{fig: Stokes_PI_deeponet_modified_deeponet_local_NTK_weights_pred}
\end{figure}

We train the DeepONet by minimizing the loss function \ref{eq: Stokes_loss} for $2 \times 10^5$ iterations of gradient descent using Adam optimizer. The test error of the trained DeepONets with different network architectures and weighting schemes is summarized in Table \ref{tab: Stokes_l2_error}. In contrast to the two previous examples, the results of "Fixed weights" are not included because there are too many terms in the loss function to tune weights manually.
Among all trained models, one can see that the modified DeepONet architecture with NTK weights yields the best predictive accuracy. 
Remarkably, this result is 50x more accurate than the baseline physics-informed DeepONet \cite{wang2021learning}. Another key observation is that using NTK weights or the modified DeepONet architecture alone is not sufficient for accurately approximating  the solution operator. This strongly suggests that appropriate weighting schemes and network architectures together play  a vital role in enhancing the trainability, as well as the performance of physics-informed DeepONets. 
Figure \ref{fig: Stokes_PI_deeponet_MLP_no_weights_pred} and Figure \ref{fig: Stokes_PI_deeponet_modified_deeponet_local_NTK_weights_pred} provide a  more detailed visual assessment of one representative predicted solution  obtained by the best and the worst trained models. As it can be seen, the baseline DeepONet completely fails to predict the correct velocity and pressure field, while the modified DeepONet architecture produces predictions that achieve an excellent agreement with the corresponding reference solution.  Additional predicted solutions for different input shapes are shown in Appendix Figure \ref{fig: Stokes_PI_deeponet_modified_deeponet_local_NTK_weights_pred_5} - \ref{fig: Stokes_PI_deeponet_modified_deeponet_local_NTK_weights_pred_144}.

\begin{figure}
    \centering
    \includegraphics[width=0.8\textwidth]{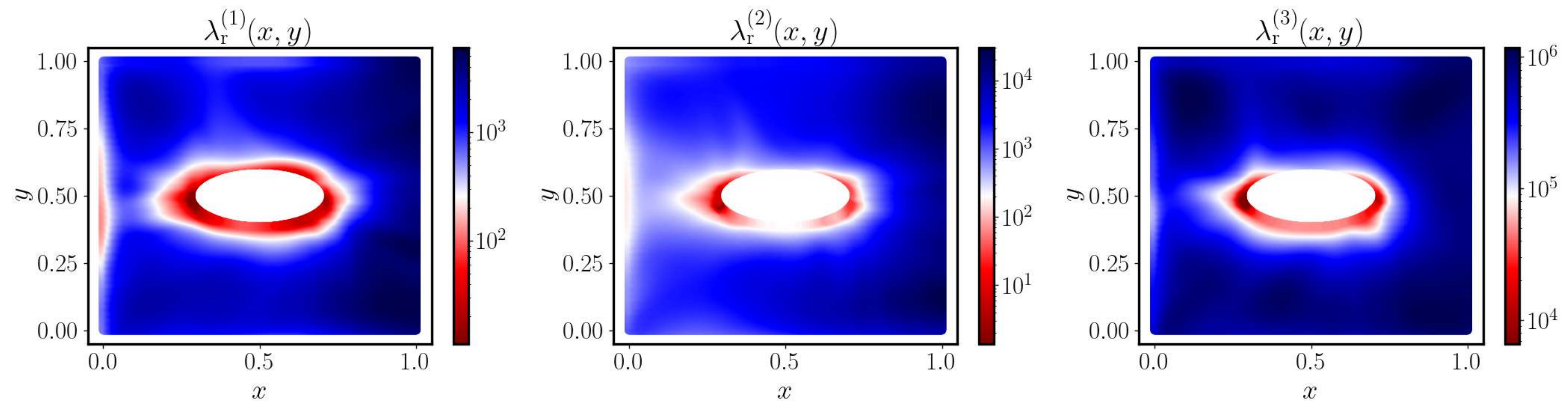}
    \caption{{\em Stokes equation:} Weight distribution defined in equation (\ref{eq: stokes_lam_r}) for the same input sample as in Figure \ref{fig: Stokes_PI_deeponet_MLP_no_weights_pred}.}
    \label{fig: Stokes_lam_r}
\end{figure}

To further elucidate the important role played by the NTK weights, we compute the weight distribution over the domain during training. According to the PDE residual loss, we can define
\begin{align}
    \label{eq: stokes_lam_r}
    \lambda_r^{(k)}(\partial \bm{\Gamma})(x, y) &=  \frac{\|\bm{H}\|_\infty}{\|\nabla_{\bm{\theta}} \mathcal{R}_{\bm{\theta}}^{(k)}(\partial \bm{\Gamma})(x,y) \|_2^2}, \quad k = 1,2, 3,
\end{align}
which denote the weights for the PDE residuals enforcing the momentum equations and the continuity equation, respectively. The weight distributions for the best trained model are presented in Figure \ref{fig: Stokes_lam_r}. One can observe red belts surrounding the obstacle, which implies that these weights attain small magnitude near it. 
% In some sense, the NTK weights achieves a similar effect as using a Signed Distance Function (SDF) for loss weighting in the SimNet solver \cite{hennigh2020nvidia}. The latter is empirically proven to yield better loss convergence and more accurate predictions especially for complex geometries. 
Another key observation is that the weight distribution  $\lambda_\text{r}^{(3)}$ corresponding to the continuity equation is two orders of magnitude greater than the other two weight distributions   $\lambda_\text{r}^{(1)}$ and  $\lambda_\text{r}^{(2)}$, respectively; a fact that suggests the difficulty and the necessity of imposing the divergence-free condition for physics-informed DeepONets. More visualizations for different obstacle geometries are provided in Appendix Figure \ref{fig: Stokes_PI_deeponet_modified_deeponet_local_NTK_weights_pred_5} - \ref{fig: Stokes_PI_deeponet_modified_deeponet_local_NTK_weights_pred_144} from which a similar behavior can be observed.

\begin{table}[]
\renewcommand{\arraystretch}{1.4}
    \centering
    \begin{tabular}{|c|c|c|c|}
    \hline
        \diagbox{Method}{Architecture}   & MLP & Modified MLP & Modified DeepONet\\
        \hline
       \multirow{3}{*}{ $\lambda = 1$}  &  $73.53 \% \pm 3.75\%   $   & $73.72 \% \pm 3.83\%   $    &  $55.21 \% \pm 7.79\%   $   \\
                                     &  $93.17 \% \pm 11.58\%   $  & $92.97  \% \pm 11.84\%   $    &  $62.07 \% \pm 5.63\%   $ \\
                                     &  $100.00 \% \pm 1.06\%   $  & $100.00 \% \pm 1.14\%   $    &  $71.78 \% \pm 6.21\%   $ \\
        \hline
    \multirow{3}{*}{  $\lambda = \frac{\|\bm{H}\|_\infty}{\text{diag}(\bm{H}) }$ }&  $54.12 \% \pm 61.17\%   $   & $2.94 \% \pm 1.12\%   $    &  $ \mathbf{1.69 \% \pm 0.34\% }  $    \\
                                     &  $100.00 \% \pm 0.01\%   $  & $10.36 \% \pm 2.15\%   $    &  $\mathbf{6.05 \% \pm 1.31\%}   $ \\
                                     &  $90.45 \% \pm 2.79\%   $  & $5.22 \% \pm 2.91\%   $    &  $\mathbf{3.83 \% \pm 1.89\%  } $\\
          \hline
\multirow{3}{*}{ $\lambda = \sqrt{\frac{\|\bm{H}\|_\infty}{\text{diag}(\bm{H}) }}$ } &  $20.49 \% \pm 7.42\% $   & $17.10 \% \pm 7.17\%   $  &  $ {4.19 \% \pm 1.80\% } $   \\
                                     &  $30.87 \% \pm 11.91\%   $  & $31.34 \% \pm 14.11\%   $    &  $10.86 \% \pm 2.53\%   $ \\
                                     &  $28.13 \% \pm 7.51\%   $  & $22.73 \% \pm 7.57\%   $    &  $6.45 \% \pm 2.98\% $ \\
          \hline
    \end{tabular}
    \caption{{\em Stokes equation:} Test errors of trained physics-informed DeepONets using different weighting schemes and network architectures. We remark that, in each cell, three rows represents the test error corresponding to the velocity $u,v$ and the pressure $p$, respectively. }
    \label{tab: Stokes_l2_error}
\end{table}

\section{Discussion}
\label{sec: discussion}

This work proposes a novel training algorithm to calibrate the convergence rate and back-propagated gradients of DeepONets and physics-informed DeepONets, by assigning appropriate weights to each individual term in their loss function during training. This is accomplished by performing a rigorous analysis of the training dynamics of DeepONets via Neural Tangent Kernel (NTK) theory. In addition, we design a novel DeepONet architecture which better preserves the propagation of inputs signals through the network's layers. Our empirical evidence suggests that this architecture is consistently better than conventional DeepONets in terms of approximating abstract non-linear operators. Taken  together, our developments provide new insights into the training of DeepONets and physics-informed DeepONets and yield significant improvement in their predictive accuracy by a factor of 10-50x across a range of benchmarks involving learning the solution operator of parametric PDEs with inherently different dynamics. 

% optimial weights is problem dependent and architecture dependent

Despite some promising results reported here, there are numerous open questions worth further investigation.  From a theoretical standpoint, in  contrast to conventional DeepONets guaranteed by universal approximation theorem of operators \cite{lu2021learning,lanthaler2021error}, little is known about  the approximation properties of the physics-informed DeepONets trained by minimizing a loss function where PDE constraints are encoded. Given a series of established a-priori error estimates for physics-informed neural networks (PINNs) \cite{shin2020convergence, mishra2020estimates}, in a parallel thrust, it would be interesting to provide rigorous mathematical justifications for physics-informed DeepONets.
From a methodological standpoint, there is still plenty of room for improving the performance of physics-informed DeepONets. For one thing, the proposed training algorithm is based on a simple NTK analysis of  full-batch gradient descent. The effect of mini-batch gradient descent with momentum on the training dynamics remains poorly understood.
From an applications standpoint, physics-informed DeepONets can be employed as a fast and differentiable surrogate for tackling general PDE-constrained optimization problems that routinely arise in science and engineering (e.g., design and control optimization problems), which generally require numerous evaluations of costly forward solvers or experiments. 

% We believe that addressing these open questions will not only cast new insight into the optimization process of physics-informed DeepONets but also pave a new way to developing scientific machine learning algorithms with better robustness and accuracy guarantees, as needed for many critical applications in computational science and engineering.

% Moreover, while the remarkable success of ResNets, CNNs, RNNs and Transformers stem from their extraction of key principles in computer vision and natural language processing, the key principles that benefit networks for learning functions or operators is wide open. That is currently the main reason why PINNs or physics-informed DeepONets mainly employ fully-connected neural networks as backbone. Our work has made some preliminary attempts along this direction, but the question of how to design general and powerful network architectures specialized in physics-informed machine learning still remains wide open.

\section*{Acknowledgments}
This work received support from DOE grant DE-SC0019116, AFOSR grant FA9550-20-1-0060, and DOE-ARPA grant DE-AR0001201. We would like to thank Dr. Mohamed Aziz Bhouri for the data preparation using Fenics \cite{alnaes2015fenics, mitusch2019dolfin}.
We would also like to thank the developers of the software that enabled our research, including JAX \cite{jax2018github}, Matplotlib \cite{hunter2007matplotlib}, and NumPy \cite{harris2020array}.

\section*{Competing Interests}
The authors declare that they have no competing interests.

\section*{Author Contributions}
SW and PP conceptualized the research and designed the numerical studies. SW implemented the methods and conducted the numerical experiments. HW assisted with the numerical studies. PP provided funding and supervised all aspects of this work. SW and PP wrote the manuscript.

\section*{Data and materials availability}
All methods needed to evaluate the conclusions in the paper are present in the paper and/or the Appendix. All code and data accompanying this manuscript will be made publicly available at \url{https://github.com/PredictiveIntelligenceLab/ImprovedDeepONets}.

\bibliographystyle{unsrt}
\bibliography{references}

\begin{thebibliography}{10}

\bibitem{lanthaler2021error}
Samuel Lanthaler, Siddhartha Mishra, and George~Em Karniadakis.
\newblock Error estimates for {DeepONet}: A deep learning framework in infinite
  dimensions.
\newblock {\em arXiv preprint arXiv:2102.09618}, 2021.

\bibitem{kovachki2021universal}
Nikola Kovachki, Samuel Lanthaler, and Siddhartha Mishra.
\newblock On universal approximation and error bounds for fourier neural
  operators.
\newblock {\em arXiv preprint arXiv:2107.07562}, 2021.

\bibitem{yu2021arbitrary}
Annan Yu, Chlo{\'e} Becquey, Diana Halikias, Matthew~Esmaili Mallory, and Alex
  Townsend.
\newblock Arbitrary-depth universal approximation theorems for operator neural
  networks.
\newblock {\em arXiv preprint arXiv:2109.11354}, 2021.

\bibitem{lu2021learning}
Lu~Lu, Pengzhan Jin, Guofei Pang, Zhongqiang Zhang, and George~Em Karniadakis.
\newblock Learning nonlinear operators via {DeepONet} based on the universal
  approximation theorem of operators.
\newblock {\em Nature Machine Intelligence}, 3(3):218--229, 2021.

\bibitem{kovachki2021neural}
Nikola Kovachki, Zongyi Li, Burigede Liu, Kamyar Azizzadenesheli, Kaushik
  Bhattacharya, Andrew Stuart, and Anima Anandkumar.
\newblock Neural operator: Learning maps between function spaces.
\newblock {\em arXiv preprint arXiv:2108.08481}, 2021.

\bibitem{owhadi2020ideas}
Houman Owhadi.
\newblock Do ideas have shape? plato's theory of forms as the continuous limit
  of artificial neural networks.
\newblock {\em arXiv preprint arXiv:2008.03920}, 2020.

\bibitem{kadri2016operator}
Hachem Kadri, Emmanuel Duflos, Philippe Preux, St{\'e}phane Canu, Alain
  Rakotomamonjy, and Julien Audiffren.
\newblock Operator-valued kernels for learning from functional response data.
\newblock {\em Journal of Machine Learning Research}, 17(20):1--54, 2016.

\bibitem{wang2021learning}
Sifan Wang, Hanwen Wang, and Paris Perdikaris.
\newblock Learning the solution operator of parametric partial differential
  equations with physics-informed {DeepONets}.
\newblock {\em Science Advances}, 7(40):eabi8605, 2021.

\bibitem{wang2021long}
Sifan Wang and Paris Perdikaris.
\newblock Long-time integration of parametric evolution equations with
  physics-informed {DeepONets}.
\newblock {\em arXiv preprint arXiv:2106.05384}, 2021.

\bibitem{glorot2010understanding}
Xavier Glorot and Yoshua Bengio.
\newblock Understanding the difficulty of training deep feedforward neural
  networks.
\newblock In {\em Proceedings of the thirteenth international conference on
  artificial intelligence and statistics}, pages 249--256. JMLR Workshop and
  Conference Proceedings, 2010.

\bibitem{he2016deep}
Kaiming He, Xiangyu Zhang, Shaoqing Ren, and Jian Sun.
\newblock Deep residual learning for image recognition.
\newblock In {\em Proceedings of the IEEE conference on computer vision and
  pattern recognition}, pages 770--778, 2016.

\bibitem{ioffe2015batch}
Sergey Ioffe and Christian Szegedy.
\newblock Batch normalization: Accelerating deep network training by reducing
  internal covariate shift.
\newblock In {\em International conference on machine learning}, pages
  448--456. PMLR, 2015.

\bibitem{salimans2016weight}
Tim Salimans and Durk~P Kingma.
\newblock Weight normalization: A simple reparameterization to accelerate
  training of deep neural networks.
\newblock {\em Advances in neural information processing systems}, 29:901--909,
  2016.

\bibitem{lecun2012efficient}
Yann~A LeCun, L{\'e}on Bottou, Genevieve~B Orr, and Klaus-Robert M{\"u}ller.
\newblock Efficient backprop.
\newblock In {\em Neural networks: Tricks of the trade}, pages 9--48. Springer,
  2012.

\bibitem{di2021deeponet}
P~Clark Di~Leoni, Lu~Lu, Charles Meneveau, George Karniadakis, and Tamer~A
  Zaki.
\newblock {DeepONet} prediction of linear instability waves in high-speed
  boundary layers.
\newblock {\em arXiv preprint arXiv:2105.08697}, 2021.

\bibitem{li2020fourier}
Zongyi Li, Nikola Kovachki, Kamyar Azizzadenesheli, Burigede Liu, Kaushik
  Bhattacharya, Andrew Stuart, and Anima Anandkumar.
\newblock Fourier neural operator for parametric partial differential
  equations.
\newblock {\em arXiv preprint arXiv:2010.08895}, 2020.

\bibitem{wang2021understanding}
Sifan Wang, Yujun Teng, and Paris Perdikaris.
\newblock Understanding and mitigating gradient flow pathologies in
  physics-informed neural networks.
\newblock {\em SIAM Journal on Scientific Computing}, 43(5):A3055--A3081, 2021.

\bibitem{wang2020and}
Sifan Wang, Xinling Yu, and Paris Perdikaris.
\newblock When and why {PINNs} fail to train: A neural tangent kernel
  perspective.
\newblock {\em arXiv preprint arXiv:2007.14527}, 2020.

\bibitem{mcclenny2020self}
Levi McClenny and Ulisses Braga-Neto.
\newblock Self-adaptive physics-informed neural networks using a soft attention
  mechanism.
\newblock {\em arXiv preprint arXiv:2009.04544}, 2020.

\bibitem{wang2021deep}
Sifan Wang and Paris Perdikaris.
\newblock Deep learning of free boundary and stefan problems.
\newblock {\em Journal of Computational Physics}, 428:109914, 2021.

\bibitem{jacot2018neural}
Arthur Jacot, Franck Gabriel, and Cl{\'e}ment Hongler.
\newblock Neural tangent kernel: Convergence and generalization in neural
  networks.
\newblock In {\em Advances in neural information processing systems}, pages
  8571--8580, 2018.

\bibitem{du2019gradient}
Simon Du, Jason Lee, Haochuan Li, Liwei Wang, and Xiyu Zhai.
\newblock Gradient descent finds global minima of deep neural networks.
\newblock In {\em International Conference on Machine Learning}, pages
  1675--1685. PMLR, 2019.

\bibitem{allen2019convergence}
Zeyuan Allen-Zhu, Yuanzhi Li, and Zhao Song.
\newblock A convergence theory for deep learning via over-parameterization.
\newblock In {\em International Conference on Machine Learning}, pages
  242--252. PMLR, 2019.

\bibitem{cao2019towards}
Yuan Cao, Zhiying Fang, Yue Wu, Ding-Xuan Zhou, and Quanquan Gu.
\newblock Towards understanding the spectral bias of deep learning.
\newblock {\em arXiv preprint arXiv:1912.01198}, 2019.

\bibitem{xu2019frequency}
Zhi-Qin~John Xu, Yaoyu Zhang, Tao Luo, Yanyang Xiao, and Zheng Ma.
\newblock Frequency principle: {F}ourier analysis sheds light on deep neural
  networks.
\newblock {\em arXiv preprint arXiv:1901.06523}, 2019.

\bibitem{rahaman2019spectral}
Nasim Rahaman, Aristide Baratin, Devansh Arpit, Felix Draxler, Min Lin, Fred
  Hamprecht, Yoshua Bengio, and Aaron Courville.
\newblock On the spectral bias of neural networks.
\newblock In {\em International Conference on Machine Learning}, pages
  5301--5310, 2019.

\bibitem{lee2019wide}
Jaehoon Lee, Lechao Xiao, Samuel Schoenholz, Yasaman Bahri, Roman Novak, Jascha
  Sohl-Dickstein, and Jeffrey Pennington.
\newblock Wide neural networks of any depth evolve as linear models under
  gradient descent.
\newblock In {\em Advances in neural information processing systems}, pages
  8572--8583, 2019.

\bibitem{wang2020eigenvector}
Sifan Wang, Hanwen Wang, and Paris Perdikaris.
\newblock On the eigenvector bias of {F}ourier feature networks: From
  regression to solving multi-scale {PDEs} with physics-informed neural
  networks.
\newblock {\em arXiv preprint arXiv:2012.10047}, 2020.

\bibitem{chen1995universal}
Tianping Chen and Hong Chen.
\newblock Universal approximation to nonlinear operators by neural networks
  with arbitrary activation functions and its application to dynamical systems.
\newblock {\em IEEE Transactions on Neural Networks}, 6(4):911--917, 1995.

\bibitem{baydin2018automatic}
Atilim~Gunes Baydin, Barak~A Pearlmutter, Alexey~Andreyevich Radul, and
  Jeffrey~Mark Siskind.
\newblock Automatic differentiation in machine learning: a survey.
\newblock {\em Journal of machine learning research}, 18, 2018.

\bibitem{cai2020deepm}
Shengze Cai, Zhicheng Wang, Lu~Lu, Tamer~A Zaki, and George~Em Karniadakis.
\newblock Deepm\&mnet: Inferring the electroconvection multiphysics fields
  based on operator approximation by neural networks.
\newblock {\em arXiv preprint arXiv:2009.12935}, 2020.

\bibitem{iserles2009first}
Arieh Iserles.
\newblock {\em A first course in the numerical analysis of differential
  equations}.
\newblock Number~44. Cambridge university press, 2009.

\bibitem{kingma2014adam}
Diederik~P Kingma and Jimmy Ba.
\newblock Adam: A method for stochastic optimization.
\newblock {\em arXiv preprint arXiv:1412.6980}, 2014.

\bibitem{fort2020deep}
Stanislav Fort, Gintare~Karolina Dziugaite, Mansheej Paul, Sepideh Kharaghani,
  Daniel~M Roy, and Surya Ganguli.
\newblock Deep learning versus kernel learning: an empirical study of loss
  landscape geometry and the time evolution of the neural tangent kernel.
\newblock {\em arXiv preprint arXiv:2010.15110}, 2020.

\bibitem{leclerc2020two}
Guillaume Leclerc and Aleksander Madry.
\newblock The two regimes of deep network training.
\newblock {\em arXiv preprint arXiv:2002.10376}, 2020.

\bibitem{schoenholz2016deep}
Samuel~S Schoenholz, Justin Gilmer, Surya Ganguli, and Jascha Sohl-Dickstein.
\newblock Deep information propagation.
\newblock {\em arXiv preprint arXiv:1611.01232}, 2016.

\bibitem{yang2018physics}
Yibo Yang and Paris Perdikaris.
\newblock Physics-informed deep generative models.
\newblock {\em arXiv preprint arXiv:1812.03511}, 2018.

\bibitem{driscoll2014chebfun}
Tobin~A Driscoll, Nicholas Hale, and Lloyd~N Trefethen.
\newblock Chebfun guide, 2014.

\bibitem{cox2002exponential}
Steven~M Cox and Paul~C Matthews.
\newblock Exponential time differencing for stiff systems.
\newblock {\em Journal of Computational Physics}, 176(2):430--455, 2002.

\bibitem{alnaes2015fenics}
Martin Aln{\ae}s, Jan Blechta, Johan Hake, August Johansson, Benjamin Kehlet,
  Anders Logg, Chris Richardson, Johannes Ring, Marie~E Rognes, and Garth~N
  Wells.
\newblock The fenics project version 1.5.
\newblock {\em Archive of Numerical Software}, 3(100), 2015.

\bibitem{shin2020convergence}
Yeonjong Shin, J{\'e}r{\^o}me Darbon, and George~Em Karniadakis.
\newblock On the convergence of physics informed neural networks for linear
  second-order elliptic and parabolic type {PDEs}.
\newblock 2020.

\bibitem{mishra2020estimates}
Siddhartha Mishra and Roberto Molinaro.
\newblock Estimates on the generalization error of physics informed neural
  networks (pinns) for approximating pdes.
\newblock {\em arXiv preprint arXiv:2006.16144}, 2020.

\bibitem{mitusch2019dolfin}
Sebastian~K Mitusch, Simon~W Funke, and J{\o}rgen~S Dokken.
\newblock dolfin-adjoint 2018.1: automated adjoints for fenics and firedrake.
\newblock {\em Journal of Open Source Software}, 4(38):1292, 2019.

\bibitem{jax2018github}
James Bradbury, Roy Frostig, Peter Hawkins, Matthew~James Johnson, Chris Leary,
  Dougal Maclaurin, George Necula, Adam Paszke, Jake Vander{P}las, Skye
  Wanderman-{M}ilne, and Qiao Zhang.
\newblock {JAX}: composable transformations of {P}ython+{N}um{P}y programs,
  2018.

\bibitem{hunter2007matplotlib}
John~D Hunter.
\newblock Matplotlib: A {2D} graphics environment.
\newblock {\em IEEE Annals of the History of Computing}, 9(03):90--95, 2007.

\bibitem{harris2020array}
Charles~R Harris, K~Jarrod Millman, St{\'e}fan~J van~der Walt, Ralf Gommers,
  Pauli Virtanen, David Cournapeau, Eric Wieser, Julian Taylor, Sebastian Berg,
  Nathaniel~J Smith, et~al.
\newblock Array programming with numpy.
\newblock {\em Nature}, 585(7825):357--362, 2020.

\end{thebibliography}

\clearpage

\appendix

\section{Nomenclature}
Table \ref{tab: Notations} summarizes the main symbols and notation used in this work.

\begin{table}[h]
\renewcommand{\arraystretch}{1.4}
    \centering
    \begin{tabular}{ll} 
    \Xhline{3\arrayrulewidth} 
    Notation     & Description \\
    \Xhline{3\arrayrulewidth} 
        $\bm{u}(\cdot)$ & an input function \\  
        $\bm{s}(\cdot)$ & a solution to a parametric PDE \\
        $G$         & an operator \\
        $G_{\bm{\theta}}$  &  an unstacked DeepONet representation of the operator $G$ \\
        $\bm{\theta}$ &  all trainable parameters of a DeepONet \\
        $\{\bm{x}_i\}_{i=1}^m$  & $m$ sensor points where input functions $\bm{u}(\bm{x})$ are evaluated\\
        $[u(\bm{x}_1), u(\bm{x}_2), \dots, u(\bm{x}_m)]$ & an input of the branch net, representing the input function $u$ \\ 
        $\bm{y}$          & an input of the trunk net, a point in the domain of $G(u)$ \\
        $N$                & number of input samples in the training data-set \\
        $m$              &  number of locations for evaluating the input functions $u$        \\
        $P$               &  number of locations for evaluating the output functions $G(u)$        \\
        $Q$                & number of collocation points for evaluating the PDE residual \\
        PDE & Partial differential equation \\ 
        GRF             &  a Gaussian random field                   \\
        SDF             & a signed distance function \\
        MLP & Multi-layer perceptron \\
        $l$              & length scale of a Gaussian random field   \\
          $k$              & output scale of a Gaussian random field   \\
        $\bm{K}$  &  Neural Tangent Kernel of an MLP  \\
        $\bm{H}$  &  Neural Tangent Kernel of a physics-informed DeepONet
\\
    \Xhline{3\arrayrulewidth}
    \end{tabular}
    \caption{{\em Nomenclature}: Summary of the main symbols and notation used in this work.}
    \label{tab: Notations}
\end{table}

\clearpage
\section{Hyper-parameter settings}\label{app: hp_settings}

\begin{table}[h]
\renewcommand{\arraystretch}{1.4}
    \centering
    \begin{tabular}{c|ccccccc}
        \Xhline{3\arrayrulewidth}
      Case   & Input function space &  m & P & Q & \# u Train & \# u Test & Iterations  \\
      \hline
     Anti-derivative  & GRF$(l=0.2, k \in [10^{-2}, 10^2])$ & $10^2$ & 1 & - & $10^4$ & $10^3$  & $4 \times 10^{4}$ \\
    Advection equation & GRF$(l=0.2)$ & $10^2$ & $10^2$ & $2 \times 10^3$ & $10^3$ & $10^2$ & $3 \times 10^{5}$\\
    Burger's equation & $ \mathcal{N}\left(0,  25^2(-\Delta+5^2 I)^{-4}\right) $ & $10^2$ & $10^2$ & $2 \times 10^3$  & $10^3$  & $10^2$   &  $2 \times 10^{5}$\\\
        Stokes equation &  Ellipse & $10^2$ & $10^2$ & $2 \times 10^3$ & $  10^3$ & $10^2$ & $2 \times 10^{5}$ \\
    \Xhline{3\arrayrulewidth}
    \end{tabular}
    \caption{Default hyper-parameter settings for each benchmark employed in this work  (unless otherwise stated).}
    \label{tab: parameters_case}
\end{table}

\begin{table}[h]
\renewcommand{\arraystretch}{1.4}
    \centering
    \begin{tabular}{c|ccccccc}
        \Xhline{3\arrayrulewidth}
        Case   & Architecture &  Trunk depth &  Trunk width & Branch depth & Branch width  \\
        \hline
        \multirow{3}{*}{Advection equation} &  MLP & 7 & 100 & 7 & 100 \\
                &  Modified MLP & 7 & 100 & 7 & 100 \\
                 & Modified DeepONet & 7 & 100 & 7 & 100 \\
           \Xhline{3\arrayrulewidth}
           \multirow{3}{*}{Burger's equation} &  MLP & 7 & 100 & 7 & 100 \\
        &  Modified MLP & 7 & 100 & 7 & 100 \\
         & Modified DeepONet & 7 & 100 & 7 & 100 \\
   \Xhline{3\arrayrulewidth}
           \multirow{3}{*}{Stokes equation} &  MLP & 7 & 100 & 7 & 100 \\
        &  Modified MLP & 7 & 100 & 7 & 100 \\
         & Modified DeepONet & 7 & 100 & 7 & 100 \\
   \Xhline{3\arrayrulewidth}
    \end{tabular}
    \caption{Physics-informed DeepONet architectures for each benchmark employed in this work (unless otherwise stated). }
    \label{tab: Physics_informed_DeepONet_size}
\end{table}

\clearpage

\section{Computational cost}
\label{sec: computational_cost}

{\bf Training:} Table \ref{sec: computational_cost} summarizes the computational cost  (hours) of training physics-informed DeepONet models with different network architectures and weighting schemes. 
The size of different models as well as network architectures are listed Table \ref{tab: Physics_informed_DeepONet_size}, respectively. All networks are trained using a single NVIDIA RTX A6000 graphics card.

\begin{table}[h]
\renewcommand{\arraystretch}{1.4}
    \centering
    \begin{tabular}{c|c|c| c}
     \Xhline{3\arrayrulewidth}
       Case  &  Architecture & Weighting scheme & Training time (hours)  \\
       \hline
      \multirow{6}{*}{ Advection equation} & \multirow{2}{*}{MLP} & Fixed weights &  1.98 \\
                                                &                       & NTK weights  & 4.63  \\
                                                & \multirow{2}{*}{Modified MLP} & Fixed weights & 2.88  \\
                                                &                       & NTK weights  &  7.65  \\
                                                & \multirow{2}{*}{Modified DeepONet} & Fixed weights &  3.25 \\
                                                &                       & NTK weights  &  7.51 \\
    \hline
    \multirow{6}{*}{ Burger's equation} & \multirow{2}{*}{MLP} & Fixed weights & 2.21  \\
                                                &                       & NTK weights  &  7.00  \\
                                                & \multirow{2}{*}{Modified MLP} & Fixed weights  & 3.05  \\
                                                &                       & NTK weights  & 9.82  \\
                                                & \multirow{2}{*}{Modified DeepONet} & Fixed weights & 3.88   \\
                                                &                       & NTK weights  & 9.49  \\
    \hline
 \multirow{6}{*}{ Stokes equation} & \multirow{2}{*}{MLP} & Fixed weights &  2.36  \\
                                                &                       & NTK weights  & 14.38  \\
                                                & \multirow{2}{*}{Modified MLP} & Fixed weights & 4.67  \\
                                                &                       & NTK weights  &   20.86 \\
                                                & \multirow{2}{*}{Modified DeepONet} & Fixed weights & 5.60   \\
                                                &                       & NTK weights  &  20.58 \\
    \Xhline{3\arrayrulewidth}
    \end{tabular}
    \caption{Computational cost (hours) for training physics-informed DeepONet models across the different becnhmarks and architectures employed in this work. Reported timings are obtained on a single NVIDIA RTX A6000 graphics card.}
    \label{tab: computational_cost}
\end{table}

\clearpage

\section{Proof of Lemma \ref{prop: deeponet_NTK_property}}
\label{proof: deeponet_NTK_property}
\begin{proof}
First we observe that $\bm{H}(\bm{\theta})$ is a Gram matrix. Indeed, we define 
\begin{align}
    v_i = T^{(i)}(u^{(i)}(\bm{x}_i), G_{\bm{\theta}}(\bm{u}^{(i)})(\bm{y}_i)), \quad i=1.2, \dots, N^*.
\end{align}
Then by the definition of $\bm{H}(\bm{\theta})$, we have
\begin{align}
    \bm{H}_{ij}(\bm{\theta}) = \langle v_i, v_j \rangle
\end{align}

\begin{enumerate}[label=(\alph*)]
    \item By the definition of a Gram matrix.
    \item Let $\|\bm{H}(\bm{\theta})\|_\infty =  \langle v_i, v_j \rangle$ for some $i, j$ and $\langle v_k, v_k \rangle = \max_{1 \leq k \leq N^*} \bm{H}_{kk}(\bm{\theta}) $/ for some $k$. Then we have
    \begin{align}
        \max_{1 \leq k \leq N^*} \bm{H}_{kk}(\bm{\theta}) = \langle v_k, v_k \rangle \leq \|\bm{H}(\bm{\theta})\|_\infty =  \langle v_i, v_j \rangle \leq \|v_i\| \|v_j\| \leq \|v_k\|^2 = \langle v_k, v_k \rangle
    \end{align}
Therefore, we have 
\begin{align}
         \|\bm{H}(\bm{\theta})\|_\infty = \max_{1\leq k \leq N^*} \bm{H}_{kk}(\bm{\theta})
\end{align}
\end{enumerate}

\end{proof}

\clearpage

\clearpage
\section{Proof of Lemma \ref{lemma: ODE}}\label{proof: ODE}

\begin{proof}
Recall the definition of the loss function \ref{eq: rearranged_loss},
\begin{align}
    \mathcal{L}(\bm{\theta}) = \frac{2}{N^*} \sum_{i=1}^{N^*} \left| T^{(i)}(u^{(i)}(\bm{x}_i), G_{\bm{\theta}}(\bm{u}^{(i)})(\bm{y}_i))\right|^2
\end{align}
Now we consider the corresponding gradient flow
\begin{align}
    \frac{d \bm{\theta}}{ dt} &= - \nabla \mathcal{L}(\bm{\theta}) \\
                              &= - \frac{4}{N^*} \sum_{k=1}^{N^*} \left( T^{(i)}(u^{(i)}(\bm{x}_i), G_{\bm{\theta}}(\bm{u}^{(i)})(\bm{y}_i)) \right) \frac{d  T^{(i)}(u^{(i)}(\bm{x}_i), G_{\bm{\theta}}(\bm{u}^{(i)})(\bm{y}_i)) }{d \bm{\theta}}
\end{align}
For $1 \leq j \leq N^*$, note that
\begin{align}
    &\frac{d T^{(j)}(u^{(j)}(\bm{x}_j), G_{\bm{\theta}}(\bm{u}^{(j)})(\bm{y}_j))}{ dt } \\
    &=   \frac{d T^{(j)}(u^{(j)}(\bm{x}_j), G_{\bm{\theta}}(\bm{u}^{(j)})(\bm{y}_j))}{ d \bm{\theta} } \cdot  \frac{d \bm{\theta}}{ dt } \\
    &= \frac{d T^{(j)}(u^{(j)}(\bm{x}_j), G_{\bm{\theta}}(\bm{u}^{(j)})(\bm{y}_j))}{ d \bm{\theta} } \cdot \left[ - \frac{4}{N^*} \sum_{i=1}^{N^*} \left( T^{(i)}(u^{(i)}(\bm{x}_i), G_{\bm{\theta}}(\bm{u}^{(i)})(\bm{y}_i)) \right) \frac{d  T^{(i)}(u^{(i)}(\bm{x}_i), G_{\bm{\theta}}(\bm{u}^{(i)})(\bm{y}_i)) }{d \bm{\theta}} \right] \\
    &= - \frac{4}{ N^*} \sum_{i=1}^{N^*} \left( T^{(i)}(u^{(i)}(\bm{x}_i), G_{\bm{\theta}}(\bm{u}^{(i)})(\bm{y}_i)) \right) \left\langle  \frac{d  T^{(i)}(u^{(i)}(\bm{x}_i), G_{\bm{\theta}}(\bm{u}^{(i)})(\bm{y}_i)) }{d \bm{\theta}}, \frac{d T^{(j)}(u^{(j)}(\bm{x}_j), G_{\bm{\theta}}(\bm{u}^{(j)})(\bm{y}_j))}{ d \bm{\theta} } 
    \right\rangle
\end{align}
By Definition \ref{def: deeponet_ntk}, \ref{def: notation}, we conclude that
\begin{align}
    \frac{d \bm{T}\left(\bm{U}(\bm{X}), G_{\bm{\theta}(t)} (\bm{U}) (\bm{Y}) )\right)   }{ d t} =- \frac{4}{ N^*} \bm{K}(\bm{\theta}) \cdot \bm{T}(\bm{U}(\bm{X}), G_{\bm{\theta}(t)} (\bm{U}) (\bm{Y}) ))
\end{align}
where $\bm{K}$ is a $N^* \times N^*$ matrix whose entries are given by  
\begin{align}
     \bm{K}_{ij}(\bm{\theta}) =  \left\langle \frac{d T^{(i)} \left(u^{(i)} (\bm{x}_i), G_{\bm{\theta}} (\bm{u}^{(i)})(\bm{y}_i)  \right)  }{d \bm{\theta}}  , \frac{d T^{(j)} \left(u^{(j)} (\bm{x}_j), G_{\bm{\theta}} (\bm{u}^{(j)})(\bm{y}_j) \right)  }{d \bm{\theta}}                  \right\rangle
\end{align}

\end{proof}

\clearpage
\section{Antiderivative}

\begin{figure}[h]
    \centering
    \includegraphics[width=0.4\textwidth]{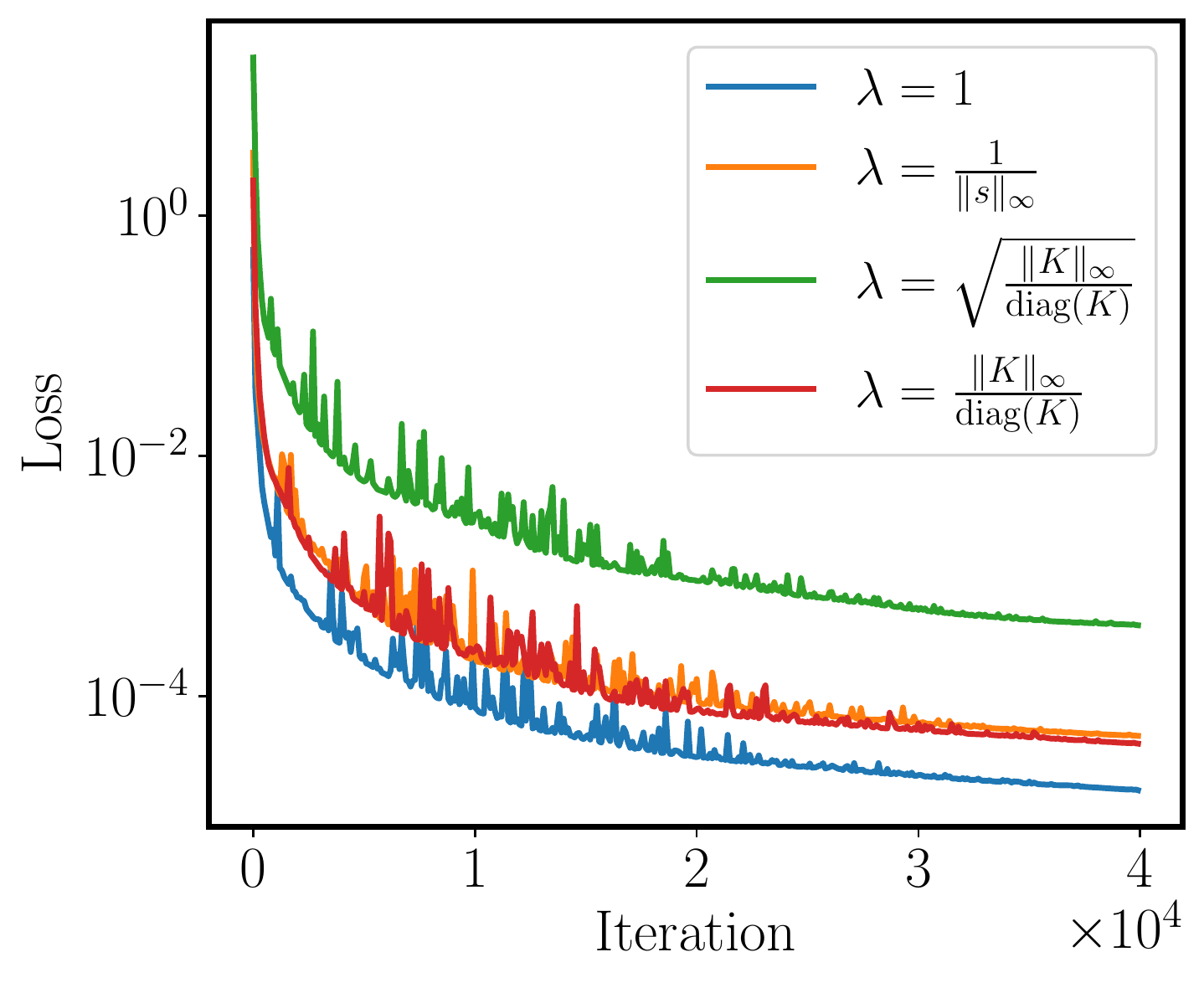}
    \caption{{\em Anti-derivative operator:}  Training loss convergence of a DeepONet models using different loss schemes for $4 \times 10^4$ 40,000 iterations of gradient descent using the Adam optimizer. Here we remark that all losses are weighted, and, therefore, the magnitude of these losses is not informative.}
    \label{fig: ODE_losses}
\end{figure}

\clearpage
\section{Advection equation}

\begin{figure}[h]
    \centering
    \includegraphics[width=0.9\textwidth]{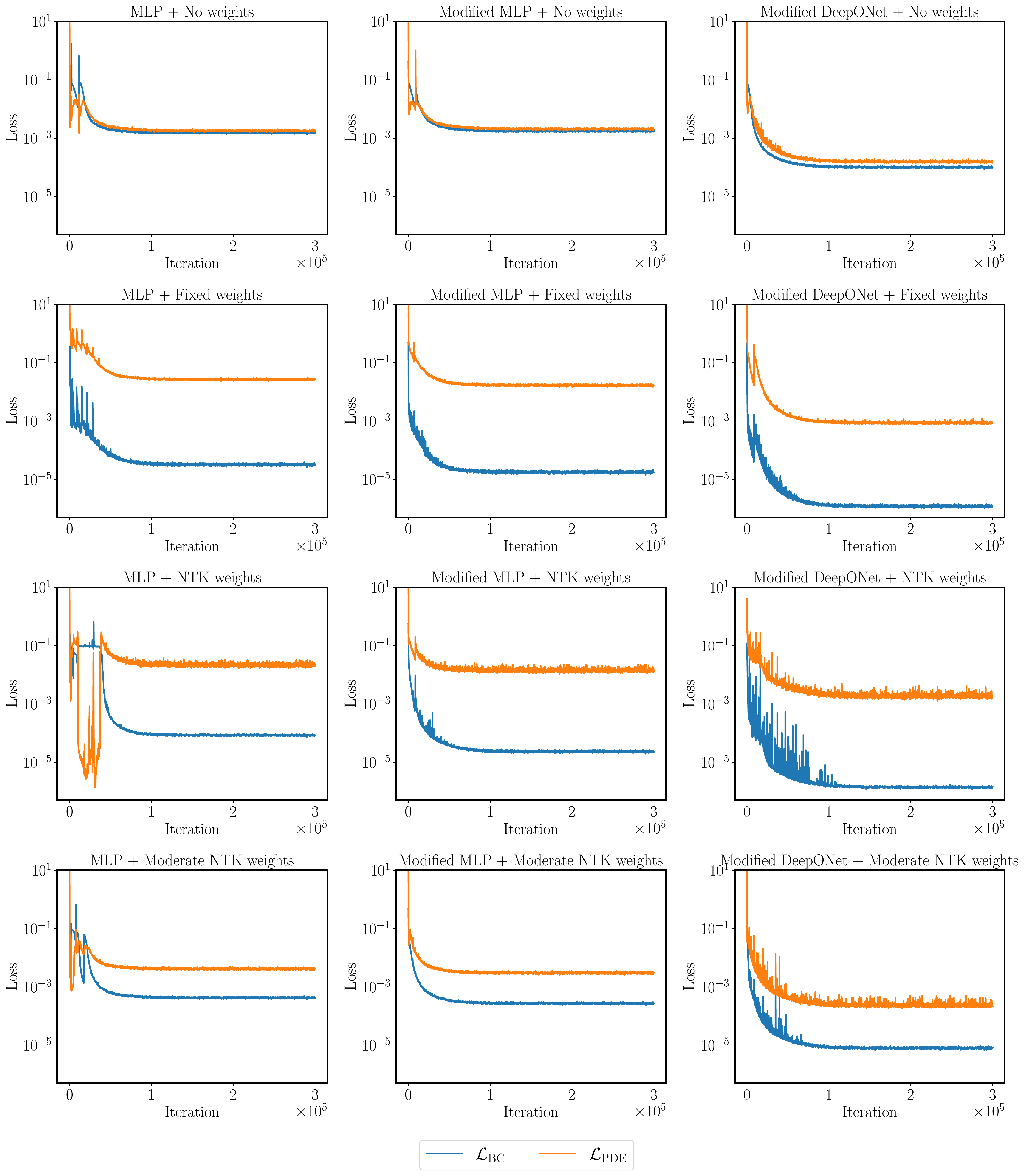}
    \caption{{\em Advection equation:}  Training loss convergence of a DeepONet models using different DeepONet architectures and weighting schemes for $3 \times 10^5$ iterations of gradient descent using the Adam optimizer. Here we remark that all losses are unweighted.}
    \label{fig: ADV_losses}
\end{figure}

\begin{figure}
    \centering
    \includegraphics[width=0.4\textwidth]{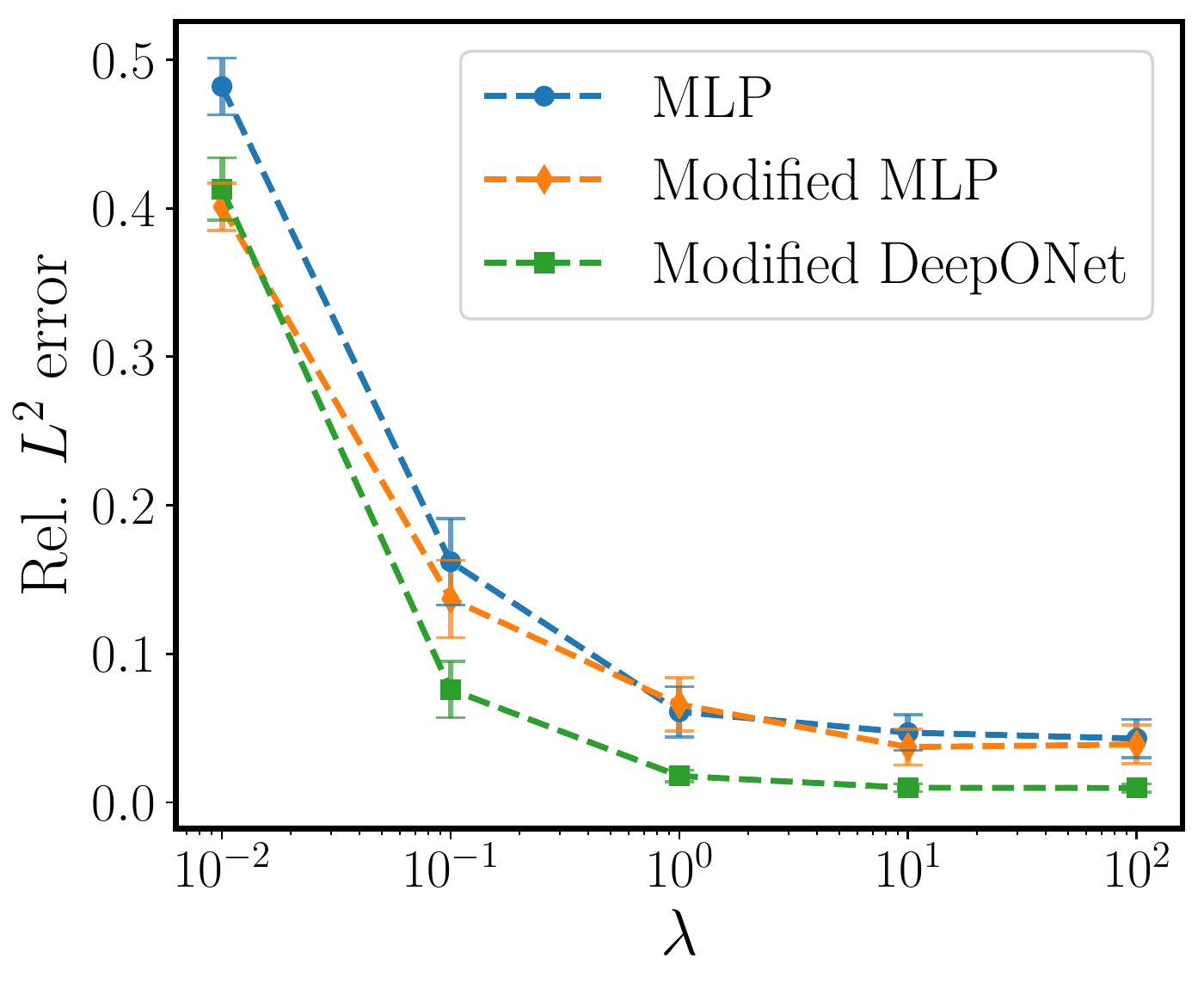}
    \caption{{\em Advection equation:}  Relative $L^2$ error of physics-informed DeepONets represented by different architectures and trained using different fixed weights $\lambda_{\text{bc}} = \lambda_{\text{ic}} = \lambda \in [10^{-2}, 10^{2}]$, averaged over the test data-set. }
    \label{fig: ADV_fixed_weights_error}
\end{figure}

\begin{figure}
    \centering
    \includegraphics[width=0.9\textwidth]{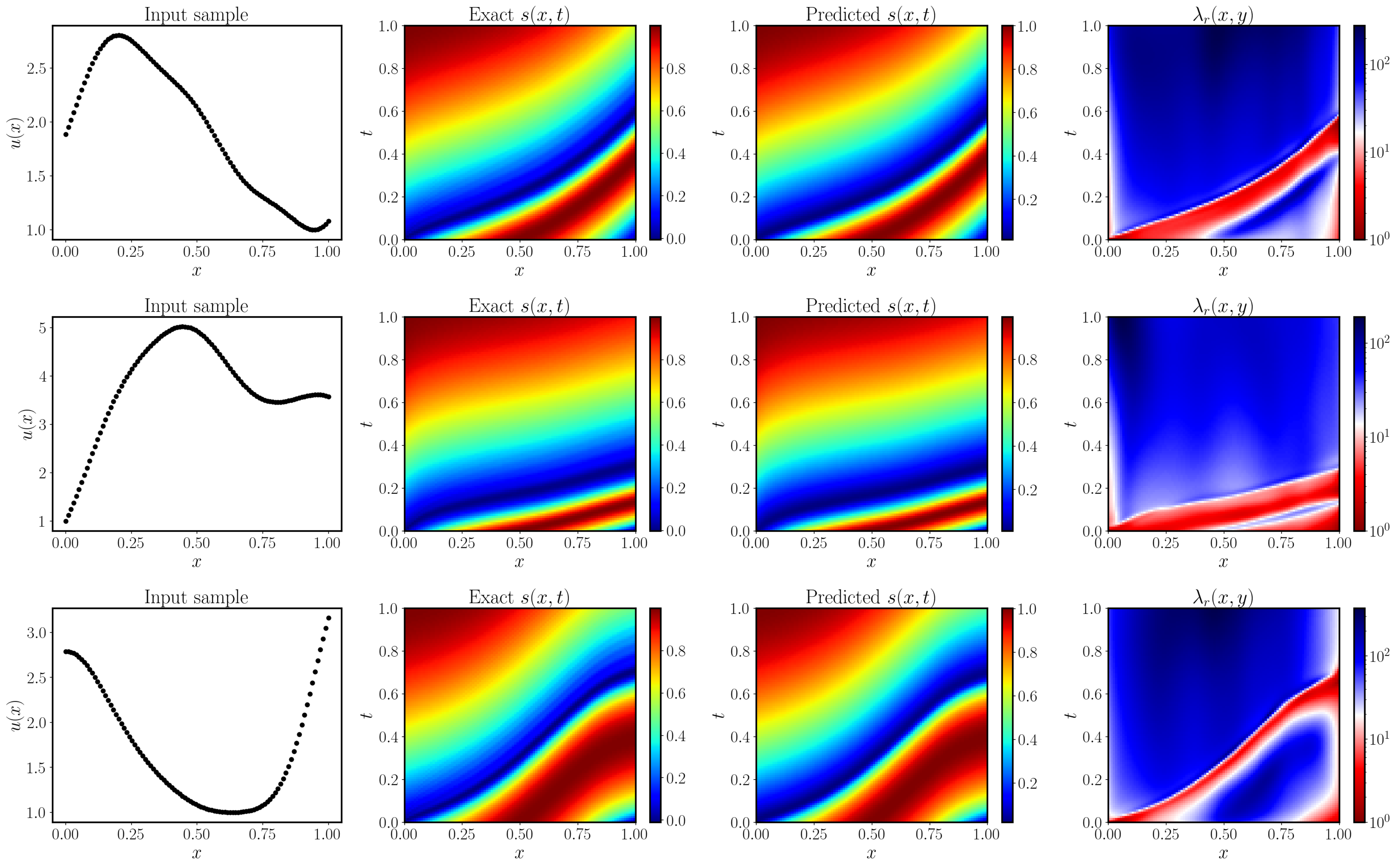}
    \caption{{\em Advection equation:} Predicted solution of the best trained physics-informed DeepONet for three different examples in the test data-set.}
    \label{fig: ADV_examples}
\end{figure}

\clearpage
\section{Burger's equation}

\begin{figure}[h]
    \centering
    \includegraphics[width=0.9\textwidth]{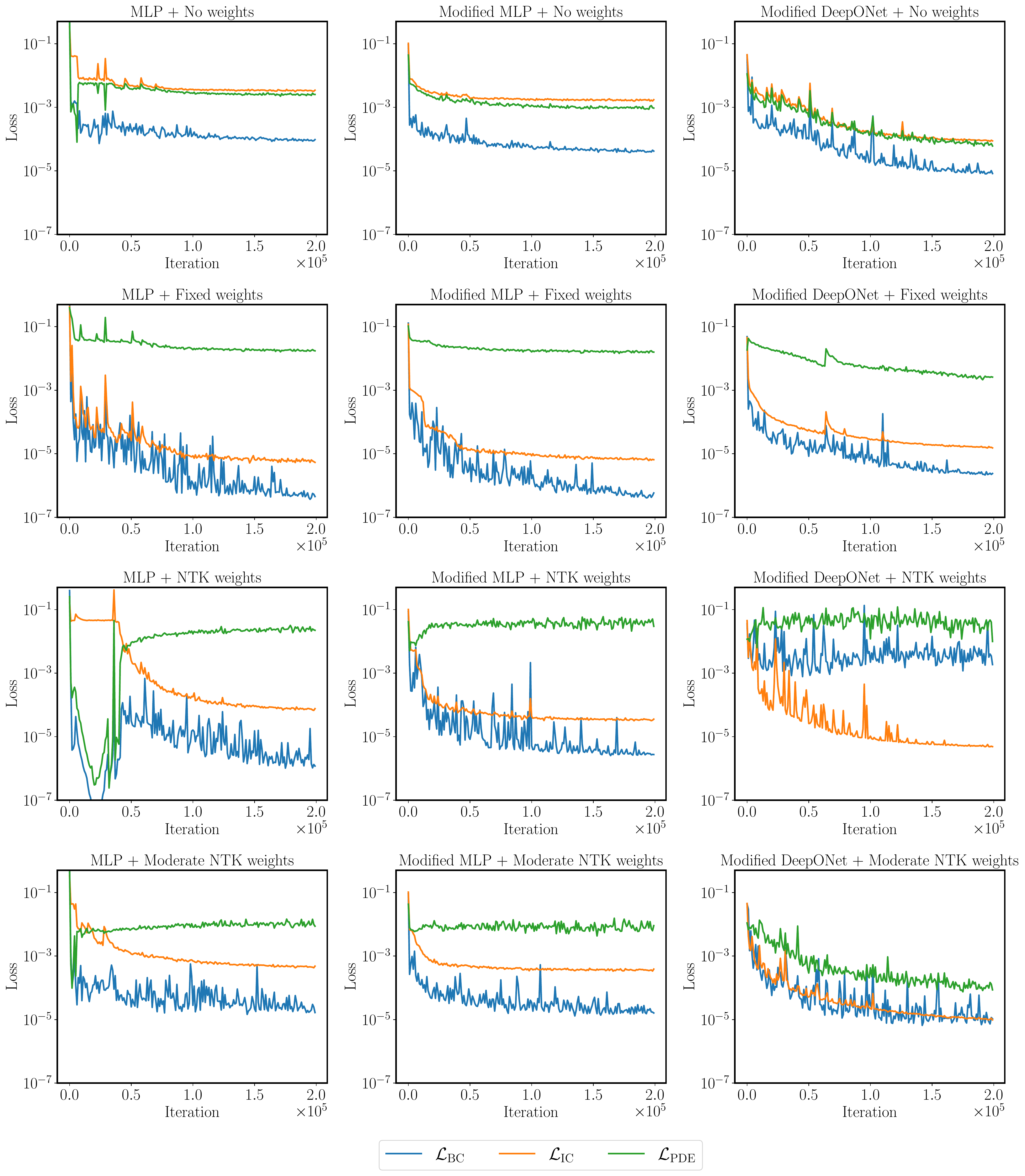}
    \caption{{\em Burger's equation:}  Training loss convergence of a DeepONet models using different DeepONet architectures and weighting schemes for $2 \times 10^5$ iterations of gradient descent using the Adam optimizer. Here we remark that all losses are unweighted.}
    \label{fig: Burger_losses}
\end{figure}

\begin{figure}[h]
    \centering
    \includegraphics[width=0.5\textwidth]{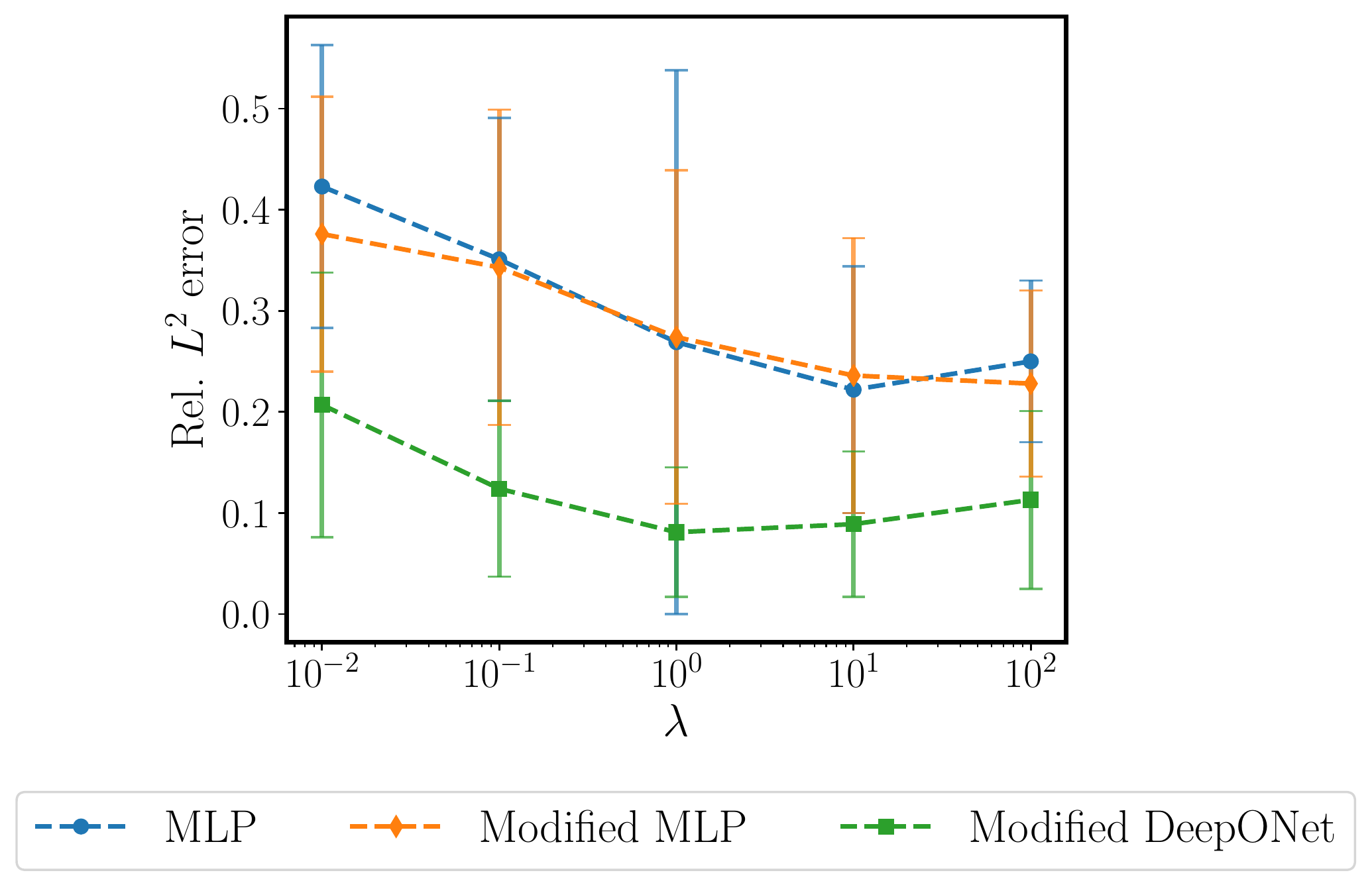}
    \caption{{\em Burger's equation:}  Relative $L^2$ error of physics-informed DeepONets represented by different architectures and trained using different fixed weights $\lambda_{\text{bc}} = \lambda_{\text{ic}} = \lambda \in [10^{-2}, 10^{2}]$, averaged over the test data-set. }
    \label{fig: Burger_fixed_weights_error}
\end{figure}

\begin{figure}[h]
    \centering
    \includegraphics[width=0.9\textwidth]{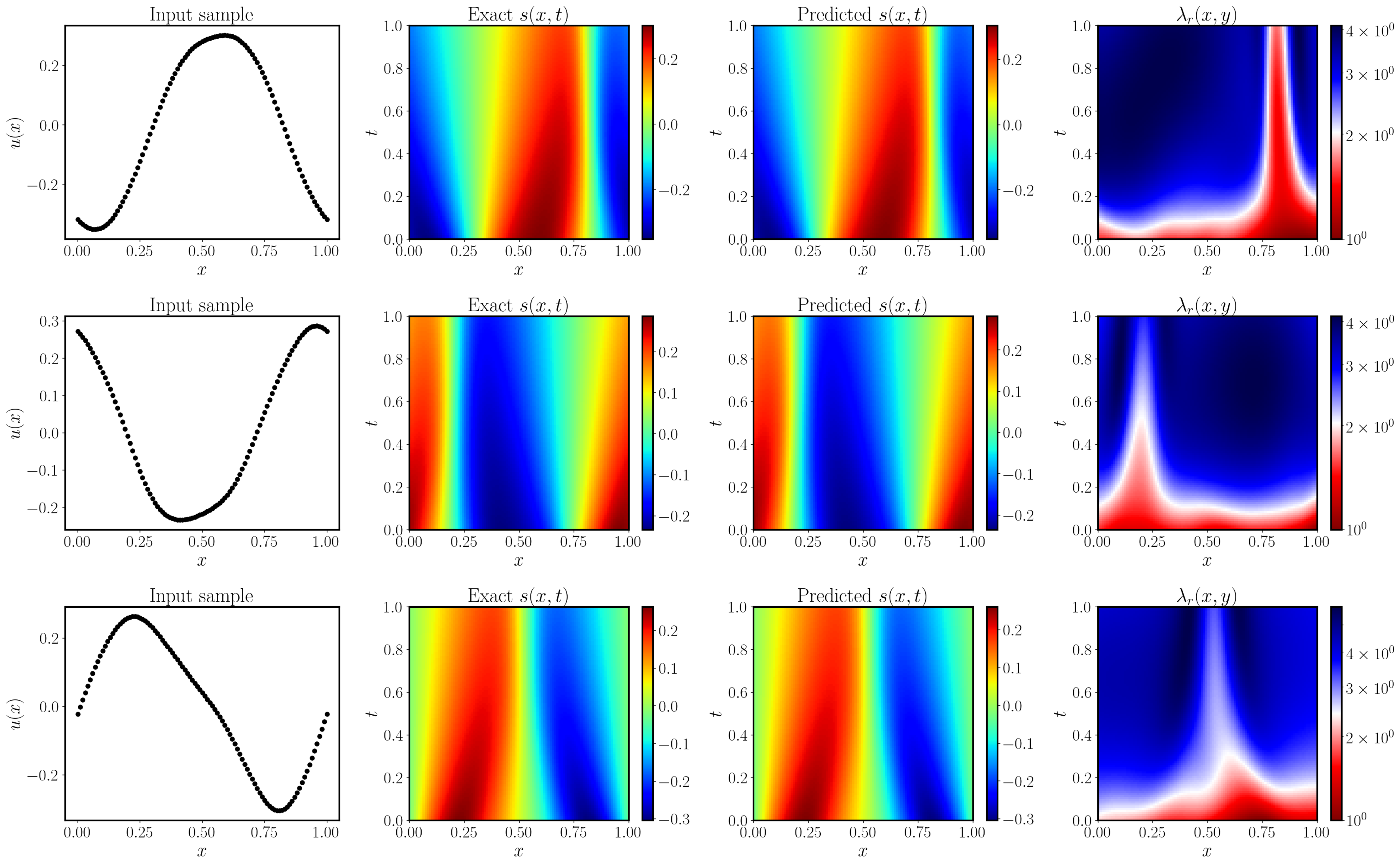}
    \caption{{\em Burger's equation ($\nu = 0.01$):} Predicted solution of the best trained physics-informed DeepONet for three different examples in the test data-set.}
    \label{fig: Buger_01_examples}
\end{figure}

\begin{figure}[h]
    \centering
    \includegraphics[width=0.9\textwidth]{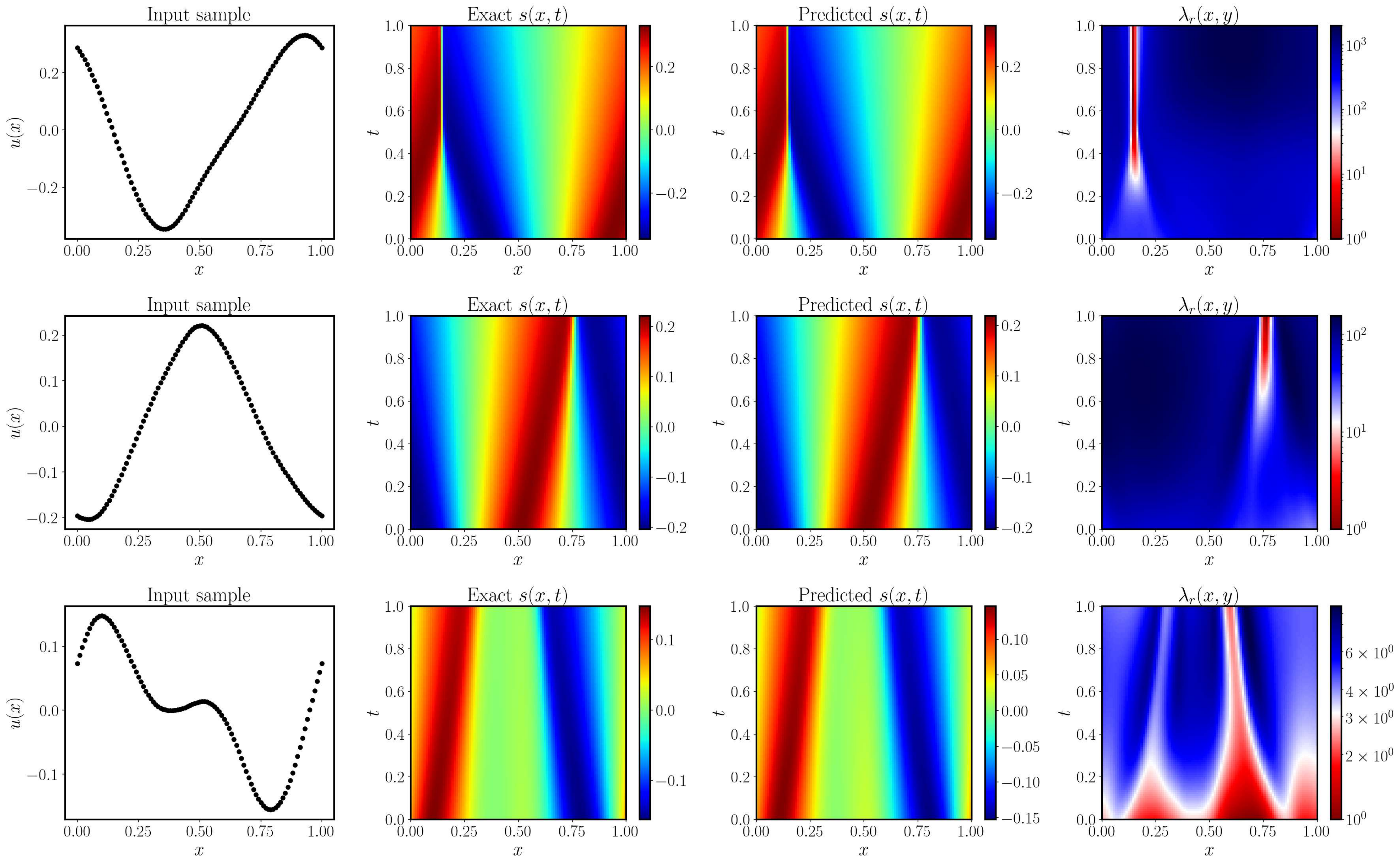}
    \caption{{\em Burger's equation ($\nu = 0.001$):} Predicted solution of the best trained physics-informed DeepONet for three different examples in the test data-set.}
    \label{fig: Burger_001_examples}
\end{figure}

\begin{figure}[h]
    \centering
    \includegraphics[width=0.9\textwidth]{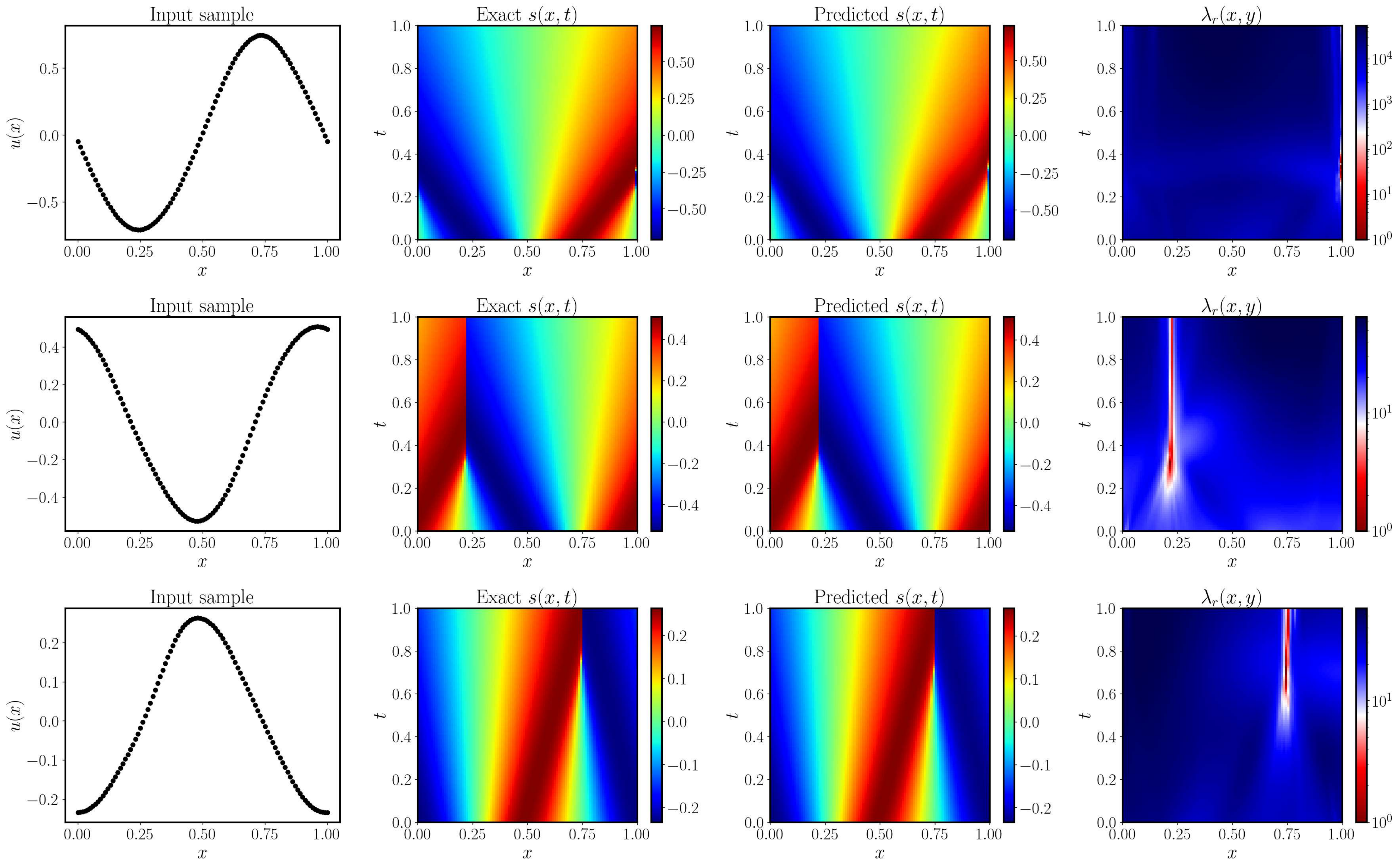}
    \caption{{\em Burger's equation ($\nu = 0.0001$):} Predicted solution of the best trained physics-informed DeepONet for three different examples in the test data-set.}
    \label{fig: Buger_0001_examples}
\end{figure}

\clearpage
\section{Stokes flow}

\begin{figure}[h]
    \centering
    \includegraphics[width=0.9\textwidth]{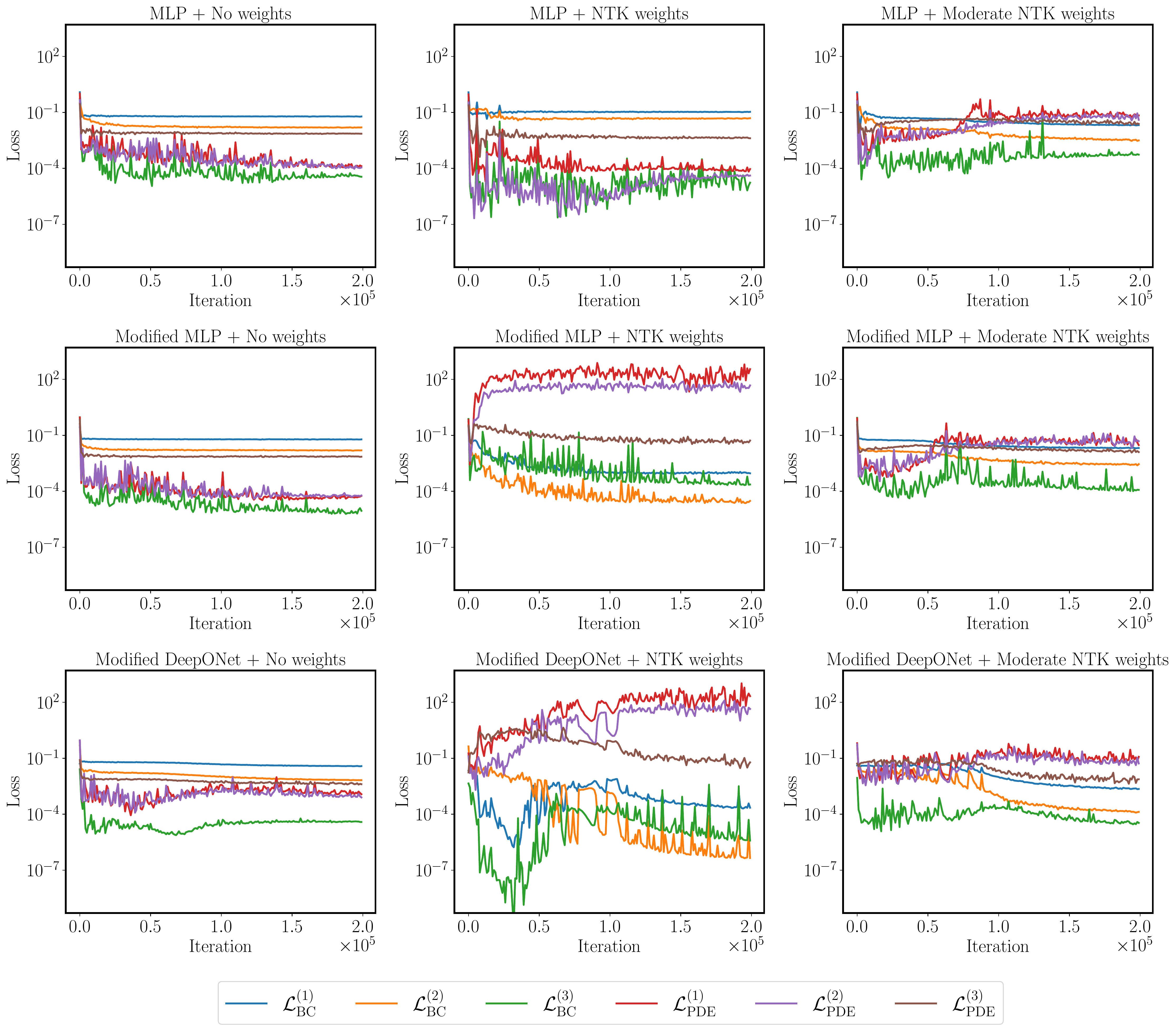}
    \caption{{\em Stokes equation:}  Training loss convergence of a DeepONet models using different DeepONet architectures and weighting schemes for $2 \times 10^5$ iterations of gradient descent using the Adam optimizer. Here we remark that all losses are unweighted.}
    \label{fig: Stokes_losses}
\end{figure}

\begin{figure}[h]
    \centering
    \includegraphics[width=0.8\textwidth]{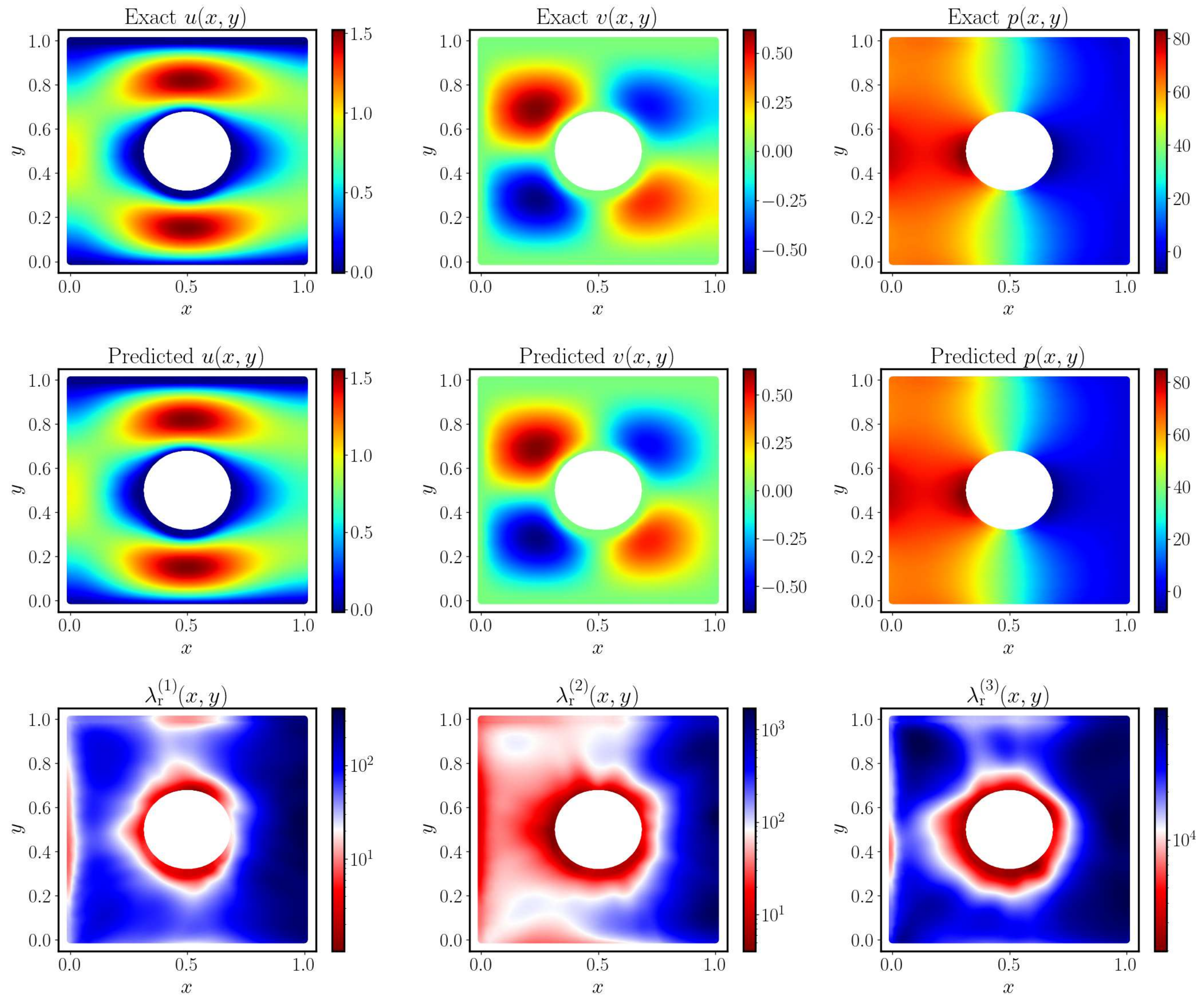}
    \caption{{\em Stokes equation:} Predicted solution of the best trained physics-informed DeepONet for one example in the test data-set.}
    \label{fig: Stokes_PI_deeponet_modified_deeponet_local_NTK_weights_pred_5}
\end{figure}

\begin{figure}[h]
    \centering
    \includegraphics[width=0.8\textwidth]{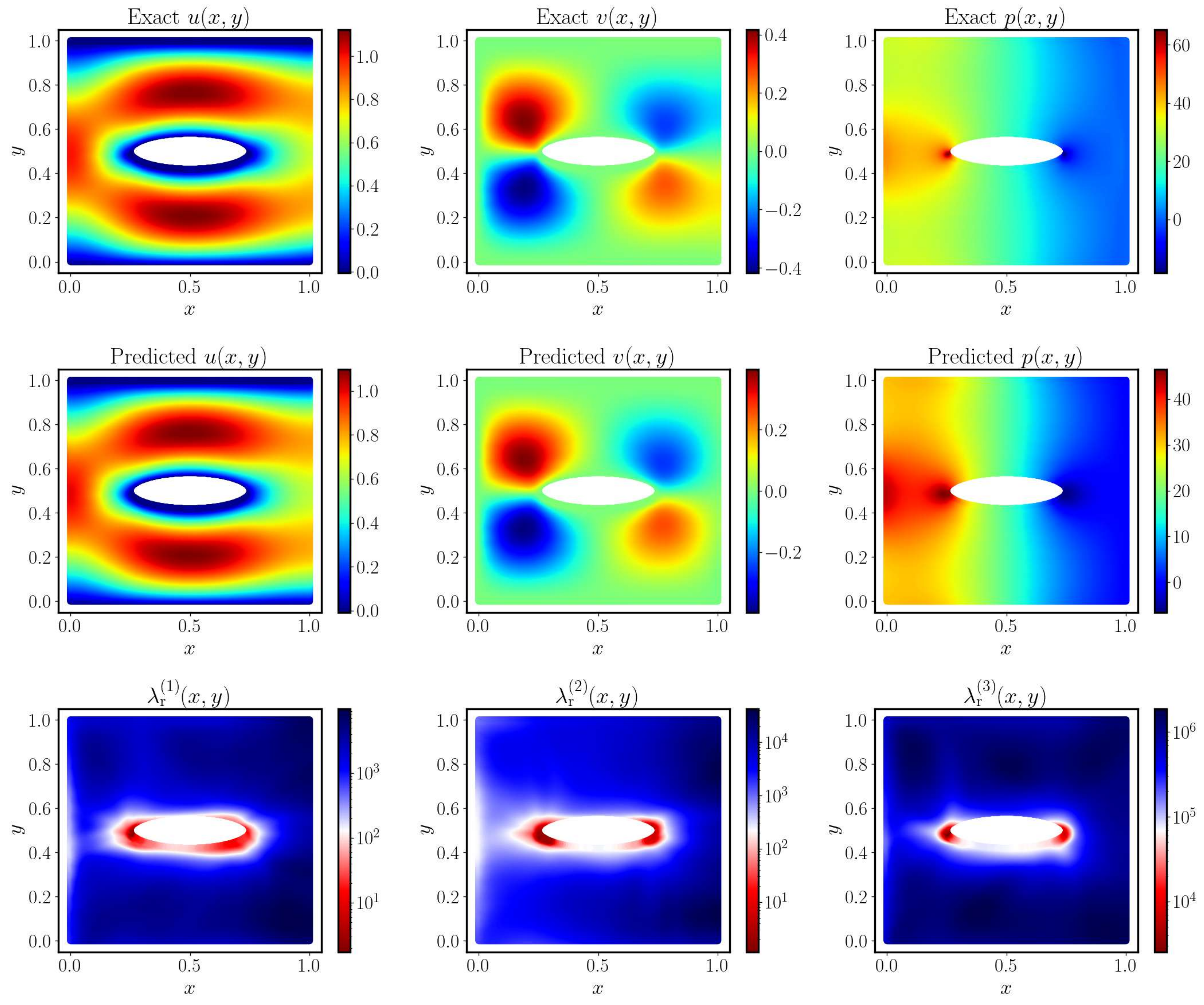}
    \caption{{\em Stokes equation:} Predicted solution of the best trained physics-informed DeepONet for one example in the test data-set.}
    \label{fig: Stokes_PI_deeponet_modified_deeponet_local_NTK_weights_pred_91}
\end{figure}

\begin{figure}[h]
    \centering
    \includegraphics[width=0.8\textwidth]{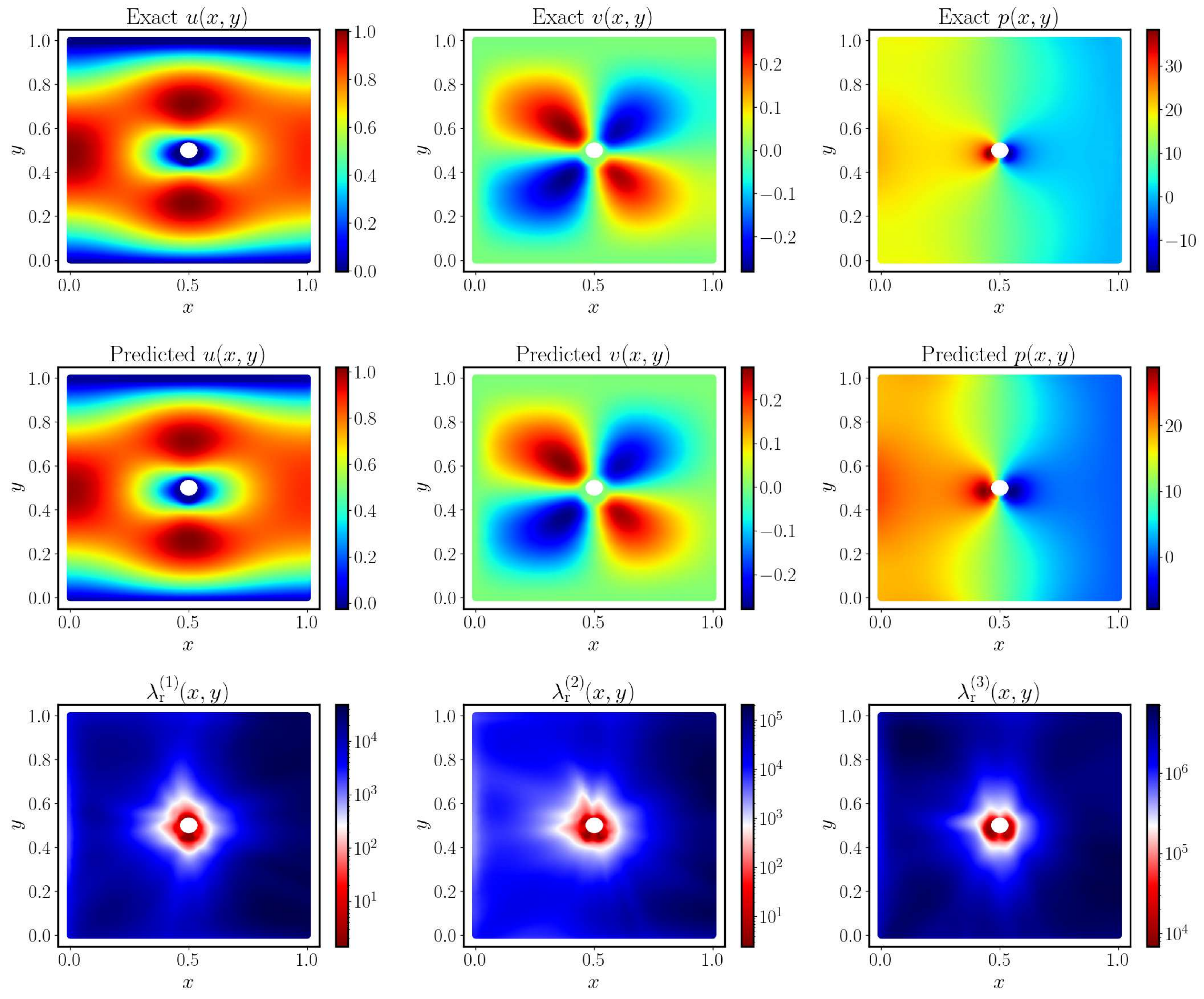}
    \caption{{\em Stokes equation:} Predicted solution of the best trained physics-informed DeepONet for one example in the test data-set.}
    \label{fig: Stokes_PI_deeponet_modified_deeponet_local_NTK_weights_pred_144}
\end{figure}

\end{document}